\documentclass[sn-mathphys]{sn-jnl}% Math and Physical Sciences Reference Style
\usepackage{graphicx}
\usepackage{amsmath}
\usepackage{amssymb}
\usepackage{pifont}
\usepackage{xcolor}
\usepackage{bm}
\usepackage{multicol}
\usepackage{multirow}
\usepackage{array}
\usepackage{url}
\usepackage{booktabs}
\usepackage{soul}
\usepackage[inline]{enumitem}
\usepackage{arydshln}
\usepackage{longtable}

\newcommand{\cmark}{\ding{51}}%
\newcommand{\xmark}{\ding{55}}%

\usepackage{subcaption}
\usepackage{algpseudocode}
\jyear{2022}%

\theoremstyle{thmstyleone}%
\theoremstyle{thmstyletwo}%

\theoremstyle{thmstylethree}%

\begin{document}

\title[Actors-Objects-Environment Network]{AOE-Net: Entities Interactions Modeling with Adaptive Attention Mechanism for Temporal Action Proposals Generation}

\author*[1]{\mbox{\fnm{Khoa}~\sur{Vo}}}\email{khoavoho@uark.edu}
\author[1]{\mbox{\fnm{Sang}~\sur{Truong}}}\email{sangt@uark.edu}\equalcont{These authors contributed equally to this work}

\author[1]{\mbox{\fnm{Kashu}~\sur{Yamazaki}}}\email{kyamazak@uark.edu}\equalcont{These authors contributed equally to this work}

\author[2]{\fnm{Bhiksha}~\sur{Raj}}\email{bhiksha@cs.cmu.edu}

\author[3,4]{\\\fnm{Minh-Triet}~\sur{Tran}}\email{tmtriet@hcmus.edu.vn}

\author[1]{\fnm{Ngan}~\sur{Le}}\email{thile@uark.edu}

\affil[1]{\orgdiv{AICV Lab}, \orgname{University of Arkansas}, \orgaddress{\city{Fayetteville}, \state{Arkansas}, \country{USA}}}

\affil[2]{\orgname{Carnegie Mellon University}, \orgaddress{\city{Pittsburgh}, \state{Pennsylvania}, \country{USA}}}

\affil[3]{\orgname{University of Science}, \orgaddress{\city{Ho Chi Minh City}, \country{Vietnam}}}

\affil[4]{\orgname{Vietnam National University}, \orgaddress{\city{Ho Chi Minh City}, \country{Vietnam}}}


\abstract{Temporal action proposal generation (TAPG) is a challenging task, which requires localizing action intervals in an untrimmed video. Intuitively, we as humans, perceive an action through the interactions between actors, relevant objects, and the surrounding environment. Despite the significant progress of TAPG, a vast majority of existing methods ignore the aforementioned principle of the human perceiving process by applying a backbone network into a given video as a black-box. In this paper, we propose to model these interactions with a multi-modal representation network, namely, \emph{Actors-Objects-Environment Interaction Network (AOE-Net)}. Our AOE-Net consists of two modules, i.e., perception-based multi-modal representation (PMR) and boundary-matching module (BMM). Additionally, we introduce \emph{adaptive attention mechanism (AAM)} in PMR to focus only on main actors (or relevant objects) and model the relationships among them. PMR module represents each video snippet by a visual-linguistic feature, in which main actors and surrounding environment are represented by visual information, whereas relevant objects are depicted by linguistic features through an image-text model.
BMM module processes the sequence of visual-linguistic features as its input and generates action proposals.
Comprehensive experiments and extensive ablation studies on ActivityNet-1.3 and THUMOS-14 datasets show that our proposed AOE-Net outperforms previous state-of-the-art methods with remarkable performance and generalization for both TAPG and temporal action detection. To prove the robustness and effectiveness of AOE-Net, we further conduct an ablation study on egocentric videos, i.e. EPIC-KITCHENS 100 dataset. Our source code is publicly available at \url{https://github.com/UARK-AICV/AOE-Net}.}


\maketitle

\section{Introduction}

\label{sec:intro}
\begin{figure}
     \centering
     \begin{subfigure}[b]{0.95\textwidth}
         \centering
         \includegraphics[width=\textwidth]{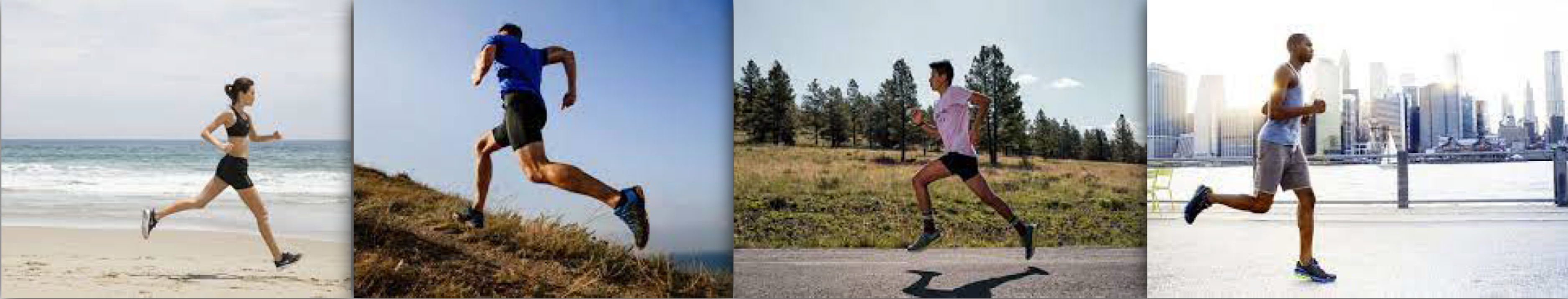}
         \caption{Examples of actions (e.g jogging) are independent to environments.}
         \label{fig:motivation1}
     \end{subfigure}
     \hfill
     \begin{subfigure}[b]{0.95\textwidth}
         \centering
         \includegraphics[width=\textwidth]{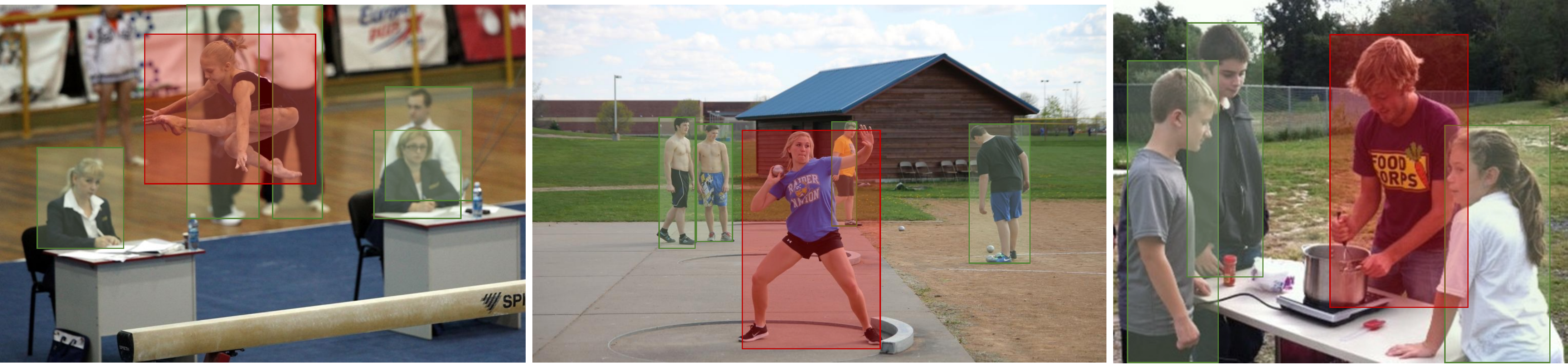}
         \caption{Examples of how actors contribute to form actions i.e. among all actors (green and red boxes) in the scenes, only main actors (red boxes) actually commit actions.}
         \label{fig:motivation2}
     \end{subfigure}
     \hfill
     \begin{subfigure}[b]{0.95\textwidth}
         \centering
         \includegraphics[width=\textwidth]{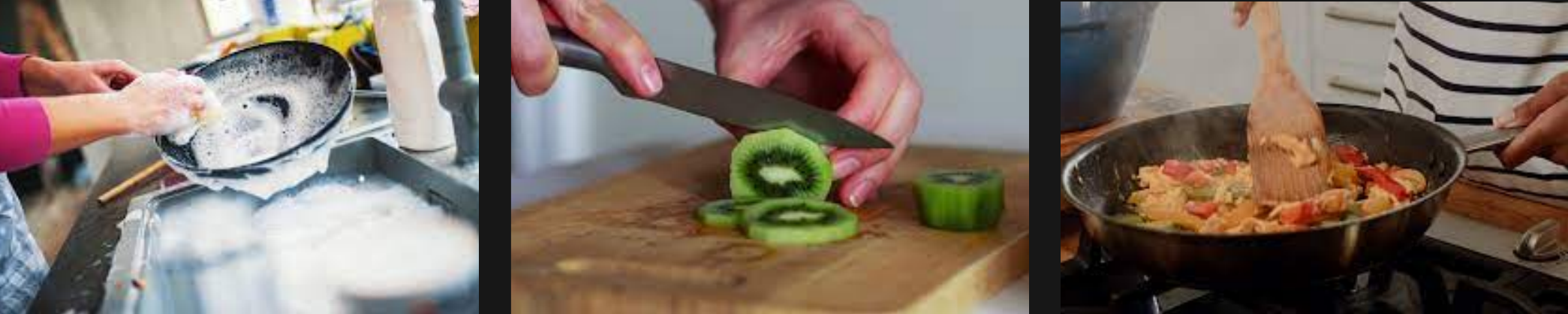}
         \caption{Examples of actions in egocentric videos where actors are not visible.}
         \label{fig:motivation3}
     \end{subfigure}
     \hfill
     \begin{subfigure}[b]{0.95\textwidth}
         \centering
         \includegraphics[width=\textwidth]{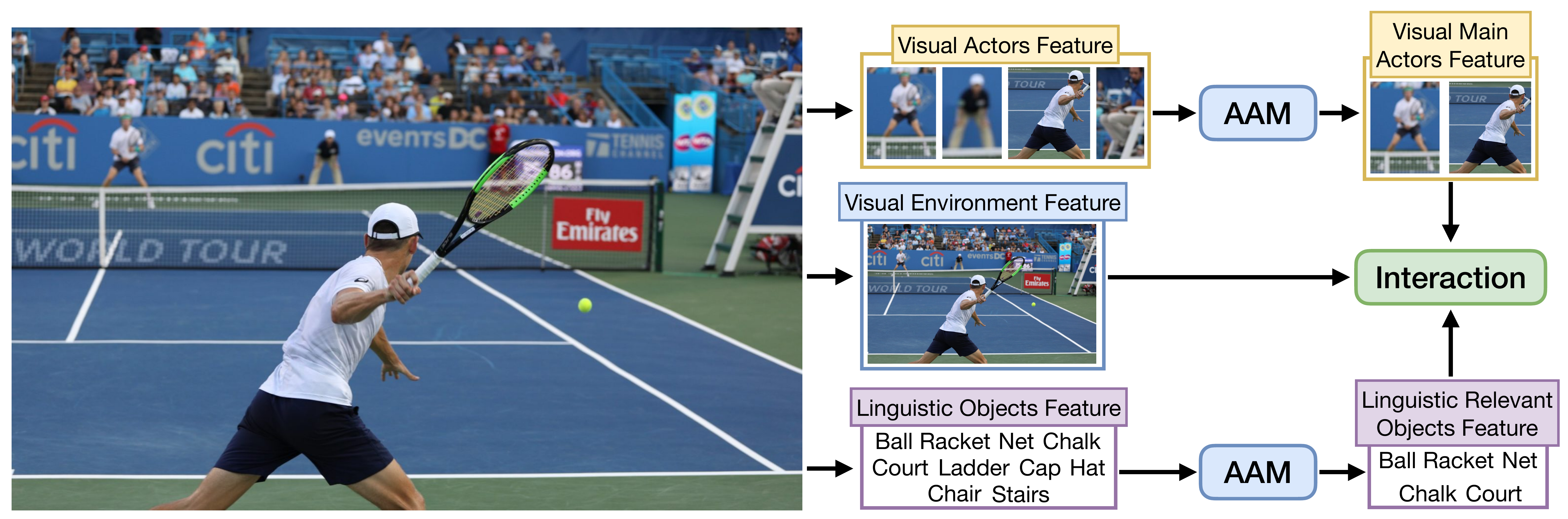}
         \caption{Our proposed AOE-Net is modeled by both global visual environment, local visual main actors features, linguistic relevant objects features, and the interaction among them. In AOE-Net, AAM is our proposed adaptive attention mechanism to select main actors and relevant objects.}
         \label{fig:ourAOE}
     \end{subfigure}
        \caption{Most existing TAPG methods \cite{lin2018bsn, BSN++, bmn, dbg, xu2020gtad} apply a 3D backbone network to entire spatial domain. However, as shown in (a), actors contribute more importance to an action than environment itself. The SOTA in TAPG \cite{KhoaVo_Access, KhoaVo_ICASSP} extract both local humans features and global environment feature; however, they are unable to either distinguish between main actors who actually commit actions and inessential actors (b) or address egocentric videos where actors are not visible in the scene (c). Our proposed AOE-Net, as illustrated in (d), consists of the global visual environment, local visual main actors features, and linguistic relevant objects features.}
        \label{fig:motivation}
\end{figure}

Given an untrimmed video, TAPG targets localizing temporal segments with specific starting and ending timestamps for each action or activity appearing in the video. TAPG has emerged as one of the most important problems in video analysis and understanding \cite{anchor_2, Jiyang2017, CTAP, Gao_2018_CVPR}. More specifically, TAPG is a key module for other downstream tasks including temporal action detection (TAD) \cite{caba2015activitynet, THUMOS14}, video captioning \cite{krishna2017dense}, action recognition \cite{Kinetics}, etc. In general, TAPG approaches can be divided into two main categories i.e. anchor-based approaches and boundary-based approaches. Inspired by anchor-based object detection in 2D images, anchor-based TAPG methods \cite{actionproposal_2016, FasterR_CNN_Action, anchor_1, anchor_2, anchor_3} pre-define a set of anchor segments and try to fit them into groundtruth action segments in videos. Even though a regression network has been applied to refine the proposals, anchor-based TAPG methods cannot fit all groundtruth actions with diverse lengths by a finite number of anchors. Boundary-based TAPG methods \cite{lin2018bsn, BSN++, bmn, dbg, xu2020gtad, KhoaVo_ICASSP, KhoaVo_Access} address the previous limitations by first separately localizing  starting and ending timestamps of exiting actions and then fusing them by a follow-up action evaluation module.

Despite good achievements on benchmarking datasets, boundary-based approaches \cite{lin2018bsn, BSN++, bmn, dbg, xu2020gtad} still have some limitations, in which the most major one is the overlooked video representation. In such designs, a video is split into consecutive snippets (or clips, chunks) of $\delta$ frames; then, a 3D convolutional backbone network \cite{C3D, i3d_2017, 2_stream_1, SlowFast} is simply applied to the entire spatial domain of each snippet to extract its visual representation. However, not all spatial regions in a snippet are relevant nor contribute to the formation of an action. Specifically, as shown in Fig.~\ref{fig:motivation}(a), an actor itself rather than spatial environment influences the action i.e. jogging can be created anywhere regardless of environment. To address those limitations, \cite{KhoaVo_ICASSP, KhoaVo_Access} recently propose to separately represent each snippet by both local actors features and global surrounding environment features. Both features are combined by a self-attention module to flexibly balance between local and global visual representation. 
%The actors visual features are extracted through a human detector \cite{FasterRCNN} and RoIAlign \cite{MaskRCNN_ICCV17} on the feature map extracted by \cite{C3D}. 
Although the improvements reported \cite{KhoaVo_ICASSP, KhoaVo_Access} are very promising, those paradigms are unable to discriminate main actors who actually commit actions from the inessential actors, as shown in Fig.~\ref{fig:motivation}(b). Additionally, both AEN and ABN may not be helpful in many videos where actions are not dependent on the presence of humans, i.e., egocentric videos, as shown in Fig.~\ref{fig:motivation}(c).

% language -> objects which are tiny and less motion -> visual feature is not appropriate 

Intuitively, besides actors and the environment, we, as human beings, also perceive an action through the presence of relevant objects and their interactions with the surroundings.
%In the other hand, an action is formed by main actors who actually commit actions, relevant objects, environment, and interactions between them.
However, unlike actors, relevant objects are often tiny with few pixels (e.g., less than $20 \times 20$ pixels). This causes the problem of vanishing information if we obtain objects from visual feature maps extracted by some typical CNN-based backbone. A possible solution is to leverage ``vision and language" methods \cite{mei2020vision, anderson2018bottom, radford2021learning} to represent objects by linguistic features. By this way, information about the presence of every object is fully preserved. We leverage CLIP \cite{radford2021learning}, which is a powerful model to associate both vision and language, to extract linguistic features from relevant objects existing in video snippets. As a result, relevant objects are represented by linguistic features whereas environment and main actors are represented by visual features. Fig.~\ref{fig:motivation}(d) illustrates our intuition for video representation.  
%As shown in Fig.~\ref{fig:motivation}(d), environment and main actors are presented by visual feature while relevant objects are presented by linguistic feature.

%Furthermore, vision and language are two primary capabilities of humans, i.e., humans perform actions/tasks through the interactions between vision and language \cite{mei2020vision}. Today, modeling "vision and language" (V\&L) has become one of the most popular research domains in both computer vision and natural language processing (NLP). Specifically, text generation from unstructured perceptual input, e.g., image or video, has posed an important challenge. Most existing V\&L models rely on a visual encoder to perceive the visual world by translating raw pixels into vectors. In such designs, visual representation has become bottleneck \cite{anderson2018bottom}. Recently, CLIP \cite{radford2021learning} proposes to learn visual representation with language supervision. CLIP is trained on 400M noisy image-text pairs crawled from the Internet, and it consists of a visual encoder and a text encoder. CLIP shows strong zero-shot power on ImageNet classification. Unlike main actors, objects are often tiny with few pixels (e.g., less than $20 \times 20$ pixels). This causes the problem of vanishing features because of the inclusion of deeper convolutional layers. Thus, we leverage CLIP to extract linguistic features from relevant objects existing in videos. As shown in Fig.~\ref{fig:motivation}(d), environment and main actors are presented by visual feature while relevant objects are presented by linguistic feature.

In this paper, we propose a novel \emph{Actors-Objects-Environment Interaction Network (AOE-Net)} as a simulation of human perception in modeling the video by visual features from main actors and environment as well as linguistic features from relevant objects. Our AOE-Net consists of two main modules, i.e., (i) perception-based multi-modal representation (PMR) to extract visual-linguistic (V-L) feature and model actors-objects-environment relations in each snippet, (ii) boundary-matching module (BMM) to localize and generate action proposals. To select only main actors along with choosing relevant objects and extract mutual relationships among each of these entities, we propose a novel \emph{adaptive attention mechanism (AAM)}.
%In PMR, detected actors and environment are presented by visual features through AAM, whereas relevant objects are presented by linguistic feature using AAM.

\noindent\textbf{Our contributions are summarized as follows:}
\begin{itemize} %[leftmargin=*]
    \item We propose a novel network, AOE-Net, which follows the human perception process to understand human actions.
    \item We introduce a novel and effective attention module, AAM, which simultaneously selects main actors (or relevant objects) and eliminates inessential actor(s) (or objects), then, extracts semantic relations between main actors (or relevant objects).
    \item Our proposed AOE-Net achieves the SOTA performance on common benchmarking datasets of ActivityNet-1.3 and THUMOS-14 in both TAPG and TAD tracks with a large margin compared to previous works.
    \item We investigate the robustness of AOE-Net while working on egocentric videos of EPIC-KITCHENS 100, in which the main actor is absent from the video views.
    \item We provide ablation studies on the contribution of each entity type (i.e., main actors, relevant objects, and environment) as well as various combinations among them.
    \item Extensive ablation studies and qualitative analysis of AAM are also provided to investigate its effectiveness.
\end{itemize}

\begin{figure*}[t]
\centering
  \includegraphics[width=\linewidth]{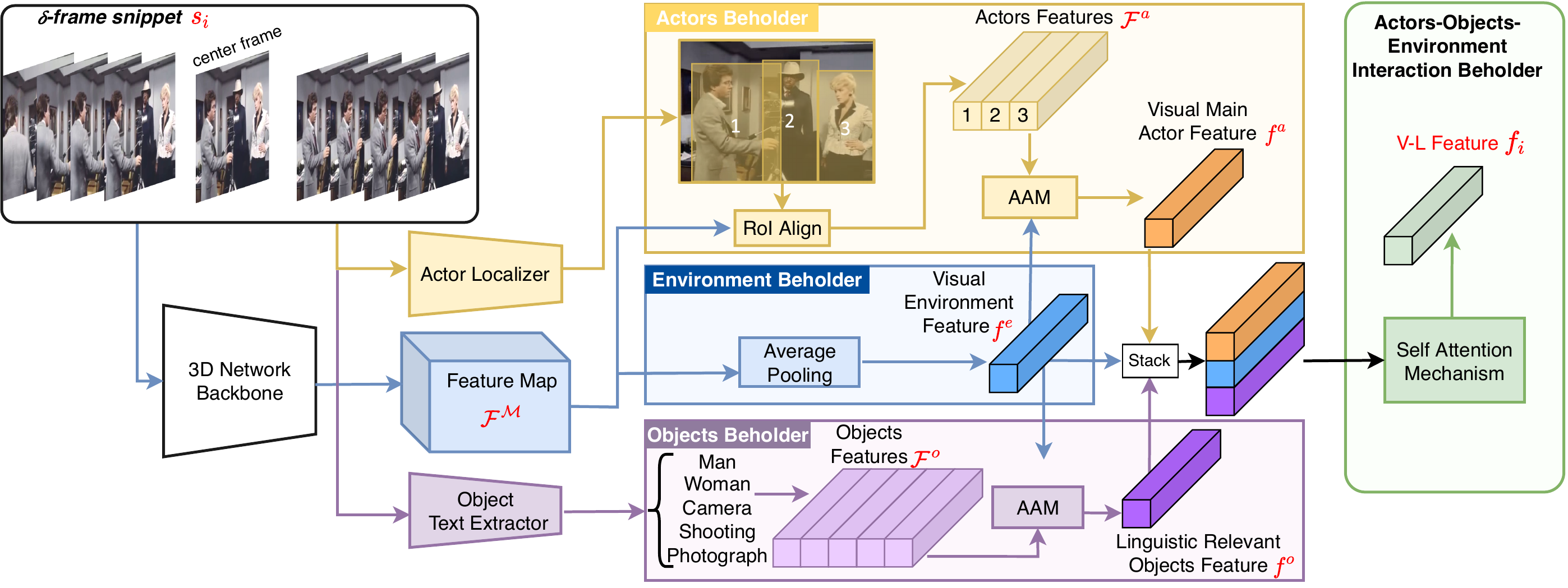}
  \caption{The architecture of our proposed PMR. Given a $\delta$-snippet $s_i$, the V-L feature is obtained by four modules: (i) actors beholder to extract local visual action feature $f^a$; (ii) environment beholder to extract global visual environment feature $f^e$; (iii) objects beholder to extract linguistic object feature $f^o$, and (iv) actors-objects-environment interaction beholder to model V-L feature as the interaction between actors, objects and the environment.}
  \label{fig:AOE-PMR}
\end{figure*}

\section{Related Works}
\noindent
\textbf{Temporal Action Proposal Generation (TAPG)} \\
As stated above, prior works can be categorized into two groups: anchor-based and boundary-based. Anchor-based methods \cite{heilbron2016fast, FasterR_CNN_Action, anchor_1, anchor_2, anchor_3} are inspired by anchor-based object detection methods \cite{FasterRCNN, RetinaNet, yolov3}, i.e., they pre-define a set of fixed segments and learn to fit them into groundtruth action segments in the video. Among them, \cite{heilbron2016fast} uses space-time interest points and dictionary learning. \cite{anchor_2} makes use of C3D \cite{C3D_3} to build a binary classification task to generate proposal segments. TURN \cite{anchor_3} divides a video into units and employs unit-level features with a temporal regression. In some of those anchor-based approaches, a regression network is also applied to refine the proposals. However, the groundtruth proposals vary a lot in terms of duration, which discourages the performance of anchor-based methods. Boundary-based methods \cite{boundary_0, lin2018bsn, BSN++, liu2019multi, bmn, dbg, xu2020gtad, KhoaVo_ICASSP, KhoaVo_Access} resolve this problem by initially localizing the starting and ending timestamps of all actions appearing in the video and then matching them by an action evaluation module, which estimates the actionness score of every possible pair of boundaries. Among them, \cite{boundary_0} adopts a watershed algorithm to group contiguous high-score as proposals. \cite{lin2018bsn} first predicts each temporal point as either starting or ending point of an action and then evaluates proposals. \cite{CTAP} combines both anchor-based and boundary-based approaches to reorganize the candidate set. \cite{liu2019multi} employs a bilinear matching module to perform TAPG at two distinct granularities. Generally, the boundary-based methods provide superior performance over anchor-based methods.

Our AOE-Net belongs to the second category. Different from existing boundary-based methods, AOE-Net is based on V-L feature by leveraging the human perception principle.

\vspace{0.2cm}
\noindent
\textbf{Attention Module} \\
Attention Models (Atts) have a long history~\cite{itti1998model} and have become an important concept in neural networks \cite{chaudhari2021attentive}. Atts can be divided into two main groups: Soft-Attention Models (Soft-Atts) and Hard-Attention Models (Hard-Atts). \cite{bahdanau2014neural} was one of the first Soft-Atts that was applied to machine translation. Because of its differentiable architecture, which helps the whole model learn in an end-to-end fashion, Soft-Atts has become an essential component in a large number of applications (e.g., speech \cite{cho2015describing}, NLP \cite{galassi2020attention}, computer vision \cite{chaudhari2019attentive}).
%Notably, \cite{attention_is_all_you_need} proposes a self-attention network based on Soft-AN in machine translation. Because of its flexible capability to model the relations between elements of input, \cite{attention_is_all_you_need} has been popularly applied into language models and computer vision.
Because self-attention networks \cite{attention_is_all_you_need} are able to learn the relations between input elements regardless of their quantity, their popularity is increasing in not only language models but also in computer vision.
Hard-Atts was first introduced in \cite{xu2015show} and \cite{Saccader_NIPS2019} for digit and object classifications, respectively. 
%In these works, a policy network is trained by reinforcement learning methods to recurrently attend to different regions of the image for several timesteps before classifying it. 
Hard-Atts aims to mask out irrelevant elements of the inputs by sampling the input elements with probabilities to reduce the distractions. This is an advanced benefit over Soft-Atts; however, Hard-Att in \cite{xu2015show} is indifferentiable. Recently, \cite{patro2018differential} proposes a Hard-Att that can be trained by normal gradient back-propagation, with a fundamental observation that the L2-norm values of more important features are usually higher than those of less important features in a feature map.

In this work, we propose an \emph{adaptive attention model (AAM)}, which leverages both the differentiable Hard-Att \cite{patro2018differential} and the self-attention network \cite{attention_is_all_you_need}.

\section{Our Method}
%\subsection{Problem Statement}
%\label{subsec:problem}
% CLIP as Linguistic Representation

Given an input video $\mathcal{V}=\{v_i\}_{i=1}^{N}$, where $N$ is the number of frames, we follow the standard settings from existing works to divide $\mathcal{V}$ into a sequence of $\delta-$frame \textit{snippets} $s_i\mid_{i=1}^T$. Each snippet $s_i$ consists of $\delta$ consecutive frames, therefore, $\mathcal{V}$ has a total of $T=\bigr\lceil \frac{N}{\delta} \bigr\rceil$ snippets. Let $\phi(.)$ be an encoding function to extract the visual feature $f_i$ of a $\delta$-frame snippet $s_i$; the video $\mathcal{V}$ can be represented as $\mathcal{F}$ as follows:
\begin{equation}
        \mathcal{F} =\{f_i\}_{i=1}^{T}, \text{ where } f_i  =\phi(s_i) 
\end{equation}

Different from the existing works \cite{BSN++,dbg, gtan_cvpr2019, xu2020gtad, tsi_accv, bmn, lin2018bsn, xu2020gtad, bai2020boundary, tan2021relaxed}, which simply define $\phi(.)$ as a pre-trained backbone network (e.g., C3D \cite{C3D}, 2Stream \cite{2_stream_1}, Slow-fast \cite{SlowFast}), we model $\phi(.)$ by the proposed PMR, which is capable of representing visual information of the snippet in both global and local perspectives, using both visual and linguistic information.

Given the feature sequence $\mathcal{F}$, the boundary-matching module (BMM) has a role of localizing action proposals.
In this section, we introduce PMR in Sub-Sec.~\ref{subsec:pmr}. Then, we present the boundary-matching module in Sub-Sec.~\ref{subsec:bmm}.

\subsection{Perception-based Multi-modal Representation (PMR)}
\label{subsec:pmr}

PMR aims to extract features based on the principle of how a human perceives an action (i.e., identify the main actors at each temporal period, recognize relevant objects and understand interactions between main actors, relevant objects, and the environment) to specify when the action starts and ends. In this paper, we are interested in discovering two modalities of vision and language to extract V-L feature. PMR consists of four main components: (i) environment beholder; (ii) actors beholder; (iii) objects beholder; and (iv) actors-objects-environment interaction beholder. The overall architecture of PMR is shown in Fig.~\ref{fig:AOE-PMR}.

%To encode both the appearance and motion taking place in a $\delta$-frame snippet, we adapt a 3D network pre-trained on action recognition benchmarking datasets as a backbone feature extractor. Since the task of the Environment beholder is to extract global semantic information of each snippet, we process the snippet through all convolutional blocks and several fully connected layers of the backbone network to obtain a single feature vector $f^{e}$, which captures information of all spatial dimensions of the snippet.
%\hl{Should move this to implementation details}
%For ActivityNet-1.3 dataset \cite{caba2015activitynet}, we employ C3D \cite{C3D} pre-trained on Kinetics-400 \cite{Kinetics} as our backbone network and remove its softmax and classification layers to keep a more semantic feature vector, which is in 2048 dimensions, $f^{e}\in\mathbb{R}^{2048}$.

%Contrarily, for the THUMOS-14 dataset \cite{THUMOS14}, we follow \cite{lin2018bsn, bmn, xu2020gtad} to employ TSN \cite{TSN2016ECCV} pre-trained on untrimmed action recognition track of ActivityNet-1.3 dataset. Additionally, to conduct a fair comparison with previous works, we keep the same settings as it was used in previous works to extract a 400-dimension feature vector from the last layer of TSN \cite{TSN2016ECCV}, $f^{e}\in\mathbb{R}^{400}$.

%\vspace{1mm}
%\textbf{(i) Environment Beholder:}
\subsubsection{Environment Beholder:}
\label{subsubsec:env}
This component has the role of capturing the global visual information of an input $\delta$-frame snippet. To extract the spatio-temporal information of the snippet, we adopt a 3D network pre-trained on action recognition benchmarking datasets as a backbone feature extractor. First, the snippet is processed through all convolutional blocks of the 3D network to obtain a feature map $\mathcal{F^M}$ at the final block; then, an average pooling operator is employed to produce a spatio-temporal feature vector $f^{e}$.

%\vspace{1mm}
%\textbf{(ii) Actors Beholder:}
\subsubsection{Actors Beholder:}
\label{subsubsec:actors}
This component semantically extracts visual main actors representation $f^a$. In most cases, an action cannot happen if a human (main actor) is absent notwithstanding environments (Fig.~\ref{fig:motivation}(a)). However, when an action occurs, it does not necessarily signal that every actor in the scene has committed the action (Fig.~\ref{fig:motivation}(b)). Herein, the actors beholder first localizes all existing actors in a $\delta$-frame snippet. To do so, we apply a human detector onto the middle frame assuming that the actors would not move fast enough to be mislocated with a small $\delta$. We denote $\mathcal{B}=\{b_i\}_{i=1}^{N_B}$ as a set of detected human bounding boxes, where $N_B \geq 0$. Afterwards, each of the detected bounding boxes, $b_i$, is aligned onto feature map $\mathcal{F^M}$, which is obtained by the 3D network backbone from environment beholder, using RoIAlign \cite{MaskRCNN_ICCV17}. Then, each bounding box feature is average-pooled into a single feature vector $f^a_i$. Finally, we obtain a set of actor features $\mathcal{F}^a=\{f^a_i\}^{N_B}_{i=1}$.

To adaptively select an arbitrary number of main actors and extract their mutual relationships, we apply our proposed AAM (described in Sub-Sec. \ref{sec:aam}), which is elaborately explained in Sub-Sec.~\ref{sec:aam} and illustrated in Fig.~\ref{fig:AOE-AAM}.

\begin{figure}[!t]
    \centering
    \includegraphics[width=0.9\textwidth]{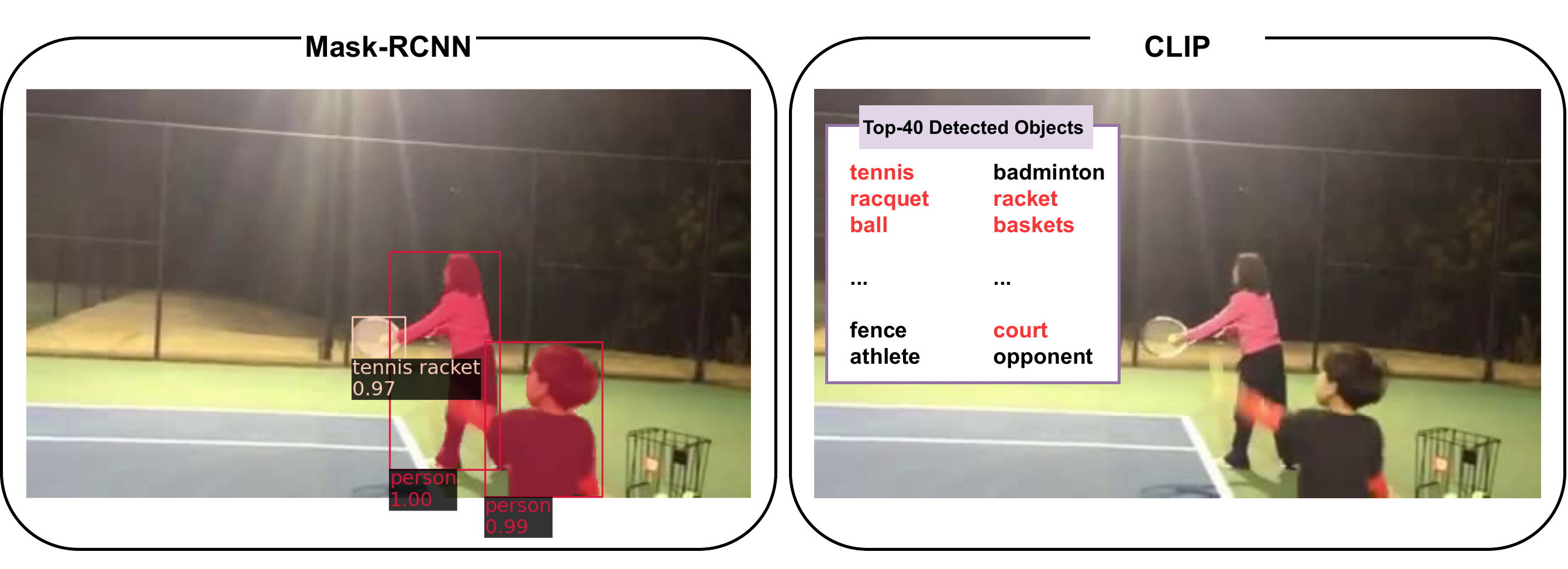}
    \caption{An example of objects detected by Mask-RCNN \cite{MaskRCNN_ICCV17} (left) vs. CLIP (right), where the most relevant objects selected by our AMM are highlighted in \textbf{\textcolor{red}{bold red}}).}
    \label{fig:maskrcnn_vs_clip}
\end{figure}

%\vspace{1mm}
%\textbf{(iii) Objects Beholder:}
\subsubsection{Objects Beholder:}
\label{subsubsec:objects}
Different from the environment and actors, objects may be tiny with a few pixels and therefore may vanish in the feature map $\mathcal{F}^\mathcal{M}$. Hence, in this objects beholder, we propose to use linguistic information from relevant objects, which is considerably more informative than visual information. We leverage CLIP~\cite{radford2021learning} as a pre-trained model to extract linguistic information.

CLIP~\cite{radford2021learning} is trained with a large number of image and description pairs, thus, CLIP effectively learns the correlation between the global scene information and local scene elements. Many scene elements are presented as small objects in the scene and they are hardly captured by an object detector. With CLIP, scene elements can be inferred by globally encoding the entire scene information. Thus, once the entire scene is captured, the small objects of scene elements are obtained accordingly.

For example, given an image of people playing tennis shown in Fig. \ref{fig:maskrcnn_vs_clip} as below, it is unfeasible to detect a small object such as a tennis ball using an object detector. As shown in Fig. \ref{fig:maskrcnn_vs_clip} (left), Mask-RCNN [49] is only able to detect humans and tennis racket while the tennis ball is not captured. Whereas, CLIP already encoded tennis scene elements including tennis ball when modeling tennis games. As shown in Fig. \ref{fig:maskrcnn_vs_clip} (right), CLIP captures tennis ball and other related objects such as basket, court, fence, etc. In this example, we choose top $K=40$ detected objects by CLIP. The most relevant objects selected by AMM are shown in \textcolor{red}{\textbf{bold red}}.

%We are taking top $K$ correlated words of a scene, so when $K$ is large enough (we empirically select $K=40$), there will be a high possibility that related objects are included in those words. Below, we include a figure to demonstrate the superiority of CLIP compared to a typical objects detector like Mask-RCNN [49].
%Because of this, CLIP might not be able to detect an abnormal object in the scene.  

\begin{figure}[t]
\centering
  \includegraphics[width=0.8\linewidth]{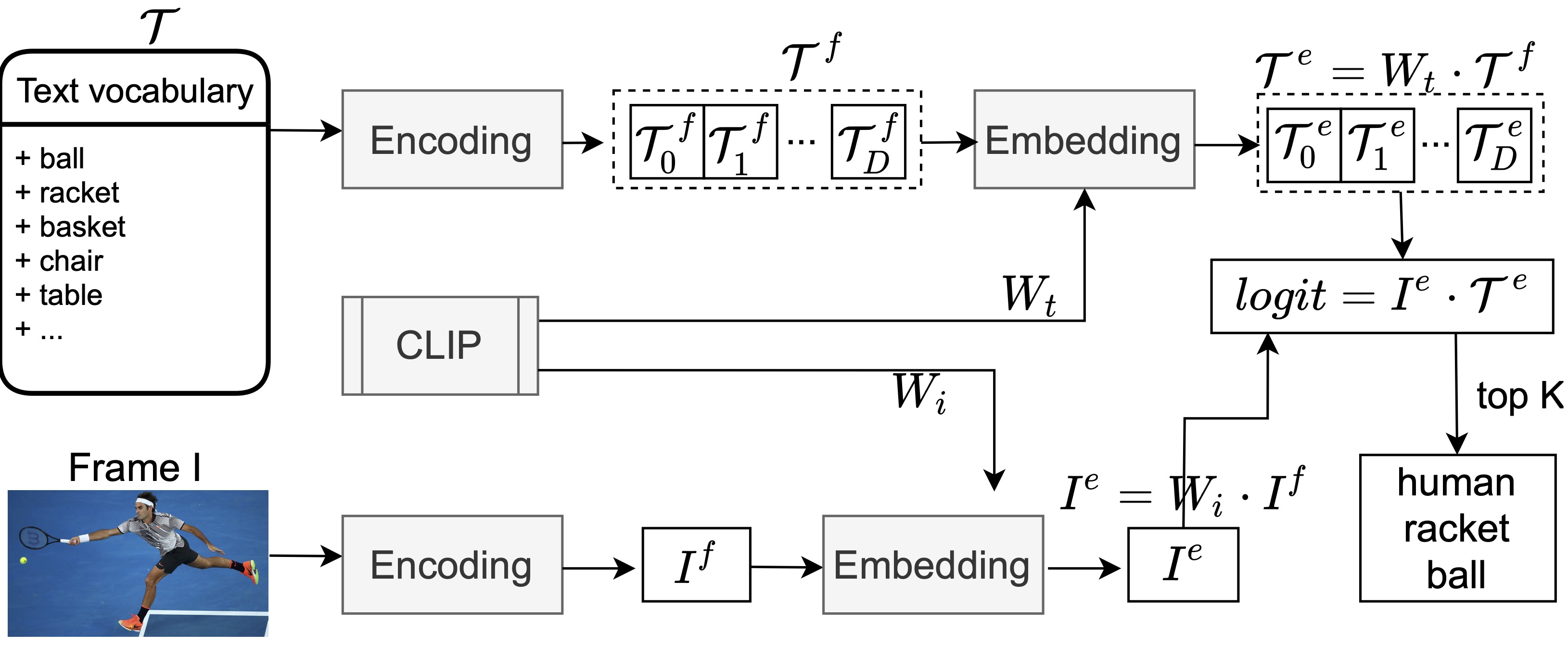}
  \caption{Illustration of object text extraction where Encoding, Embedding, and CLIP are pre-trained models from \cite{attention_is_all_you_need}, \cite{dosovitskiy2020image}, \cite{radford2021learning}, respectively.}% \vspace*{-15pt}}
  \label{fig:obj_text}
\end{figure}

%To capture the appearance of objects in each snippet, one can leverage any general objects detector that are trained on a special
Object text extraction is the first step of this component as illustrated in Fig.~\ref{fig:obj_text}. However, our task just focuses on human activities and their related objects. Therefore, we utilize the corpus of ActivityNet Captioning dataset \cite{krishna2017dense} to construct the object text vocabulary $\mathcal{T} = \{\mathcal{T}_i\}_{i=1}^{D}$.

ActivityNet Captioning dataset \cite{krishna2017dense} annotates the same set of videos in ActivityNet-1.3 \cite{caba2015activitynet}. In its training split, there are a total of 37,447 sentences to densely describe every event in each video, these captions is composed by a vocabulary of up to 10,648 words. In order to create a vocabulary which majorly contains objects and human activities, we eliminate stop words, pronouns, numbers, and infrequent words (which appears 5 times or lower in the whole dataset). Afterwards, we remove words that do not present in the vocabulary used by CLIP~\cite{radford2021learning}. Fortunately, thanks to the byte pair \cite{sennrich-etal-2016-neural} encoding used in CLIP \cite{radford2021learning}, there are very few words that are removed after in this step. To this end, the vocabulary for our objects beholder consists of $D=3,544$ words is extracted from the ActivityNet Captioning dataset \cite{krishna2017dense}.

Each word $\mathcal{T}_i\in\mathcal{T}$ is encoded by a Transformer network \cite{attention_is_all_you_need} into a text feature $\mathcal{T}^f_i$. Let $W_t$ be a text projection matrix pre-trained by CLIP, the embedding text vocabulary is computed as $\mathcal{T}^e = W_t \cdot \mathcal{T}^f$, where $\mathcal{T}^f = \{\mathcal{T}^f_i\}_{i=1}^{D}$. Let $W_i$ be an image projection matrix pre-trained by CLIP, a middle frame $I$ of the $\delta$-frame snippet is first encoded by Vision Transformer \cite{dosovitskiy2020image} to extract visual feature $I^f$, and then embedded by $W_i$, i.e., $I^e = W_i\cdot I^f$. The pairwise cosine similarities between embedded $I^e$ and $\mathcal{T}^e$ is then computed. Top $K$ similarity scores are chosen as output objects text represented by feature $\mathcal{F}^o=\{\mathcal{T}^f_i\}_{i=1}^K$. Ablation study on $K$ will be discussed in Sub-Sec~\ref{sec:topK}. Similar to the actors beholder, we apply the proposed AAM (described in Sub-Sec. \ref{sec:aam}) to select relevant objects from $\mathcal{F}^o$, then model the semantic relations among them, and finally obtain linguistic feature $f^o$.

%\vspace{0.2cm}
%\textbf{(iv) Actors-Object-Environment (AOE) Interaction beholder:}
\subsubsection{Actors-Objects-Environment (AOE) Beholder:}
\label{subsubsec:aoe}
This component aims to model the relations between global visual environment feature $f^e$, local visual of main actors features $f^a$, and linguistic relevant objects features $f^o$. Firstly, we stack three types of features together as $\mathcal{F}^{aoe}=[f^a,f^o,f^e]$. Then, we employ the self-attention model \cite{attention_is_all_you_need} followed by an average pooling layer to fuse the stack of features $\mathcal{F}^{aoe}$ into $f_i$. $f_i$ is a V-L feature that represents the input snippet $s_i$ through both visual (environment and actors modalities) and linguistic (objects modality) ways.

\begin{figure}[t]
\centering
  \includegraphics[width=0.95\linewidth]{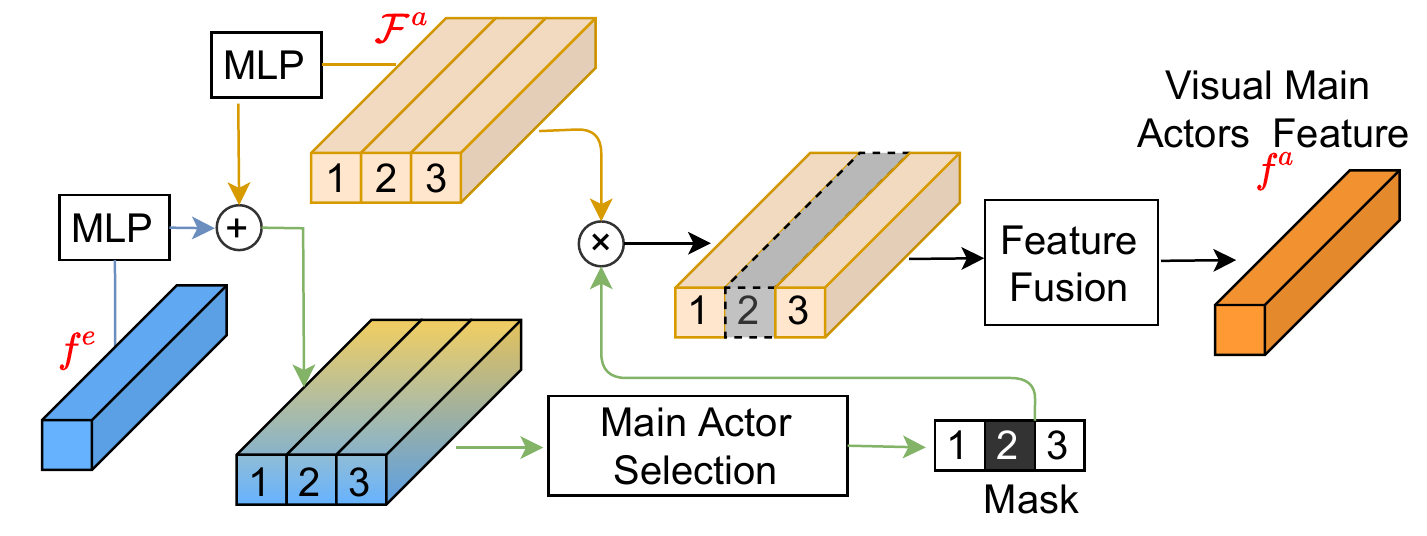}
  \caption{Illustration of proposed AAM. We choose actors features $F^a$ and environment feature $f^e$ as an example. AAM aims to select main actors features, followed by fusing arbitrary main actors features, to obtain visual main actors representation $f^a$.}
  \label{fig:AOE-AAM}
\end{figure}

\begin{algorithm}[t]
\caption{AAM to extract the representation of main actors in a snippet.}
\label{algo:aam}
\hrule
\begin{algorithmic}[1]
\algrenewcommand\algorithmicrequire{\textbf{Data: }}
\algrenewcommand\algorithmicensure{\textbf{Result: }}
\Require Feature vector $f^e$ and features set $\mathcal{F}^a$ represent environment and all actors that appear in an input snippet, respectively.
\Ensure Feature vector $f^a$ represents main actors.

\State $\hat{f}^e \gets MLP_{\theta_e}(f^e)$ %\tcp*{embed $f^e$ to common space with every $f^a_i$ in $\mathcal{F}^a$}
\State set $\tilde{\mathcal{F}}^a$, $H^a$ to empty list \Comment{\scriptsize{$\mathcal{F}^a$ stores selected main actors, $H^a$ stores scores of every actor}}
%set $\tilde{\mathcal{F}}^a$ to empty list \\%\tcp*{$\mathcal{F}^a$ will store selected main actors}
\For{each $f^a_i$ in $\mathcal{F}^a$}
    \State $\hat{f}^a_i \gets MLP_{\theta_a}(f^a_i)$ %\tcp*{embed $f^a_i$ to common space with $f^e$}
    \State $h^a_i \gets \mid\mid \hat{f}^a_i \oplus \hat{f}^e \mid\mid_2$   \Comment{\scriptsize{$\oplus$: element-wise addition}}
    \State append $h^a_i$ to $H^a$
\EndFor

\State $H^a \gets softmax(H^a)$ %\tcp*{rescale scores to sum up to $1.0$}
\State $\tau \gets \frac{1}{\mid\textbf{h}^a\mid}$ %\tcp*{compute adaptive threshold $\tau$}

\For{each $h^a_i$ in $H^a$}
    \If{$h^a_i > \tau$}
        \State append $f^a_i$ to $\tilde{\mathcal{F}}^a$ %\tcp*{select $f^a_i$ if its score higher than $\tau$}
    \EndIf
\EndFor
\State $f^a \gets self\_attention(\tilde{\mathcal{F}}^a)$ %\tcp*{fuse main actor features by self-attention \cite{attention_is_all_you_need}}
\end{algorithmic}
\end{algorithm}

\subsection{Adaptive Attention Mechanism (AAM)}
\label{sec:aam}

Given $M$ actors (or objects) obtained in the input snippet, only a few of those, i.e., $\hat{M}$ main actors (or relevant objects), actually contribute to an action. Because $\hat{M}$ is unknown and continuously changes throughout the input video, we propose AAM that inherits the merits from adaptive hard attention \cite{adahard_eccv2018} to select an arbitrary number of main actors (or objects) and a soft self-attention mechanism \cite{attention_is_all_you_need} to extract relationships among them. Take actors beholder as an instance, AAM is described by the pseudocode in Algorithm~\ref{algo:aam} and illustrated in Fig.~\ref{fig:AOE-AAM}.
%Given $N_B$ detected actors, only a few of those, i.e., main actors, actually contribute to an action. Because the number of main actors is unknown and continuously changes throughout an input video, we propose AAM that inherits the merits from adaptive hard attention to select an arbitrary number of main actors and a soft self-attention mechanism \cite{attention_is_all_you_need} to extract relationships among them. Take actors beholder as an instance; AAM is described by pseudocode in Algorithm~\ref{algo:aam} and illustrated in Fig.~\ref{fig:AOE-AAM}. In the case of objects beholder, the actors features $\mathcal{F}^a$ is replaced by the object feature $\mathcal{F}^o$.

To begin, the environment feature $f^{e}$ and actors features $\mathcal{F}^{a}$ are embedded into the same dimensional space by a multi-layer perceptrons (MLPs) parameterized by $\theta_e$ and $\theta_a$, respectively: %. $MLP_\theta(\cdot)$:
\begin{align}
\hat{f}^{e} &= MLP_{\theta_e}(f^e) \label{eq:env_mlp}\\
\hat{F}^{a} &= \{\hat{f}^{a}_i\}_{i=1}^{M} \text{ where } \hat{f}^{a}_i = MLP_{\theta_a}(f^a_i)
\end{align}

Then, $\hat{f}^{e}$ is combined with each feature $\hat{f}_i^a$ of $\hat{F}^{a}$ by element-wise addition (i.e., $\oplus$) to form a collaborative feature. Afterwards, we can compute the attention score $h_i^a$ corresponding to $\hat{f}_i^a$ using the L2-norm of its corresponding collaborative feature. These computational steps can be presented through the following equation:
\begin{equation}
    h^a_i = \mid\mid \hat{f}^a_i \oplus \hat{f}^e \mid\mid_2
\end{equation}

It is proven in \cite{adahard_eccv2018} that features with the greater L2-norm values carry more meaningful information and better contribute to later modules. % , i.e., $A_c = \{a^c_i\}_{i = 1}^{N_B}, \text{, where } a^c_i=||f^c_i||_2$.

%Then, both $\hat{f}^{e}$ and $\hat{F}^{a}$ are combined by element-wise addition (i.e., $\oplus$) to form a collaborative feature \hl{check if F-hat or F. Line 5 in Algorithm use h while here is fc}
%$H^a$: $H^a = \{h^a_i\}_{i=1}^{M}$, where $h^a_i = \hat{f}^a_i \oplus \hat{f}^e$. Afterwards, we compute the L2-norm of each collaborative feature $F^c$. It is proven that features with the greatest L2-norm values carry meaningful information and better contribute to later modules \cite{adahard_eccv2018}, i.e., $A_c = \{a^c_i\}_{i = 1}^{N_B}, \text{, where } a^c_i=||f^c_i||_2$.

Next, we re-scale all L2-norm values by softmax function to be summed up to $1.0$, because L2-norm values are unbounded:
\begin{equation}
H^a = \{ h^a_i\}_{i=1}^{M} \text{, where } h^a_i = \frac{e^{h^a_i}}{\Sigma^{M}_{i=1}e^{h^a_i}}
\end{equation}

To obtain the features of an arbitrary number of main actors, we create an adaptive threshold based on the total number of actors $\tau = \frac{1}{\mid \mathcal{F}^a \mid}$ and retrieve only features $f^a_i \in \mathcal{F}^a$ with corresponding score higher than $\tau$:
\begin{equation}
\tilde{\mathcal{F}}^a = \{ f^a_i \mid h^a_i \geq \tau \}
\label{eq:act_select}
\end{equation}

After that, we fuse a set of main actors feature vectors $\tilde{\mathcal{F}}^a$ into a single feature vector $f^a$ by leveraging the self-attention Transformer Encoder proposed in \cite{attention_is_all_you_need}.
%$\tilde{F}^a$ is fed into three MLPs to build three intermediate features $Q$, $K$, and $V$.
%Each of these has a specific role to form the relations between actors and re-weight each of them based on their relevance in the relations. 
%Each $q_i \in Q$, corresponding to $f^a_i$, is used to generate an attention mask. Each value $v_i \in V$ is re-weighted by this attention mask. The whole process is defined as:
%\begin{equation}
%    \mathcal{A}(q_i, K, V)=softmax(\frac{q_i\cdot K^T}{\sqrt{d_K}})V
%\end{equation}
%, where $d_K$ is the number of dimensions of features in $K$, following \cite{attention_is_all_you_need}. Finally, we obtain the actors visual feature $f^a$ as in Eq.\ref{eq:fa}
%\begin{equation}
%    f^a = \frac{1}{|Q|} \sum_{i}^{|Q|}\mathcal{A}(q_i, K, V)
%\label{eq:fa}
%\end{equation}.
%

In the case of objects beholder, the input actors features $\mathcal{F}^a$ is replaced by the objects features $\mathcal{F}^o$.

\begin{table}[!t]
\centering
\caption{The detailed architecture of BMM with three components. $\mathcal{F}$ is the input feature obtained from PMR. $T$ and $D$ are the temporal length of the video and maximum duration of proposals in terms of the number of snippets.} % \vspace*{-3mm}}
\resizebox{0.9\linewidth}{!}{
\footnotesize
\begin{tabular}{p{3.0cm}|p{2.4cm}|p{2.2cm}}
\hline
 \textbf{Layers} & \textbf{Input}  & \textbf{Output} \\
\hline \hline
 1DConv.  \newline $256\times3/1$, ReLU & $\mathcal{F}: F \times T$ & $O_1: 256\times T$\\
\hdashline
 1DConv. \newline $128\times3/1$, ReLU & $O_1:256\times T$ & $O_2: 128\times T$\\
\hline
 1DConv.  \newline $256\times3/1$, ReLU & $O_2:128\times T$ & $O_3: 256\times T$\\
\hdashline
 1DConv.  \newline $2\times3/1$ , Sigmoid &  $O_3:256\times T$ & $O_T: 2\times T$\\
\hline
 Matching layer & $O_2:128\times T$ & $O_5:128\times32\times D \times T$\\
\hdashline
 3DConv.  \newline $512\times32\times1\times1/(32,0,0)$ , ReLU &  $O_5:128\times32\times D\times T$ & $O_6:512\times 1 \times D \times T$\\
\hdashline
 squeeze   &  $O_6: 512\times 1 \times D \times T$ & $O_7: 512\times D \times T$\\
\hdashline
 2DConv.  \newline $128\times1\times1/(0,0)$ , ReLU &  $O_7:512\times D \times T$ & $O_8:128 \times D \times T$\\
\hdashline
 2DConv.  \newline $128\times3\times3/(1,1)$ , ReLU &  $O_8:128\times D \times T$ & $O_9:128\times D \times T$\\
\hdashline
 2DConv.  \newline $2\times1\times1/(0,0)$ , Sigmoid &  $O_9: 128\times D \times T$ & $O_P:1\times D \times T$\\
\bottomrule
\end{tabular}}
\label{tab:AOE-BMM}
\end{table}

%\vspace{-2mm}
\subsection{Boundary-Matching Module (BMM)}
\label{subsec:bmm}

\begin{figure*}[t]
\centering
  \includegraphics[width=\linewidth]{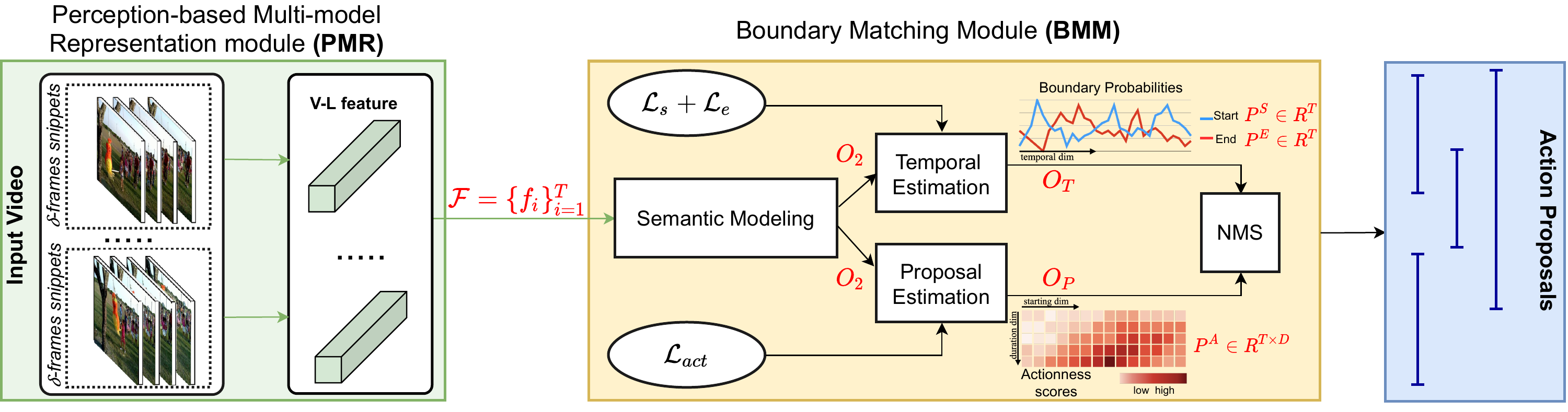}
  \caption{The overall architecture of our proposed AOE-Net, consisting of perception-based multi-model representation module (PMR) and boundary-matching module (BMM).} % \vspace*{-3mm}}
  \label{full_architecture}
\end{figure*}

BMM is responsible for localizing action boundary and generating action proposals in videos. In our AOE-Net, BMM module is adopted from previous works i.e. BSN~\cite{lin2018bsn}, BMN~\cite{bmn}, ABN ~\cite{KhoaVo_ICASSP}, AEN~\cite{KhoaVo_Access}, AEI~\cite{khoavo_aei_bmvc21} because of its standard and simple design. BMM takes the output V-L features sequence $\mathcal{F} = \{ f_i\}_{i=1}^T$ from PMR module as its input. Our BMM contains three components: semantic modeling, temporal estimation (TE), and proposal estimation (PE) as illustrated in Fig.~\ref{full_architecture}. The first component models the semantic relationship between snippets. The TE component assesses each snippet $s_i\mid_{i=1}^T$ to evaluate probabilities of action starting ($P^S_i$) and action ending ($P^E_i$) that exist in $s_i$. Meanwhile, the PE component evaluates every interval $[i,j]$ in the video to estimate its actionness score $P^A_{i,d}$, where $d=j-i$. The detailed architecture of BMM is provided in Table \ref{tab:AOE-BMM}. The semantic modeling component is implemented by two 1-D Conv. layers and outputs a feature map $O_2 \in R^{128 \times T}$. The later components, TE and PE, take $O_2$ as their input and generate $O_T \in R^{2 \times T}$ and $O_P \in R^{1 \times D \times T}$, respectively. The output $O_T$ presents probabilities of action starts ($P^S \in R^T$) and action ends ($P^E \in R^T$). The output $O_P$ contains actionness scores $P^A \in R^{D\times T}$.

At the inference stage, we search through $P^S$ and $P^E$ to select temporal locations $i$ whose $P^S_i$ or $P^E_i$ are local maximums to form sets of potential starting and ending temporal locations, respectively. Then, starting and ending locations $(s, e)$ %from those lists that satisfy timing constraint 
(e.g. $s\leq e \leq T$) are paired and become a candidate proposal with the score $s=P^S_s \cdot P^E_e \cdot P^A_{s, e-s}$. 
Based on the timestamps and scores of candidate proposals, we finally apply NMS \cite{SoftNMS, NMS} to produce the final set of temporal action proposals.

%In our AOE-Net, we adopt BMN \cite{bmn} as our boundary-matching module because of its standard and simple design.

%\vspace{-.2mm}
\subsection{Training Methodology}
\label{subsec:train_method}
\noindent\textbf{Training Labels Generation from Groundtruth:}\\
%Given a list of $N_G$ groundtruth action segments $G=\{g_i=(g^s_i, g^e_i)\}_{i=1}^{N_G}$ of input video $\mathcal{V}$, we generate the groundtruth starting labels $L^S \in [0,1]^T$, ending labels $L^E \in [0,1]^T$ and actionness labels $L^A \in [0,1]^{T\times D}$ ($D$ is a pre-defined maximum proposal length). $L^S_i$ (or $L^E_i$) carrying a value of $1$ means that there is a groundtruth starting (or ending) boundary of any action at snippet index $i$ and vice versa. Likewise, $L^A_{i,d}$ carrying a value of $1$ means that there is a groundtruth action starts at snippet index $i$ with a length of $d$ snippets.
We follow \cite{bmn, lin2018bsn} to generate the ground truth labels for training process including starting labels, ending labels for $\mathcal{L}_s, \mathcal{L}_e$ and duration labels for $\mathcal{L}_{act}$.

The starting and ending labels are generated for every snippet of the video, which are called $L_S=\{l^s_n\}_{n=1}^T$ and $L_E=\{l^e_n\}_{n=1}^T$, respectively. The boundary timestamps (starting and ending) of every action instance $a_i=(s_i, e_i)$ are rescaled into $T$-snippet range by multiplying them with $\frac{T \cdot \text{fps}}{L}$ where $\text{fps}$ is the frame rate of the video and the action instance $a_i \in \mathcal{A}$,  $\mathcal{A}=\{a_i\}_{i=1}^{M}$. After rescaling, the action instance $a_i$ becomes a new action instance $a^\delta_i=(s_i^\delta, e_i^\delta)$. For every snippet $t_n \in T$, we denote a temporal region $r_{n}=[t_n-1,t_n+1]$. Analogously, for every pair of boundaries $(s_i^\delta, e_i^\delta)$ of action $a_i^\delta$, we denote regions $r^s_i=[s_i^\delta-\frac{3}{2}, s_i^\delta+\frac{3}{2}]$ and $r^e_i=[e_i^\delta-\frac{3}{2}, e_i^\delta+\frac{3}{2}]$ as their corresponding starting region and ending region. By this formulation, we have two sets of regions $R_S=\{r^s_i\}^M_{i=1}$ and $R_E=\{r^e_i\}^M_{i=1}$ for starting and ending boundaries, respectively. Finally, starting label $l^s_n$ and ending label $l^e_n$ of a snippet $t_n$ are calculated by the following functions:

\begin{multicols}{2}
\centering
\begin{equation}
\nonumber
l^{\text{s}}_{n} = 
\begin{cases}
1,& \sum\limits^{M}_{i=1} \frac{\mid r_{n} \cap r_i^{\text{s}} \mid }{\mid r_i^{\text{s}} \mid} \geq 0.5
\\
0,              & \text{otherwise}
\end{cases}
\quad\quad\quad
    l^{\text{e}}_{n} = 
\begin{cases}
    1,&  \sum\limits^{M}_{i=1} \frac{ \mid r_{n} \cap r_i^{\text{e}} \mid }{ \mid r_i^{\text{e}} \mid } \geq 0.5 \\
    0, & \text{otherwise}
\end{cases}
\end{equation}
\end{multicols}

The duration labels for a video are gathered into a matrix $L_D \in \{0, 1\}^{D \times T}$ where $D$ is the maximum length of proposals being considered in number of snippets, as suggested in \cite{bmn}, we set $D=T$ in all of our experiments. With an element at position $(t_i, t_j)$ stands for a proposal action $a_p=(t_s=\frac{t_j\cdot T}{t_v}, t_e=\frac{(t_j+t_i)\cdot T}{t_v})$, it will be assigned by $1$ if its Interaction-over-Union with any ground truth action in $\mathcal{A}=\{a_i\}_{i=1}^{M}$ reaches a local maximum, or $0$ otherwise.

\vspace{2mm}
\noindent\textbf{Loss Functions:}\\
To train our AOE-Net with the groundtruth labels, we define the loss function $\mathcal{L}_{AOE}$ as in Eq.~\ref{eq:Laoe} where $\mathcal{L}_{s}$, $\mathcal{L}_{e}$, and $\mathcal{L}_{act}$ are loss functions corresponding to starting boundary, ending boundary and actionness score.

\begin{equation}
    \mathcal{L}_{AOE} = \mathcal{L}_{s}(P^S, L^S) + \mathcal{L}_{e}(P^E,L^E) + \mathcal{L}_{act}(P^A,L^A)
\label{eq:Laoe}
\end{equation}

We use weighted binary log-likelihood loss $\mathcal{L}_{wb}$ for $\mathcal{L}_{s}$ and $\mathcal{L}_{e}$, which is defined as follows:

\begin{equation}
\small
    \mathcal{L}_{wb}(P, L) =  \sum^{N}_{i=1}\left[\frac{L_i}{N^+}\log P_i + \frac{(1-L_i)}{N^-}\log (1-P_i)\right]
\label{eq:L_wb}
\end{equation}
\noindent
where $N^+$ and $N^-$ are the number of positives and negatives in groundtruth labels, respectively. Conversely, $\mathcal{L}_{act}(P,L)$ is defined as follows:
\begin{equation}
    \mathcal{L}_{act}(P,L)=\mathcal{L}_{wb}(P,L)+\lambda \mathcal{L}_2(P,L)
\end{equation}
, where $\mathcal{L}_2$ is the mean squared error loss and $\lambda$ is set to $10$.

To reduce time cost in the training phase of our proposed AOE-Net, actors features $\mathcal{F}^a$, objects features set $\mathcal{F}^o$ and environment feature $f^e$ are extracted in advance. Then, AAM and AOE Interaction beholder of PMR module is trained with BMM module in an end-to-end framework.

\section{Experiments}
%We conduct the experiments and comparisons between our AEI and SOTA methods on both TAPG and TAD tasks.
% Our BMM (the second module in AEI) is built upon two network architectures: CNN-based and GCN-based. We denote these architectures as \textbf{AEI-B} and \textbf{AEI-G}, respectively.
% \newline
% \newline

\subsection{Datasets and Metrics}
\noindent
\textbf{Datasets}

\noindent
% We conduct experiments on ActivityNet-1.3 \cite{caba2015activitynet} and THUMOS-14 \cite{THUMOS14} datasets. The former contains 20,000 videos with 200 annotated activities while the latter consists of 414 videos with 20 actions. For both datasets, we follow previous works \cite{lin2018bsn,bmn, dbg} to preprocess videos with the snippet length set to $\delta=16$.
Our experiments on TAPG and TAD are carried out using both ActivityNet-1.3 \cite{caba2015activitynet} and THUMOS-14 \cite{THUMOS14} datasets. The former features 20K videos and 200 activities that have been annotated, whereas the latter has 414 videos and 20 types of actions. We follow prior works \cite{lin2018bsn,bmn, dbg} for videos preprocessing with the snippet length set to $\delta=16$ in all experiments. To prove the effectiveness of our proposed AOE-Net on egocentric videos, we also conduct an experiment on TAPG task of EPIC-KITCHENS 100 dataset \cite{damen2021rescaling}, which consists of 100 video hours, 20M frames, 90K actions in 700 variable-length videos captured in 45 environments using head-mounted cameras.
 
\vspace{2mm}
\noindent\textbf{Metrics}

\noindent In TAPG, we use two common metrics, i.e., AR@AN and AUC, to evaluate the proposed AOE-Net as well as compare it with SOTA approaches. The former metric is the average recall (AR) calculated at a specific average number of proposals (AN) preserved by each video. The latter one is the area under the AR versus the AN curve score. AR@100 and AUC are the most commonly used metrics in ActivityNet-1.3. In THUMOS-14, however, just AR@AN is utilized to compare approaches; nonetheless, multiple AN are chosen from a list of [50, 100, 200, 500, 1000].

In TAD, we use mean Average Precision (mAP) to benchmark approaches. Following the common settings \cite{lin2018bsn, dbg, bmn, tsi_accv, MR_eccv2020}, we evaluate TAD methods in ActivityNet-1.3 with tIoU thresholds of \{0.5, 0.75, 0.95\}, and average mAP. Whereas, TAD methods in THUMOS-14 are evaluated with tIoU thresholds of \{0.3, 0.4, 0.5, 0.6, 0.7\}.

\begin{table*}[!tb]
\centering
%\vspace*{0.2cm}
\resizebox{0.95\linewidth}{!}{
\begin{tabular}{lll|ccc}
                    Methods  & Venue \& Year & Feature  & AR@100 & AUC(val) & AUC(test) \\
\hline\hline
TCN \cite{TCN}            & ICCV17 & 2Stream & --      & 59.58    & 61.56     \\
MSRA \cite{MSRA}          & CVPRW17 & P3D    & --      & 63.12    & 64.18     \\
SSTAD \cite{SSTAD_BMVC17} & BMVC17 & C3D     & 73.01  & 64.40    & 64.80     \\
CTAP \cite{CTAP}          & ECCV18 & 2Stream & 73.17  & 65.72    & --         \\
BSN   \cite{lin2018bsn}   & ECCV18 & 2Stream & 74.16  & 66.17    & 66.26 \\
SRG  \cite{SRG}           & IEEE-TCSVT19 & 2Stream & 74.65 & 66.06 & -- \\
MGG \cite{liu2019multi}   & CVPR19 & I3D & 74.54  & 66.43    & 66.47     \\
BMN   \cite{bmn}      & ICCV19 & 2Stream & 75.01  & 67.10    & 67.19     \\
DBG   \cite{dbg}      & AAAI20 & 2Stream & 76.65  & 68.23    & 68.57     \\
BSN++   \cite{BSN++}  & ACCV20 & 2Stream & 76.52  & 68.26    & --         \\
TSI++ \cite{tsi_accv} & ACCV20 & 2Stream & 76.31  & 68.35    & 68.85     \\
MR\cite{MR_eccv2020}  & ECCV20 & I3D & 75.27 & 66.51 & --         \\
AEN \cite{KhoaVo_ICASSP}  & ICASSP21 & C3D & 75.65 & 68.15 & 68.99 \\
ABN \cite{KhoaVo_Access}  & IEEE-Access21 &  C3D   & 76.72 & 69.16 & 69.26 \\
%ABN \cite{KhoaVo_Access}  & IEEE-Access21 &  C3D   & 76.72 & 69.16 & 69.26 \\
SSTAP \cite{wang2021self} & CVPR21        & I3D & 75.54 & 67.53 & -- \\
TCANet \cite{qing2021temporal} & CVPR21   &  2Stream   & 76.08 & 68.08 & -- \\
Zheng, et.al. \cite{zheng2021boundary} & NPL21 & 2Stream & 74.93 & 65.20 & -- \\
AEI \cite{khoavo_aei_bmvc21}  & BMVC21 &  C3D   & \underline{\textit{77.24}} & \underline{\textit{69.47}} & \underline{\textit{70.09}} \\
% VSGN \cite{zhao2021video} & CVPR21 & & & \\ %not report AR AUC
\midrule
\multirow{3}{*}{\textbf{AOE-Net}}  & & C3D & \textbf{77.67}  & \textbf{69.71} & \textbf{70.10} \\
& & 2Stream & 76.32 & 68.35 & 69.00 \\
& & SlowFast & 76.95 & 68.95 & 69.86 \\
%\textbf{AOE-Net} &             & 2Stream  & 76.32  & 68.35    & 69.00     \\
%             &             & SlowFast & \underline{\textit{76.95}}  & \textit{\underline{68.95}}    & \textit{\underline{69.86}}     \\
\bottomrule
\end{tabular}
}
\caption{\textbf{TAPG} comparisons on ActivityNet-1.3 \cite{caba2015activitynet} in terms of AR@100 and AUC on validation set and AUC on testing set.}
%\label{tab:features}
\label{tab:TAPG-ANet}
\end{table*}

%\vspace{2mm}
%\noindent
%\textbf{Implementation Details}
\subsection{Implementation Details}

To extract visual features from videos, we use a C3D \cite{C3D} network pre-trained on Kinetics-400 \cite{Kinetics} as the backbone network in all experiments on both ActivityNet-1.3 \cite{caba2015activitynet} and THUMOS-14 \cite{THUMOS14} (unless stated otherwise). The dimensions of the features extracted from the C3D backbone are 2048. 

In the objects beholder, to extract object text, we use CLIP \cite{radford2021learning} that was pre-trained on a large-scale dataset of 400M image-text pairs crawled from the Internet. The text feature and image feature are encoded by Transformer \cite{attention_is_all_you_need} and Vision Transformer \cite{dosovitskiy2020image} networks, respectively. In the actors beholder, to detect humans, we use a Faster-RCNN model \cite{FasterRCNN} that has been pre-trained on the COCO dataset \cite{cocodataset}. Adam optimizer was used to train our AOE-Net, and the initial learning rate is set to 0.0001 for ActivityNet-1.3 and 0.001 for THUMOS-14.

On ActivityNet-1.3, Soft-NMS (SNMS) \cite{SoftNMS} is used in post-processing for all experiments in TAPG and TAD. On THUMOS-14, following \cite{lin2018bsn, bmn}, both Soft-NMS \cite{SoftNMS} and NMS \cite{NMS} are utilized in post-processing of TAPG, whereas only NMS is applied in TAD. In the following experimental results, we emphasize the best performance in \textbf{bold} and the second-best performance in \underline{\textit{underline}}.

\begin{table}[!tb]
\centering
\resizebox{\linewidth}{!}{
\begin{tabular}{c|lll|ccccc|c}
& Methods & Venue \& Year & Feature & @50 & @100 & @200 & @500 & @1000 & Average \\
\hline \hline
\multirow{14}{*}{\rotatebox{90}{SNMS}}
%& TURN-TAP \cite{anchor_3} & CVPR18 & 18.55 & 29.00 & 39.61 & --    & --    \\
& CTAP  \cite{CTAP} & ECCV18 & 2Stream & 32.49 & 42.61 & 51.97 & -- & -- & -- \\
& BSN \cite{lin2018bsn} & ECCV18 & 2Stream & 37.46 & 46.06 & 53.21 & 60.64 & 64.52 & 52.38 \\
& MGG \cite{liu2019multi} & CVPR19 & I3D & 39.93 & 47.75 & 54.65 & 61.36 & 64.06 & 53.55 \\
& BMN \cite{bmn} & ICCV19 &  2Stream & 39.36 & 47.72 & 54.70 & 62.07 & 65.49 & 53.87 \\
& DBG \cite{dbg} & AAAI20 & 2Stream & 37.32 & 46.67 & 54.50 & 62.21 & 66.40 & 53.42 \\
& Rapnet \cite{gao2020accurate} & AAAI20 & C3D & 40.35 & 48.23 & 54.92 & 61.41 & -- & -- \\
& TSI++\cite{tsi_accv} & ACCV20 & 2Stream & 42.30 & \underline{\textit{50.51}} & 57.24 & 63.43 & -- & -- \\
& MR\cite{MR_eccv2020} & ECCV20 & I3D & 44.23 & \textbf{50.67} & 55.74 & -- & -- & -- \\
& BC-GNN \cite{bai2020boundary} & ECCV20 & 2Stream &  40.50 & 49.60 & 56.33 & 62.80 & -- & -- \\
& TCANet \cite{qing2021temporal} & CVPR21 & 2Stream & 42.05 & 50.48 & 57.13 & 63.61 & 66.88 & \underline{\textit{56.03}} \\
& SSTAP \cite{wang2021self} & CVPR21 & 2Stream & 41.01 & 50.12 & 56.69 & -- & \textbf{68.81} & -- \\
& ABN \cite{KhoaVo_Access} & IEEE-Access21 & C3D & 40.87 & 49.09 & 56.24 & 63.53 & 67.29 & 55.40 \\
& AEI-B \cite{khoavo_aei_bmvc21} & BMVC21 & C3D & \textbf{44.97} & 50.13 & \textbf{57.34} & \textbf{64.43} & 67.78 & \textbf{56.93} \\
\cline{2-10}
& \textbf{AOE} & & C3D & \underline{\textit{44.56}} & 50.26 & \underline{\textit{57.30}} & \underline{\textit{64.32}} & \underline{\textit{68.19}} & \textbf{56.93} \\
\hline \hline
\multirow{9}{*}{\rotatebox{90}{NMS}}
& BSN\cite{lin2018bsn} & ECCV18 & C3D & 27.19 & 35.38 & 43.61 & 53.77 & 59.50 & 43.89 \\
& BSN\cite{lin2018bsn} & ECCV18 & 2Stream & 35.41 & 43.55 & 52.23 & 61.35 & \underline{\textit{65.10}} & 51.53 \\
& BMN\cite{bmn} & ICCV19 & C3D & 29.04 & 37.72 & 46.79 & 56.07 & 60.96 & 46.12 \\
& BMN\cite{bmn} & ICCV19 & 2Stream & 37.15 & 46.75 & 54.84 & 62.19 & \textbf{65.22} & 53.23 \\
& DBG\cite{dbg} & AAAI20 & C3D & 32.55 & 41.07 & 48.83 & 57.58 & 59.55 & 47.92 \\
& DBG\cite{dbg} & AAAI20 & 2Stream & 40.89 & 49.24 & 55.76 & 61.43 & 61.95 & 53.85 \\
& ABN\cite{KhoaVo_Access} & IEEE-Access21 & C3D & \underline{\textit{44.89}} & 51.86 & 57.36 & 61.67 & 62.59 & 55.67 \\
& AEI-B \cite{khoavo_aei_bmvc21} & BMVC21 & C3D & \textbf{45.74} & \underline{\textit{52.39}} & \textbf{57.74} & \textbf{62.49} & 63.38 & \textbf{56.35} \\
\cline{2-10}
& \textbf{AOE} & & C3D & 44.78 & \textbf{52.41} & \underline{\textit{57.49}} & \underline{\textit{62.40}} & 63.40 & \underline{\textit{56.10}} \\
\bottomrule
\end{tabular}
}
\caption{\textbf{TAPG} comparisons on \textbf{THUMOS-14} in terms of AR@AN, where SNMS represents Soft-NMS \cite{SoftNMS}.} % \vspace*{-4mm}}
\label{tab:TAPG-Thumos}
\end{table}

%\vspace{2mm}
%\noindent
%\textbf{Performance and Comparison on TAPG}
\subsection{Performance and Comparison on TAPG}

Table \ref{tab:TAPG-ANet} presents TAPG comparison on both validation and testing sets of ActivityNet-1.3 \cite{caba2015activitynet}. The experimental results demonstrate that our approach AOE-Net with C3D~\cite{C3D} feature outperforms the existing methods in terms of AR@100 and AUC by an adequate margin. Table \ref{tab:TAPG-Thumos} shows the TAPG comparison on THUMOS-14. Compared to the existing TAPG methods, our AOE-Net performs very competitive on AR@ANs metrics with both SNMS and NMS. On SNMS, AOE-Net obtains the second best on all AR@ANs, except AR@100 where it is competitive to the best ones (50.26 vs. 50.67). On NMS, AOE-Net obtains the best on AR@100 and the second best on AR@200 and AR@500 with very close gap with the SOTA, 57.49 vs. 57.74 and 62.40 vs. 62.74, respectively. Notably, the performance on TAPG in both datasets of our AOE-Net are a very competitive with AEI-B \cite{khoavo_aei_bmvc21} and followed closely by ABN \cite{KhoaVo_Access}, both of which also incorporate local actors and global environment. This experiment strongly supports our observation and motivation on using the human perception principle to analyze human actions in untrimmed videos.

Beside solely evaluating AOE-Net on TAPG and TAD tasks, the effects of different backbone features to our AOE-Net also worth an investigation. 
The performance of our proposed AOE-Net network on different features, i.e., C3D~\cite{C3D}, 2Stream~\cite{2_stream_1} and Slowfast~\cite{SlowFast}, with the features dimensions are 2048, 2314 and 400, respectively, are reported in the bottom part of Table~\ref{tab:TAPG-ANet} on TAPG task of ActivityNet-1.3 dataset~\cite{caba2015activitynet}. As demonstrated, we notice that the performance with C3D~\cite{C3D} features are state-of-the-art, while the performance with SlowFast~\cite{SlowFast} features are closely behind. Whereas, the performance with 2Stream~\cite{2_stream_1} features is the worst in three types of backbone features.

\begin{figure*}[!tb]
 \begin{minipage}[b]{\linewidth}
 \centering
 \resizebox{0.9\linewidth}{!}{
    \begin{tabular}{l|l|ll|ll}
    \multirow{3}{*}{Methods} &  & \multicolumn{4}{c}{Evaluation}\\ 
    & & \multicolumn{2}{c|}{Seen} & \multicolumn{2}{c}{Unseen} \\
    & Training  & AR@100       & AUC       & AR@100        & AUC \\ \hline \hline
    \multirow{2}{*}{BSN  \cite{lin2018bsn}} & Seen + Unseen & 72.40 & 63.80 & 71.84 & 63.99 \\
    & Seen &  72.42 & 64.02 & 71.32 & 63.38 \\ \hline
    \multirow{2}{*}{\shortstack{BMN \cite{bmn}}}
    & Seen + Unseen   & 72.96 & 65.02 & 72.68 & 65.05 \\ 
    & Seen          & 72.47 & 64.37 & 72.46 & 64.47 \\
    \hline
    \multirow{2}{*}{\shortstack{TSI++ \cite{tsi_accv}}}
    & Seen + Unseen & 74.69 & 66.54 & 74.31 & 66.14 \\ 
    & Seen        & 73.59 & 65.60 & 73.07 & 65.05 \\ \hline
    \multirow{2}{*}{\shortstack{DBG \cite{dbg}}}
    & Seen + Unseen& 73.30 &  66.57&  67.23 &  64.59\\ 
    & Seen        & 72.95 &  66.23 &  64.77 &  62.18 \\  \hline
    \multirow{2}{*}{\shortstack{ABN \cite{KhoaVo_Access}}}
    & Seen + Unseen& 74.58 & 66.96 & 75.25 & 67.49 \\ 
    & Seen        &  74.40 & 66.69 & 73.66 & 65.49 \\ \hline \hline
    \multirow{2}{*}{\textbf{AOE-Net}} & Seen + Unseen & 76.36 &  68.31 & 77.31 & 69.07 \\
    & Seen & 76.43 & 68.42 & 74.90 & 66.92 \\
    \bottomrule
    \end{tabular}
 }
 \end{minipage} \\
 \begin{minipage}[b]{\linewidth}
  \centering
  \includegraphics[width=\textwidth]{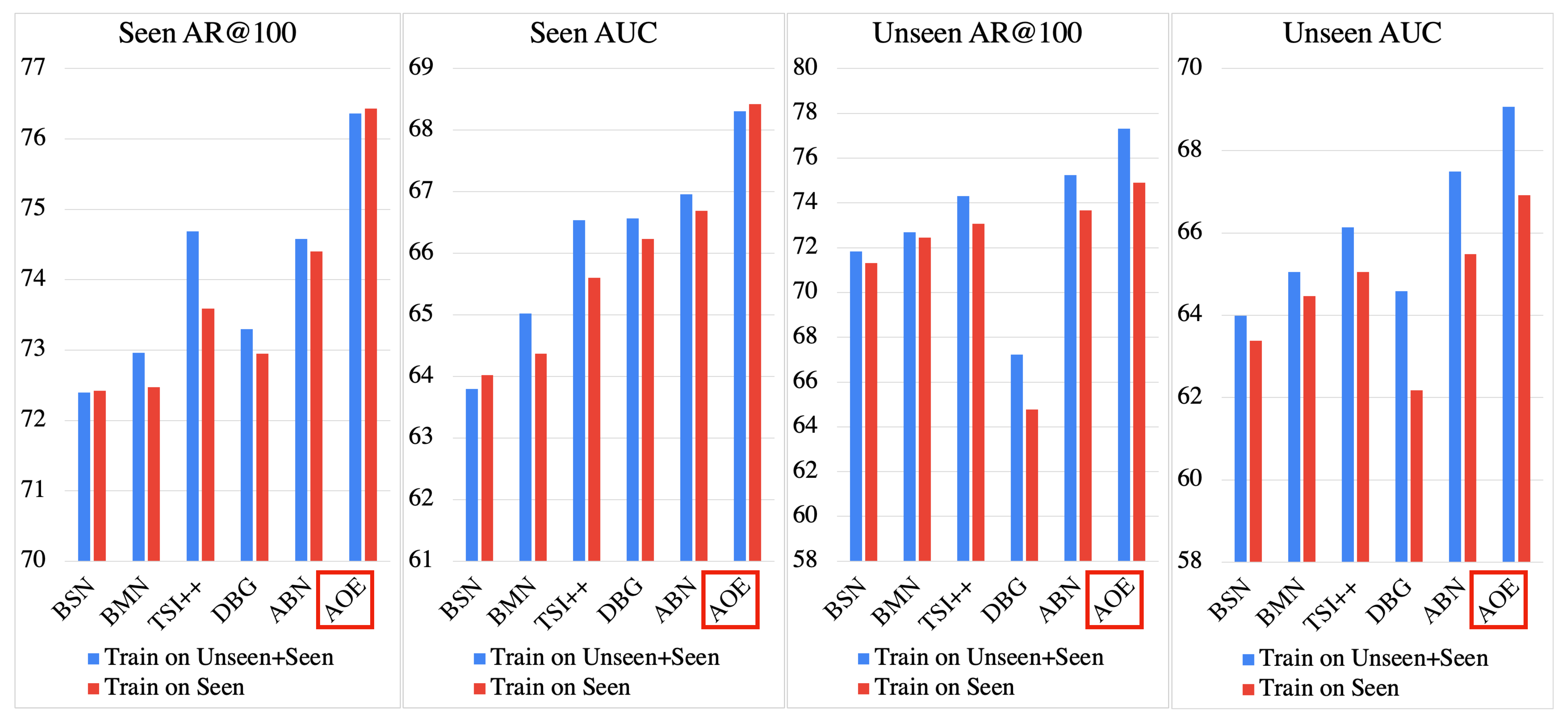}
%  \caption{\textbf{Generalizability} evaluation and comparisons on ActivityNet-1.3 in terms of AR@100 and AUC. Methods are trained on \textcolor{blue}{\textit{Unseen+Seen}} and \textcolor{red}{\textit{Seen}} training sets, respectively; and are evaluated on \textit{Seen} (first two charts) and \textit{Unseen} (last two charts) validation sets.} % Top: Detailed performance of individual experiment setting of various methods. Bottom: Visualized generalizability comparison between our proposed AOE-Net and other methods } % \vspace*{-3mm}}
%  \vspace{-2mm}
 \end{minipage}
\caption{\textbf{Generalizability} evaluation and comparisons on ActivityNet-1.3 in terms of AR@100 and AUC. Methods are trained on \textcolor{blue}{\textit{Unseen+Seen}} and \textcolor{red}{\textit{Seen}} training sets, respectively; and are evaluated on \textit{Seen} (first two charts) and \textit{Unseen} (last two charts) validation sets. Top: Detailed performance of individual experiment setting of various methods. Bottom: Visualized generalizability comparison between our proposed AOE-Net and other methods.} % \vspace*{-3mm}}
\label{fig:Generalizability}
\end{figure*}

In TAPG, \emph{generalizability} is also a significant criterion to evaluate a method. Following the same experiment setup in \cite{lin2018bsn, bmn, dbg, tsi_accv, KhoaVo_Access}, we conduct this study on ActivityNet-1.3 with two subsets, i.e., \textit{Seen}: "Sports, Exercises, and Recreation" and \textit{Unseen}: "Socializing, Relaxing, and Leisure". Our AOE-Net is trained on \textit{Unseen+Seen} and \textit{Seen} training sets, separately, and then evaluated on the \textit{Seen} and \textit{Unseen} validation sets. Fig.~\ref{fig:Generalizability} provides the performance comparison and visualization between AOE-Net with other SOTA methods. In each chart on the right, the performance of AOE-Net is shown in the last columns, which demonstrates that AOE-Net is superior to other SOTA methods. Fig.~\ref{fig:Generalizability} also shows that our AOE-Net achieves good performances on \textit{Seen} validation set with an acceptable drop on \textit{Unseen} validation set on both training configurations, suggesting that our AOE-Net is highly generalizable to unseen action types.

% \usepackage{multirow}
%\begin{table}[]
%\centering
%\begin{tabular}{l|l|ll|ll}
%\multirow{3}{*}{Methods} &  & \multicolumn{4}{c}{Evaluation}\\ 
%& & \multicolumn{2}{c|}{Seen} & \multicolumn{2}{c}{Unseen} \\
%& Training  & AR@100       & AUC       & AR@100        & AUC \\ \hline \hline
%\multirow{2}{*}{BSN  \cite{lin2018bsn}} & Seen + Unseen & 72.40 & 63.80 & 71.84 & 63.99 \\
%& Seen &  72.42 & 64.02 & 71.32 & 63.38 \\ \hline
%\multirow{2}{*}{\shortstack{BMN \cite{bmn}}}
%& Seen + Unseen   & 72.96 & 65.02 & 72.68 & 65.05 \\ 
%& Seen          & 72.47 & 64.37 & 72.46 & 64.47 \\
%\hline
%\multirow{2}{*}{\shortstack{TSI++ \cite{tsi_accv}}}
%& Seen + Unseen & 74.69 & 66.54 & 74.31 & 66.14 \\ 
%& Seen        & 73.59 & 65.60 & 73.07 & 65.05 \\ \hline
%\multirow{2}{*}{\shortstack{DBG \cite{dbg}}}
%& Seen + Unseen& 73.30 &  66.57&  67.23 &  64.59\\ 
%& Seen        & 72.95 &  66.23 &  64.77 &  62.18 \\  \hline
%\multirow{2}{*}{\shortstack{ABN \cite{KhoaVo_Access}}}
%& Seen + Unseen& 74.58 & 66.96 & 75.25 & 67.49 \\ 
%& Seen        &  74.40 & 66.69 & 73.66 & 65.49 \\ \hline \hline
%\multirow{2}{*}{\textbf{AOE-Net}} & Seen + Unseen & 76.36 &  68.31 & 77.31 & 69.07 \\
%& Seen & 76.43 & 68.42 & 74.90 & 66.92 \\
%\bottomrule
%\end{tabular}
%\caption{\hl{Add to Fig.6}}
%\end{table}

\begin{table*}[!tb]
\centering
\resizebox{0.9\linewidth}{!}{
\begin{tabular}{lll|ccc|c}
Methods & Venue \& Year & Feature & 0.50 & 0.75 & 0.95 & Average \\
\hline \hline
%SS-TAD \cite{SSTAD_BMVC17} & BMVC17 & & 44.39 & 29.65 & 7.09 & 29.20 \\
%SSN \cite{SSN} & & 39.12 & 23.48 & 5.49 & 23.98 \\
BSN \cite{lin2018bsn} & ECCV18 & 2Stream & 46.45 & 29.96 & 8.02 & 30.03 \\
GTAN \cite{gtan_cvpr2019} & CVPR19 & P3D & \underline{\textit{52.61}} & 34.14 & 8.91 & 34.31 \\
BMN \cite{bmn} & ICCV19 & 2Stream & 50.07 & 34.60 & 8.29 & 33.85 \\
GTAD \cite{xu2020gtad} & CVPR20 & 2Stream & 50.36 & 34.60 & 9.02 & 34.09 \\
P-GCN \cite{pgcn_cvpr2020} & CVPR20 & I3D & 42.90 & 28.14 & 2.47 & 26.99 \\
MR \cite{MR_eccv2020} & ECCV20 & 2Stream & 43.47 & 33.91 & 9.21 & 30.12 \\
TSI++ \cite{tsi_accv} & ACCV20 & 2Stream & 51.18 & \underline{\textit{35.00}} & 6.59 & 34.15 \\
BC-GNN \cite{bai2020boundary} & ECCV20 & 2Stream & 50.56 & 34.75 & 9.37 & 34.26 \\
RTD \cite{tan2021relaxed} & ICCV21 & 2Stream & 47.21 & 30.68 & 8.61 & 30.83 \\
ABN \cite{KhoaVo_Access} & IEEE-Access21 & C3D & 51.78 & 34.18 & \underline{\textit{10.29}} & 34.22 \\
AEI-B \cite{khoavo_aei_bmvc21} & BMVC21 & C3D & 52.3 & 34.5 & 9.7 & \textbf{34.7} \\
%MUSeS \cite{liu2019multi} & CVPR19 & 50.02 & 34.97 & 6.57 & 33.99 \\
\hline
\textbf{AOE} & & C3D & \textbf{54.42} & \textbf{35.43} & \textbf{10.35} & \underline{\textit{34.48}} \\

%& \cite{SSTAD_BMVC17} & \cite{SSN}  & \cite{lin2018bsn} & \cite{bmn} & \cite{xu2020gtad} & \cite{pgcn_cvpr2020} & \cite{MR_eccv2020} & \cite{KhoaVo_Access} & \cite{tsi_accv} & \cite{gtan_cvpr2019} &  \cite{bai2020boundary} & \cite{tan2021relaxed} & \cite{liu2021multi} & \\
\bottomrule
\end{tabular}
}
%\vspace{-2mm}
\caption{\textbf{TAD} comparisons on \textbf{ActivityNet-1.3} in terms of mAP@tIoU and mAP, where the proposals are combined with video-level classification results generated by \cite{action_protocol}.}
\label{tb:AOE-TAD-ANet}
\end{table*}

%\vspace{2mm}
%\noindent
%\textbf{Performance and Comparison on TAD}
\subsection{Performance and Comparison on TAD}

For a fair comparison, we follow the experiment settings in \cite{lin2018bsn, bmn, dbg, xu2020gtad, bai2020boundary, liu2019multi, tan2021relaxed, KhoaVo_Access} to produce labels for action proposals produced by our AOE-Net. On AcitivityNet-1.3, we adopt the top-1 video-level classification results generated by the method in \cite{action_protocol} for our proposals. Whereas on THUMOS-14, we instead label our action proposals with either UntrimmedNet \cite{untrimmetNet} (top-2 classification results) or P-GCN \cite{pgcn_cvpr2020}.

%Following the experiment settings in \cite{lin2018bsn, bmn, dbg, xu2020gtad, bai2020boundary, liu2019multi, tan2021relaxed, KhoaVo_Access}, we adopt the video-level classification results generated by the method in \cite{action_protocol} on ActivityNet-1.3 to label the proposals produced by our method. We use top-2 video-level classification results generated by UntrimmedNet \cite{untrimmetNet} to label proposals generated by AOE-Net on THUMOS-14.

Table \ref{tb:AOE-TAD-ANet} shows TAD performance comparison between AOE-Net and other SOTA methods on ActivityNet-1.3 validation set. The results emphasize that our method outperforms SOTA methods on multiple tIoU thresholds. The experiment results on THUMOS-14 test set in Table \ref{tb:AOE-TAD-Thumos} demonstrate that our AOE-Net is superior to other SOTA methods on most of the metrics with both classifiers. % In this experiment, UntrimmedNet \cite{untrimmetNet} is utilized as a classifier. 

\begin{table*}[!t]
\centering
        %The first group uses UntrimmedNet as a classifier whereas the second group use G-PCN as a classifier}
\resizebox{\linewidth}{!}{
        \begin{tabular}{l l l l| l l l l l |l} 
    &    Methods & Year & Feature  & 0.7   & 0.6  & 0.5   & 0.4   & 0.3  & Average \\ \hline
        \hline
    \multirow{14}{*}{\rotatebox{90}{UntrimmedNet \cite{untrimmetNet}}} 
    &    BSN \cite{lin2018bsn}   & ECCV18 &  2Stream  & 20.0 & 28.4 & 36.9 & 45.0 & 53.5 & 36.76\\
    &    BMN \cite{bmn}         &ICCV19 &  2Stream   & 20.5 & 29.7 & 38.8 & 47.4 & 56.0 & 38.48\\
    &    MGG \cite{liu2019multi}   & CVPR19 & 2Stream & 21.3 & 29.5 & 37.4 & 46.8 & 53.9 & 37.78\\
    &    GTAN~\cite{gtan_cvpr2019} & CVPR19 & P3D & -- & -- & 38.8 & 47.2 & 57.8 & -- \\
    &    DBG \cite{dbg}          & AAAI20 &  2Stream  & 21.7 & 30.2 & 39.8 & 49.4 & 57.8 & 39.78\\
    &    GTAD \cite{xu2020gtad}  & CVPR20 & 2Stream   & 23.4 & 30.8 & 40.2 & 47.6 & 54.5 & 39.30\\  
    &    TSI++\cite{tsi_accv}    & ACCV20 & 2Stream   & 22.4 & 33.2 & 42.6 & 52.1 & \underline{\textit{61.0}} & 42.26\\
    &    BC-GNN~\cite{bai2020boundary} & ECCV20 & 2Stream & 23.1 & 31.2 &  40.4 & 49.1 & 57.1 & 40.18\\
    &    BU-TAL\cite{zhao2020bottom} & ECCV20 & 2Stream & \textbf{28.5} & \underline{\textit{38.0}} & 45.4  & 50.7 & 53.9 & 43.30\\ 
    & TCANet \cite{qing2021temporal} & CVPR21 & 2Stream & 26.7 & 36.8 & 44.6 & 53.2 & 60.6 & 44.38\\
    & RTD\cite{tan2021relaxed} & ICCV21 & 2Stream & 25.0 & 36.4 & 45.1 & 53.1 & 58.5 & 43.62\\
    & ABN \cite{KhoaVo_Access} & IEEE-Access21 & C3D & 25.6 & 37.0 & \underline{\textit{46.1}} & \underline{\textit{54.0}} & 59.9 &  44.51 \\
    & AEI-B \cite{khoavo_aei_bmvc21} & BMVC21 & C3D & 23.4 & 35.9 & 44.7 & 52.7 & 58.7 & 43.08 \\
        \cline{2-10}
    &    \textbf{AOE-Net}  & --& C3D & \underline{\textit{25.8}} & \textbf{38.8} & \textbf{48.4} & \textbf{57.3} &  \textbf{63.4} & \textbf{46.74}\\
    \hline \hline
    % threshold 0.4, 3000 proposals
     \multirow{4}{*}{\rotatebox{90}{P-GCN \cite{pgcn_cvpr2020}}} 
    &    BSN\cite{lin2018bsn}  & ECCV18 &  I3D & --   & --   & 49.1 & 57.8 & 63.6  & -- \\
    &    MR\cite{MR_eccv2020} & ECCV20 & 2Stream  & -- & -- & 50.10 & 60.99 & 66.29 & -- \\
    &    GTAD \cite{xu2020gtad} & CVPR20 & 2Stream & 22.9 & 37.6 & \textbf{51.6} & \underline{\textit{60.4}} & \underline{\textit{66.4}} & \underline{\textit{47.78}} \\
    %& AEI-B \cite{khoavo_aei_bmvc21} & BMVC21 & C3D & 22.4 & 37.8 & 52.1 & 60.6 & 67.3 & 48.04 \\
    \cline{2-10}
    &    \textbf{AOE-Net} &-- & C3D & \textbf{23.5} & 37.4 & \underline{\textit{50.9}} & \textbf{60.6} & \textbf{67.1} & \textbf{47.89} \\ % threshold 0.4, 3000 proposals
        \bottomrule
        \end{tabular}
}
%\vspace{-2mm}
\caption{\textbf{TAD} comparisons on \textbf{THUMOS-14} in term of mAP@tIoU using two different classifiers, i.e., UntrimmedNet \cite{untrimmetNet} and P-GCN \cite{pgcn_cvpr2020}.}
\label{tb:AOE-TAD-Thumos}
\end{table*}

\begin{table}[!t]
%\addtolength{\tabcolsep}{-4pt} 
\centering
\resizebox{\linewidth}{!}{
\begin{tabular}{l|ccccc|c|c|c|c|c}
\multirow{2}{*}{\textbf{Exp}} & \multicolumn{5}{c|}{\textbf{Setting}} & \multicolumn{5}{c}{\textbf{TAPG Performance}}\\
& Act. & Env. & Obj. & AAM & Soft-Att & @50 & @100 & @200 & @500 & @1000 \\ \hline \hline
\#1 & \colorbox{cyan}{\cmark} & \xmark & \xmark & \xmark &\colorbox{cyan}{\cmark} & 25.96 & 35.14 & 43.48 & 52.37 & 58.47 \\
\#2& \xmark & \colorbox{cyan}{\cmark} & \xmark & \xmark & \xmark & 38.94 & 47.80 & 54.93 & 61.92 & 65.96 \\
\#3& \xmark & \xmark & \colorbox{cyan}{\cmark} & \xmark &\colorbox{cyan}{\cmark} & 18.06 & 26.68 & 37.14 & 49.28 & 56.99 \\ \hline
\#4& \colorbox{cyan}{\cmark} & \colorbox{cyan}{\cmark} & \xmark & \xmark &\colorbox{cyan}{\cmark} & 40.87 & 49.09 & 56.24 & 63.53 & 67.29 \\
\#5& \colorbox{cyan}{\cmark} & \colorbox{cyan}{\cmark} & \colorbox{cyan}{\cmark} & \xmark &\colorbox{cyan}{\cmark} & 42.60 & 49.86 & 56.87 & 63.76 & 67.60 \\
\#6& \colorbox{cyan}{\cmark} & \colorbox{cyan}{\cmark} & \xmark & \colorbox{cyan}{\cmark} & \xmark & 43.79 & 49.67 & 56.73 & 63.49 & 67.36 \\ \hline
\#7& \colorbox{cyan}{\cmark} & \colorbox{cyan}{\cmark} & \colorbox{cyan}{\cmark} & \colorbox{cyan}{\cmark} & \xmark & 44.56 & 50.26 & 57.30 & 64.32 & 68.19 \\ \bottomrule
\end{tabular}
}
%\vspace{-2mm}
\caption{TAPG comparisons on different network settings. Act., Env., Obj. denote actors, environment, objects beholders.} % \vspace*{-3mm}}
\label{tb:abl_1}
\end{table}

\begin{table}[!t]
\centering
%\resizebox{0.85\linewidth}{!}{
\addtolength{\tabcolsep}{-3pt} 
\begin{tabular}{l|ccccc|ccc}
\multirow{2}{*}{\textbf{Attention}} & \multicolumn{5}{c|}{\textbf{THUMOS-14}} & \multicolumn{3}{c}{\textbf{ActivityNet-1.3}} \\ \cline{2-9}
  & {@50} & {@100} & {@200} & {@500} & {@1000}  & AR @100& AUC (val)& AUC (test)\\ \hline \hline
      Hard \cite{adahard_eccv2018} & 43.74 & 49.24 & 56.63 & 63.46 & 67.25 & 77.11 & 69.02 & 69.56 \\ % AUC Test is actually 69.76
      Soft \cite{attention_is_all_you_need} & 42.60 & 49.86 & 56.87 & 63.76 & 67.60  & 76.93 & 69.06 & 69.23 \\ \hline
      \textbf{AAM}  & \textbf{44.56} & \textbf{50.26} & \textbf{57.30} & \textbf{64.32} & \textbf{68.19} & \textbf{77.67} & \textbf{69.71} & \textbf{70.10} \\ \bottomrule
\end{tabular}
%}
%\vspace{-2mm}
\caption{TAPG compare between AAM with attention\cite{adahard_eccv2018, attention_is_all_you_need}.} % \vspace*{-3mm}}
\label{tb:abl_2}
\end{table}

\begin{table}[!t]
\centering
%\resizebox{0.75\linewidth}{!}{
\begin{tabular}{l|lll}
      & AR@10 & AR@100 & AUC \\ \hline \hline
      BMN\cite{bmn} & 11.59 & 34.26 & 25.14 \\
      \textbf{AOE-Net}  & \textbf{15.99} & \textbf{37.40} & \textbf{29.20} \\ \bottomrule
\end{tabular}
%}
%\vspace{-2mm}
\caption{TAPG comparison between our AOE-Net with BMN \cite{bmn} on egocentric videos~\cite{damen2021rescaling}.} % \vspace*{-3mm}}
\label{tb:abl_3}
\end{table}

%\vspace{2mm}
%\noindent
%\textbf{Ablation Study}
\subsection{Ablation Study}

We further conduct a rich ablation study to show the effectiveness of each component in the proposed AOE-Net as well as the robustness of AOE-Net to egocentric videos. We also report the network efficiency and AOE-Net performance with different settings of hyper-parameter K. Additional ablation study will be included in supplementary.

%\noindent
\subsubsection{Contribution of each beholder}
We examine TAPG performance on THUMOS-14 with different network settings as given in Table \ref{tb:abl_1}. While the performance of each individual beholder 
%itself 
is shown in the Exps.\#1-3, different combinations of features are given in Exps.\#4-7. This emphasizes the important contribution of actors and objects in understanding human action. Comparisons between Exps.(\#4 vs.~\#6) and (\#5 vs.~\#7) highlight the strong impact of AAM.

In Exp.\#1 and Exp.\#3, as Environment Beholder is not presented, AAM consequently cannot be applied because it requires environment feature as one of its input. Therefore, we replace AAM by a simple soft self-attention layer followed by an average pooling operation to fuse multiple actors together.
Likewise, in Exp.\#4 and Exp.\#5 we also perform the above replacement strategy to emphasize the effectiveness of AAM.

%
%\noindent
\subsubsection{Effectiveness of AAM}
We continue studying the effectiveness of the proposed AAM in TAPG task on both ActivityNet-1.3 and THUMOS-14 by comparing AAM with different attention mechanisms, i.e., soft self-attention \cite{attention_is_all_you_need} (Soft), hard attention \cite{adahard_eccv2018} (Hard) as shown in Table.\ref{tb:abl_2}.

For soft self-attention mechanism, we simply remove the actors hard attention part at the beginning of our AAM, which is defined in Eq.~\ref{eq:env_mlp}-\ref{eq:act_select}, and directly feed the input set of actors features $\mathcal{F}^a$ (or objects features $\mathcal{F}^o$) into a self-attention mechanism.

In contrast, for hard-attention mechanism, we replace the self-attention part at the end of our AAM by a simple average pooling operation to average the selected actors features $\tilde{F}^a$ (or selected objects features $\tilde{F}^o$) into a single representation $f^a$ (or $f^o$).

With the higher performances on both datasets shown in Table \ref{tb:abl_2}, AAM proves its appealing advantages over soft self-attention and hard attention mechanisms.

\subsubsection{Performance of AOE-Net with different number of objects}
\label{sec:topK}
The number of input objects $K$ for objects beholder (Sub sec. \ref{subsubsec:objects}) is also a hyper-parameter that may affect the performance of our AOE-Net. If we use a large $K$, the overall model may receive more noisy information due to the increasingly incorrect detected objects. Whereas, if we use a small $K$, the objects beholder will not present enough significant information with a few objects. Thus, the contribution of objects beholder to understand actions is insufficient.

In this ablation study, we benchmark our AOE-Net on various number of objects $K$ with TAPG task and ActivityNet-1.3 \cite{caba2015activitynet} dataset. The comparison is reported in Table \ref{tb:num_objects}.

Table \ref{tb:num_objects} shows that as we increase $K$, the TAPG performance of AOE-Net also tend to improve. However, when $K > 20$, the performance starts fluctuating and is not robust due to more wrongly detected objects in each snippet. Therefore, we conclude that $K=20$ gives the best trade-off between performance and robustness.

\begin{table}[!t]
\centering
%\resizebox{0.75\linewidth}{!}{
\begin{tabular}{c|lll}
Number of Objects (K) & AR@100 & AUC (val) & AUC (test) \\
\hline \hline
0  & 77.02 & 68.98 & 69.72 \\
1  & 77.15 & 69.17 & 69.95 \\
5  & 77.45 & 69.43 & 69.96 \\
10 & 77.24 & 69.26 & 69.56 \\
20 & \textbf{77.67} & \underline{\textit{69.71}} & \underline{\textit{70.10}} \\
30 & 77.55 & 69.63 & 69.96 \\ %69.78 \\
40 & \textbf{77.67} & \textbf{69.86} & \textbf{70.22} \\
50 & 77.24 & 69.17 & 69.81 \\
\bottomrule
\end{tabular}
%}
%\vspace{-2mm}
\caption{TAPG performance of our AOE-Net on ActivityNet-1.3 \cite{caba2015activitynet} with various settings of K} % \vspace*{-3mm}}
\label{tb:num_objects}
\end{table}

%\noindent
\subsubsection{Robustness of AOE-Net to egocentric videos}
To benchmark the robustness of AOE-Net on egocentric videos, we use EPIC-KITCHENS 100 \cite{damen2021rescaling} to benchmark TAPG task. Table \ref{tb:abl_3} provides the TAPG comparison between our AOE-Net with BMN \cite{bmn}. Even actors are not shown in the egocentric videos, our AOE still obtains good TAPG performance with a big improvement compare to BMN. This proves the effectiveness of the objects beholder.
%Because in this dataset, the main actor is always behind camera view, we remove the actors beholder of our AOE-Net. 

%\noindent
\subsubsection{Network efficiency}
%Table~\ref{tb:abl_4} reports AOE-Net efficiency with model size, \#params, FLOPs, inference time on a 3-minute video with either an Intel Xeon CPU or a single NVIDIA V100 machine.
Table~\ref{tb:abl_4} reports the efficiency of AOE-Net and previous SOTA with \#parameters in millions (M) , computational cost (GFLOPs), inference time on a 3-minute video with either an Intel Core i9-9920X CPU or a single NVIDIA RTX 2080 Ti.

\begin{table}[!t]
\centering
%\resizebox{0.8\linewidth}{!}{
%\begin{tabular}{l|ccccc}
%      & Model size & \#Params& FLOPs & \multicolumn{2}{c}{Inference time (s)} \\ 
%      &  (MB) & (M) & (G) & GPU & CPU\\ 
%      \hline \hline
%      BMN~\cite{bmn} & 141 & 4.9 & 91.22 & 0.128 & 4.15 \\ %This will be in supp.
%      DBG~\cite{bmn} & 11 & 2.9 & 47.52 & 0.03 & - \\ %This will be in supp.
%      GTAD~\cite{xu2020gtad} & 22 & 5.62 & 150.28 &  &  \\ %This will be in supp.
%      \textbf{AOE-Net} & 163 & 8.8 & 94.02 0.03 & 0.21 \\ \bottomrule
%\end{tabular}
\begin{tabular}{l|ccccc}
      & \#Params& FLOPs & \multicolumn{2}{c}{Inference time (s)} \\ 
      & (M) & (G) & GPU & CPU\\ 
      \hline \hline
      %BMN~\cite{bmn} & 4.9 & 91.22 & 0.128 & 4.15 \\ %This will be in supp.
      BMN~\cite{bmn} & 4.9 & 71.22 & 0.128 & 4.15 \\ %This will be in supp.
      DBG~\cite{dbg} & 2.9 & 47.52 & 0.03 & - \\ %This will be in supp.
      GTAD~\cite{xu2020gtad} & 5.6 & 150.28 & 0.14 & - \\ %This will be in supp.
      ABN~\cite{KhoaVo_Access} & 6.9 & 87.88 & 0.07 & 0.21\\ %This will be in supp.
      AEI~\cite{khoavo_aei_bmvc21} & 6.9 & 90.62 & 0.08 & 0.21 \\ %This will be in supp.
      \hline
      \textbf{AOE-Net} & 8.8 & 94.02 & 0.12 & 0.27 \\ \bottomrule
\end{tabular}
%}
% \vspace{-2mm}
\caption{Network efficiencies of AOE-Net and several of previous works.} %\vspace*{-4mm}}
%\hl{Need to investigate more on model size, time consumption of feature extraction (env., agent, object), PMR, BMM}}
\label{tb:abl_4}
\end{table}

%\begin{table}[!t]
%\centering
%\begin{tabular}{l|lll}
% Backbone & AR@100 & AUC (val) & AUC (test) \\
%\hline \hline
%C3D & \textbf{77.67} & \textbf{69.71} & \textbf{70.10} \\
%2Stream & 76.32 & 68.35 & 69.00 \\
%SlowFast & \underline{\textit{76.95}} & \textit{\underline{68.95}} & \textit{\underline{69.86}} \\
%\bottomrule
%\end{tabular}
%%\vspace{-2mm}
%\caption{TAPG Performance of our AOE-Net on ActivityNet-1.3 \cite{caba2015activitynet} with different types of backbone feature.} % \vspace*{-3mm}}
%\label{tb:features}
%\end{table}

%\subsubsection{Performance of AOE-Net with various backbone features}
%Beside solely evaluating AOE-Net on TAPG and TAD tasks, it is also worth investigating the effects of different backbone features to our AOE-Net.
%
%The performance of our proposed AOE-Net network on different features, i.e., C3D~\cite{C3D}, 2Stream~\cite{2_stream_1} and Slowfast~\cite{SlowFast}, with the features dimensions are 2048, 2314 and 400, respectively. The comparions are reported in Table~\ref{tb:features} on TAPG task of ActivityNet-1.3 dataset~\cite{caba2015activitynet}. As demonstrated, we notice that the performance with C3D~\cite{C3D} features are state-of-the-art, while the performance with SlowFast~\cite{SlowFast} features are closely behind. Whereas, the performance with 2Stream~\cite{2_stream_1} features is the worst in three types of backbone features.
%
%
\subsection{Qualitative Analysis of AAM}

\begin{figure*}[!ht]
    \centering
    \includegraphics[width=0.95\linewidth]{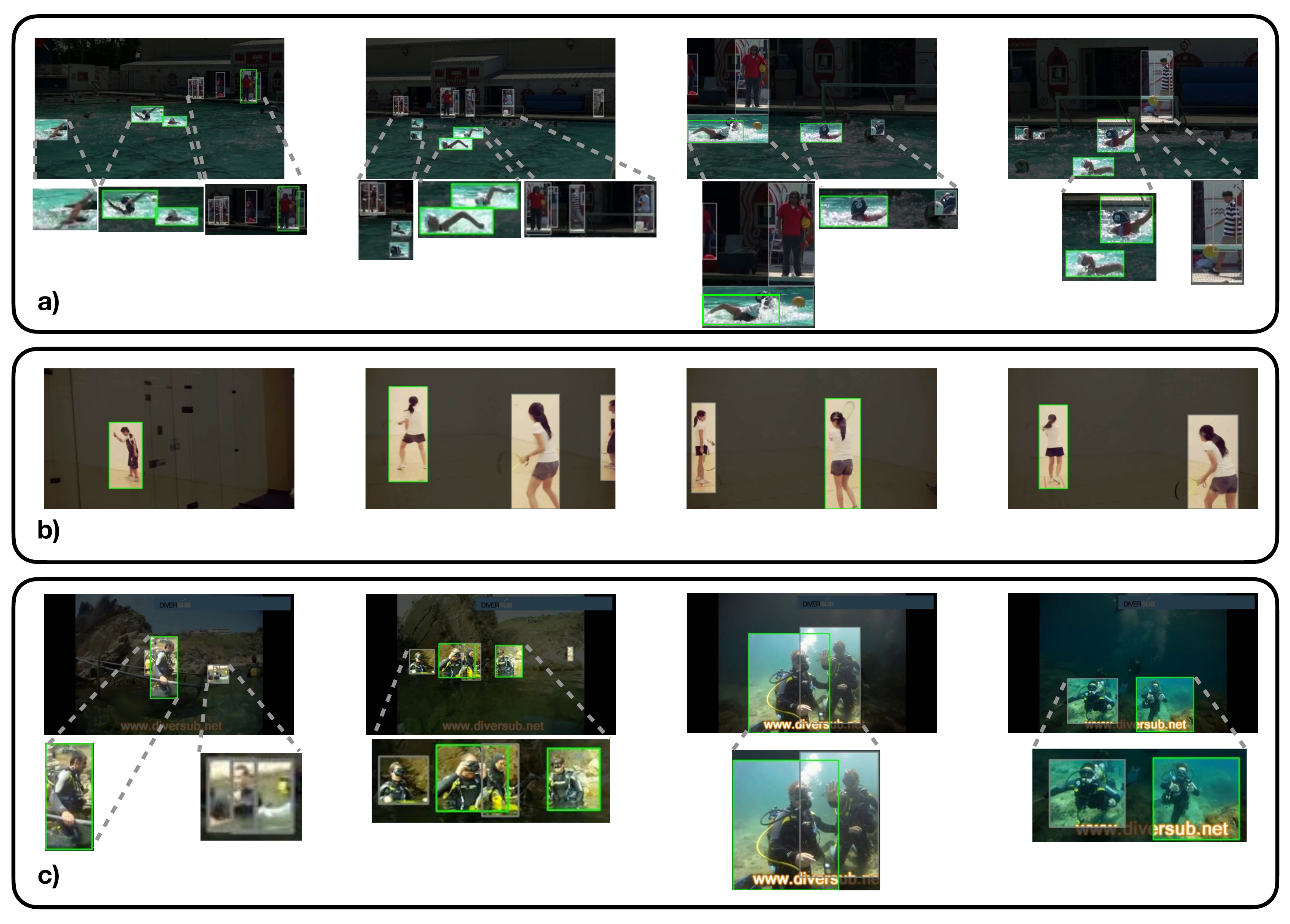}
    \caption{Visualization of main actors selection resulting by AAM on ActivityNet-1.3 \cite{caba2015activitynet}. (a), (b), and (c) are three different videos. The background is blacked out, the bounding boxes of main actors are outlined by green line and the bounding boxes of insignificant actors are outlined by grey line.}
    \label{fig:AAM_qualitative}
\end{figure*}

\noindent\textbf{Qualitative Results of AAM with Actors Beholder:}\\
Fig.~\ref{fig:AAM_qualitative} shows qualitative performances of AAM in selecting main actors in the set of detected ones. The videos are retrieved from ActivityNet-1.3 \cite{caba2015activitynet}. 
In the case of multiple actors detected (Fig.~\ref{fig:AAM_qualitative}(a) and Fig.~\ref{fig:AAM_qualitative}(c)), our proposed AAM can effectively select main actors in the scene and remove the insignificant actors. This aims to eliminate redundant information as well as select the most relevant information to feed into the boundary-matching module. Fig.~\ref{fig:AAM_qualitative}(b) illustrates the scenario where the environment is tedious and may not contribute to perceive the action. However, the local information at the bounding box around the main actor can help highlight the action. In Fig.~\ref{fig:AAM_qualitative}(b), AAM one again shows its merit when selecting the main actor who actually commits the action.

\vspace{2mm}
\noindent\textbf{Performance of AAM affected by human detector:}\\
The human detector we used is Faster-RCNN \cite{FasterRCNN} trained on COCO dataset \cite{cocodataset}. In practice, the human detector is not completely perfect in the videos due to motion blurs or low resolutions. Therefore, the AAM is also affected by the quality of detected human bounding boxes.

\begin{figure}[!t]
    \centering
    \includegraphics[width=0.7\linewidth]{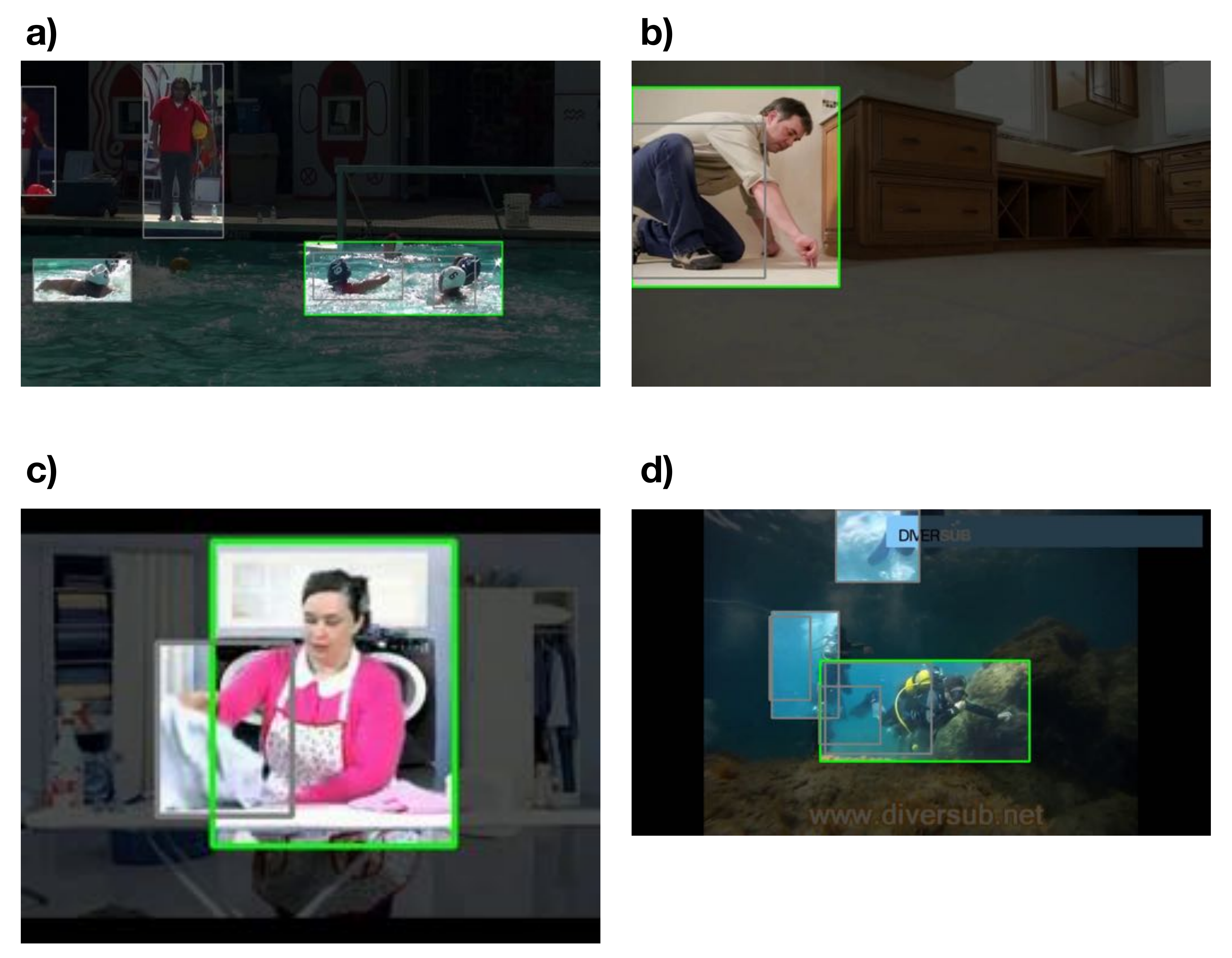}
    \caption{Visualization of AAM on ActivityNet-1.3 \cite{caba2015activitynet} on cases where the human detector poorly generates human bounding boxes. The background is blacked out, the bounding boxes of main actors are outlined by green line and the bounding boxes of insignificant actors are outlined by grey line.}
    \label{fig:AAM_bad}
\end{figure}

In Fig.~\ref{fig:AAM_bad}, we visualize frames of 4 videos where the human detector poorly produces human bounding boxes. In Fig.~\ref{fig:AAM_bad}(a), the green bounding box is localized around two athletes in the pool beside two separate bounding boxes for each athlete. Although the green bounding box is incorrect because it contains two humans (even three if we count the one behind), it is intuitively better than the individual boxes of each athlete because it contains richer information of the scene. This proves that our AAM effectively learnt to select the best bounding boxes detected regardless of its quality in terms of human detection.

In Fig.~\ref{fig:AAM_bad}(b), (c) and (d), we notice that there are badly detected bounding boxes which is just a body part of the humans instead of a whole. However, AAM could learn to eliminate these bad bounding boxes to only select the correct ones.

From the above observations, we can see that AAM can learn to avoid selecting bad bounding boxes which do not fully contain the humans. Oppositely, AAM also learns to select some badly detected bounding boxes but contains more meaningful information than the correct ones.

However, we conclude that relying on the human detector in providing locations to attend on is preventing AAM to achieve its highest potential. Therefore, in the future, it would be more beneficial if we have a better module to localize interesting spatial locations in the video frames instead of the human detector.

\noindent\textbf{Qualitative Analysis of Objects Beholder}

\begin{figure*}[!ht]
     \centering
     \begin{subfigure}[b]{0.8\textwidth}
         \centering
         \includegraphics[width=\textwidth]{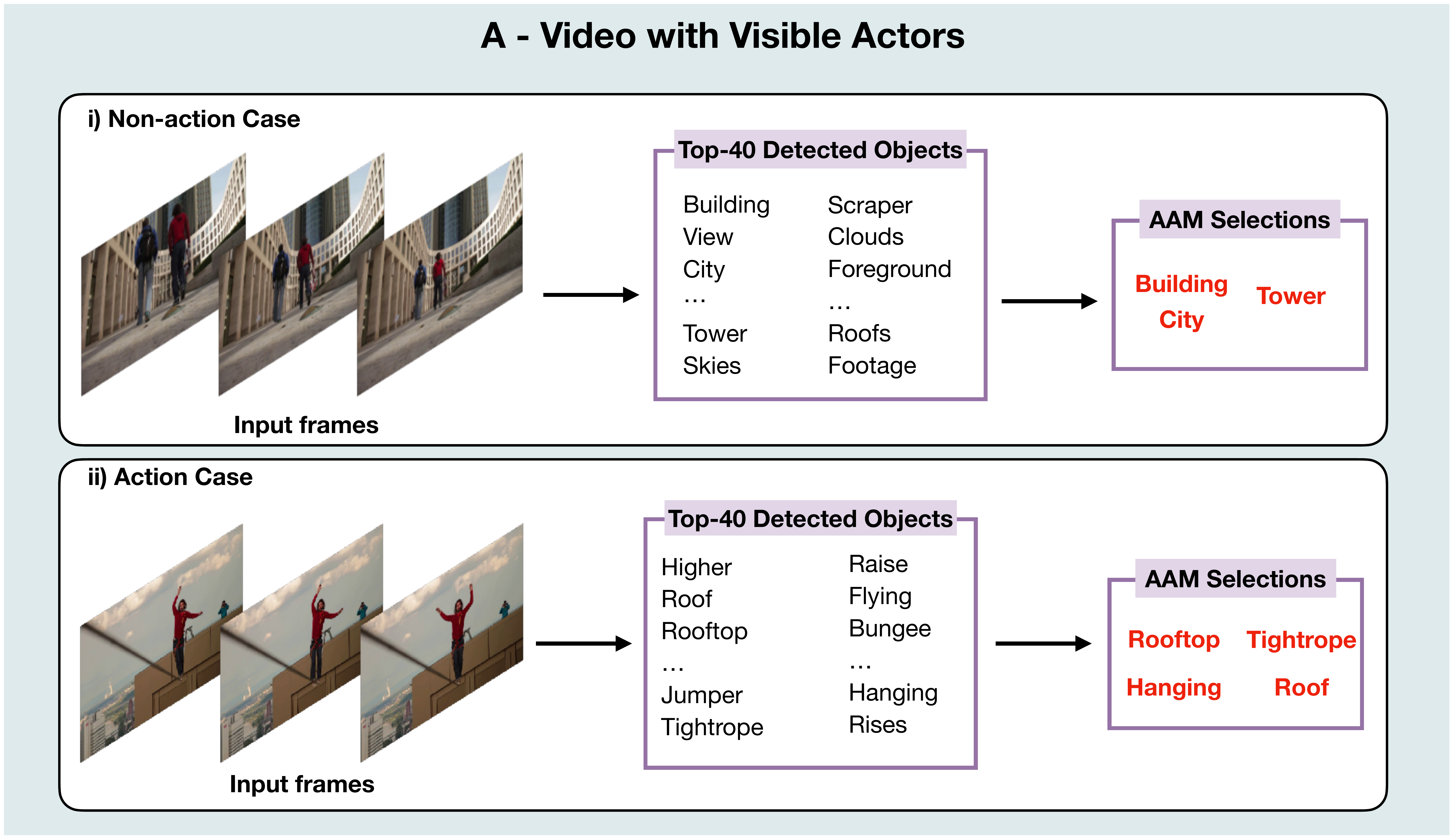}
         \label{fig:objbeh_1}
     \vspace{-3mm}
     \end{subfigure}
     \begin{subfigure}[b]{0.8\textwidth}
         \centering
         \includegraphics[width=\textwidth]{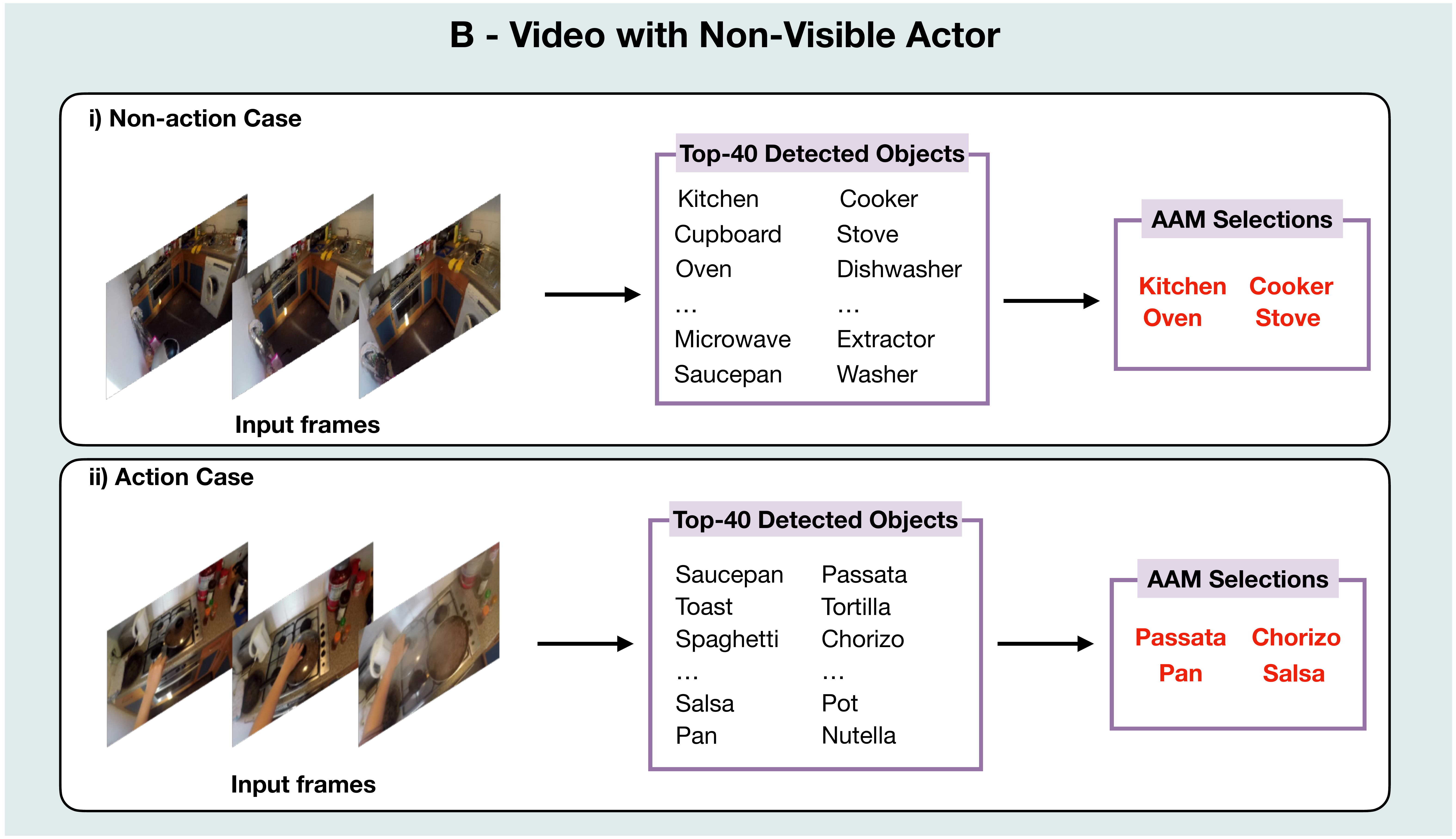}
         \label{fig:objbeh_2}
     \vspace{-3mm}
     \end{subfigure}
\caption{Qualitative results to illustrate the effectiveness of Objects Beholder with AAM in (A) videos with visible actors and (B) videos with non-visible actor. In each case, we illustrate two instances of having action and without action. The input frames are shown on the left, its objects detected by CLIP are shown in the middle, and the most relevant objects selected by AAM are shown in the right.}
\label{fig:AAM_objbeholder}
\end{figure*}

Fig.~\ref{fig:AAM_objbeholder} visualizes how AOE-Net can take advantage of Objects Beholder.
In this figure, we showcase two video for two distinct categories of (A) visible actors and (B) non-visible actors. in (A), the actors are visible and commits the action of tightrope walking, therefore, our AOE-Net can take advantage of all the beholders. Whereas in (B), the actors are not visible in the video frame and commits the action of cooking, therefore, our AOE-Net can only relies on Objects Beholder.

\noindent In each illustration of (A) and (B), we visualize either when the action does not happen (A.i and B.i) and when the action is happening (A.ii and B.ii).

\noindent As we can see in Fig.~\ref{fig:AAM_objbeholder}.(A), the objects detected in non-action case and action case are very different. Specifically in Fig.~\ref{fig:AAM_objbeholder}(A.i), the non-action scene is captured through objects like ``City", ``Building", ``Tower", etc. Whereas in Fig.~\ref{fig:AAM_objbeholder}(A.ii), the action scene includes ``Rooftop", ``Tightrope", ``Roof", and ``Hanging", etc.

\noindent Likewise, in Fig.~\ref{fig:AAM_objbeholder}.(B), the objects detected in each cases of non-action and action are very different. On one hand, in Fig.~\ref{fig:AAM_objbeholder}(B.i), detected objects are ``Kitchen", ``Cooker", ``Oven", and ``Stove" etc. On the other hand, in Fig.~\ref{fig:AAM_objbeholder}(B.ii), the action scene objects consist of ``Passata", ``Chorizo", ``Pan", and ``Salsa", etc.

\subsection{Qualitative Analysis of AOE-Net}

\begin{figure}[!ht]
     \centering
     \begin{subfigure}[b]{0.8\textwidth}
         \centering
         \includegraphics[width=\textwidth]{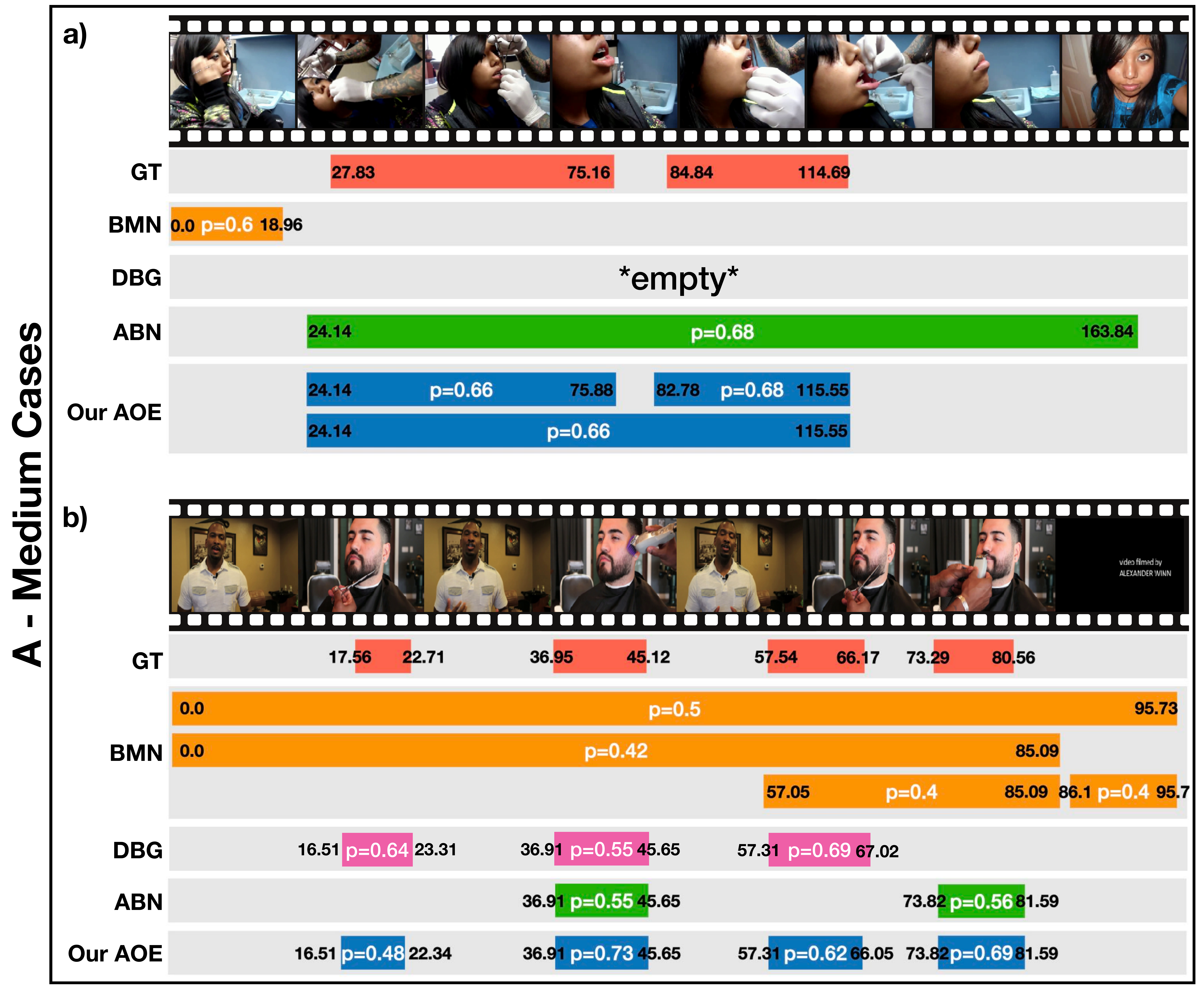}
         \label{fig:anet2}
     \vspace{-3mm}
     \end{subfigure}
     \begin{subfigure}[b]{0.8\textwidth}
         \centering
         \includegraphics[width=\textwidth]{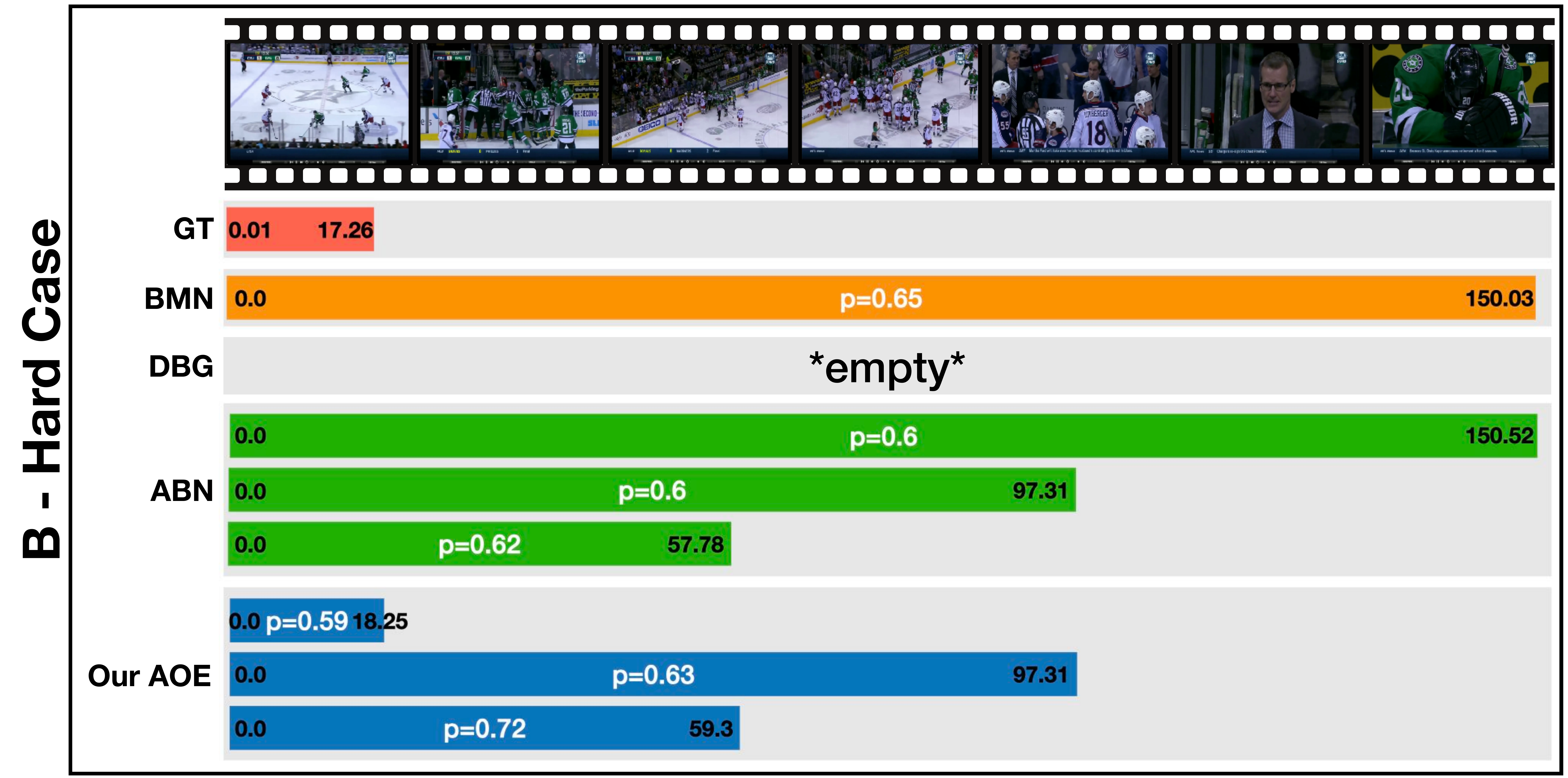}
         \label{fig:anet3}
     \vspace{-3mm}
     \end{subfigure}
     \begin{subfigure}[b]{0.8\textwidth}
         \centering
         \includegraphics[width=\textwidth]{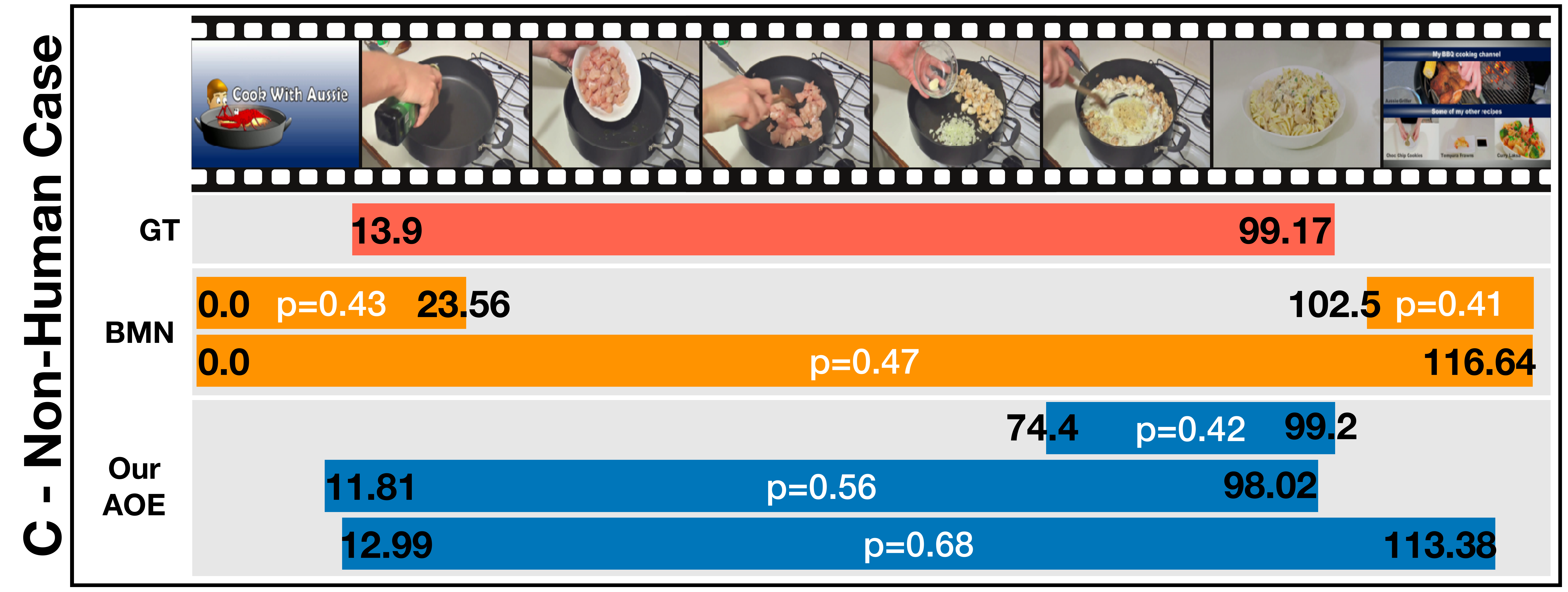}
         \label{fig:anet4}
     \end{subfigure}
\vspace{-3mm}   
\caption{Qualitative results in TAPG on ActivityNet-1.3 \cite{caba2015activitynet} dataset.}
        \label{fig:qualitative_anet}
\end{figure}

%\hl{fig. 9(a)(c) should be in 1 group whereas fig.9(b) will be another group. please analyze more the hard case of fig.9(b)}

The qualitative results of our AOE-Net in TAPG of ActivityNet-1.3 \cite{caba2015activitynet} comparing with previous SOTA works \cite{bmn, dbg, KhoaVo_Access} are illustrated in Fig.~\ref{fig:qualitative_anet}. In this figure, we showed two medium cases and a hard case and case of egocentric video. In each video, the proposals from all methods with higher scores than 0.4 are selected to show in the qualitative examples.

In videos of simple cases (Fig.~\ref{fig:qualitative_anet}-A), the actors who are receiving pierces (or trimming in video(b)) is easily observed from the whole frame, however, the piercing action is a bit difficult to be identified because it's from the hand of the doctor (or the taylor in video (b)), who is outside of the video view. Therefore, BMN \cite{bmn} and DBG \cite{dbg} completely failed to propose the exact action intervals. Likewise, ABN \cite{KhoaVo_Access} is tricked by the video and proposes an interval from the beginning of the first groundtruth action of piercing until the credit cut. On the other hand, our proposed AOE-Net can propose intervals that match with the groundtruth actions. This explains a lot for the contributions of both actors beholder and objects beholder, which provide more informative features than previous works to give good results.

In the video of hard case(Fig.~\ref{fig:qualitative_anet}-A), the actors are hockey players, who appear very small in the video frames. Therefore, the "hockey playing" activity, which appear at the beginning of this video, is very difficult to be distinguished to "celebrating" activity, which takes place right after the former. This is explanable because we need to carefully observe the movements of hockey players carefully to see this difference. Therfore, all BMN and DBG failed to recognize the groundtruth action interval. Meanwhile, ABN can propose an interval that covers the scene of the field but not necessesarily the groundtruth action. On the other hand, our proposed AOE-Net can propose an interval that closely matches the groundtruth action. This again, explains the contributions of our actors and objects beholders.

In the non-human case (Fig.~\ref{fig:qualitative_anet}-C), an actor shows their hands doing cooking on a pan in the interval of [13.9-99.17], while in [0.0-13.9] and [99.17-116.64] the advertisements are displayed. As the actor only shows their hands in the video frames to commit the action, they cannot be detected by the Actors Beholder. However, thanks to our Objects Beholder, the advertisement intervals at the beginning and the end of the video are easily perceived, hence the true action interval is detected in between. Contrarily, a previous SOTA model, BMN [3], mis-perceived the advertisement intervals as true actions and classifies them as separate action intervals, or mistakenly combines them with the true action interval in between.

%\begin{figure}
%     \centering
%     \begin{subfigure}[b]{0.8\textwidth}
%         \centering
%         \includegraphics[width=\textwidth]{Figs/anet_1.jpeg}
%         \label{fig:anet1}
%     \end{subfigure}
%     \begin{subfigure}[b]{0.8\textwidth}
%         \centering
%         \includegraphics[width=\textwidth]{Figs/anet_2.jpeg}
%         \label{fig:anet2}
%     \end{subfigure}
%     \begin{subfigure}[b]{0.8\textwidth}
%         \centering
%         \includegraphics[width=\textwidth]{Figs/anet_3.jpeg}
%         \label{fig:anet3}
%     \end{subfigure}
%\caption{Qualitative results in TAPG on ActivityNet-1.3 \cite{caba2015activitynet} dataset.}
%        \label{fig:qualitative_anet}
%\end{figure}

%\begin{figure}
%     \centering
%     \begin{subfigure}[b]{0.8\textwidth}
%         \centering
%         \includegraphics[width=\textwidth]{Figs/thumos_1.jpeg}
%         \label{fig:anet1}
%     \end{subfigure}
%     \begin{subfigure}[b]{0.8\textwidth}
%         \centering
%         \includegraphics[width=\textwidth]{Figs/thumos_2.jpeg}
%         \label{fig:anet2}
%     \end{subfigure}
%     \begin{subfigure}[b]{0.8\textwidth}
%         \centering
%         \includegraphics[width=\textwidth]{Figs/thumos_3.jpeg}
%         \label{fig:anet3}
%     \end{subfigure}
%\caption{Qualitative results in TAPG on THUMOS-14 \cite{THUMOS14} dataset.}
%        \label{fig:qualitative_thumos}
%\end{figure}

\vspace{-1mm}
\section{Conclusion and Discussion}
\vspace{-1mm}
In this paper, we attempt to simulate the human perceiving ability and proposed a novel AOE-Net to locate actions in untrimmed videos. Our AOE-Net contains two modules: PMR and BMM. PMR extracts visual-linguistic representation of each snippet with four beholders. Environment beholder and actors beholder capture global and local visual features of environment and main actors, respectively. Objects beholder extracts linguistic feature from relevant objects. The last beholder aims to model the relations between main actors, relevant objects and environment. To focus on an arbitrary number of main actor(s) or relevant objects, we introduced AAM. The qualitative and quantitative results conducted on ActivityNet-1.3 and THUMOS-14 datasets on both TAPG and TAD tasks evidently suggest that our proposed AOE-Net outperforms SOTA methods. To prove the effectiveness of AOE-Net, we provided ablation studies to show the contribution of each beholder, the effectiveness of proposed AAM, network efficiency, as well as the robustness of AOE-Net when performing on egocentric videos with EPIC-KITCHENS 100 dataset. We further investigate the performance of AOE-Net with various backbone network configurations. These results prove that replicating human perceiving ability in video understanding is a promising track to follow and further explore in the future. 

There are several potential future directions from this research. First, while main actors and relevant objects provide important impact to both TAPG and TAD tasks, it would be of great interest to investigate the impact of human body parts (e.g. hands, legs) and the interaction between them with objects in localizing human activities in untrimmed videos. Finally, integrating our method with human tracking (i.e. main actors tracking) might result in even better performance.

\bmhead{Acknowledgments}
This material is based upon work supported by the National Science Foundation under Award No OIA-1946391 and NSF 1920920.

\newpage
\bibliography{sn-bibliography}% common bib file


\begin{thebibliography}{69}
\ifx \bisbn   \undefined \def \bisbn  #1{ISBN #1}\fi
\ifx \binits  \undefined \def \binits#1{#1}\fi
\ifx \bauthor  \undefined \def \bauthor#1{#1}\fi
\ifx \batitle  \undefined \def \batitle#1{#1}\fi
\ifx \bjtitle  \undefined \def \bjtitle#1{#1}\fi
\ifx \bvolume  \undefined \def \bvolume#1{\textbf{#1}}\fi
\ifx \byear  \undefined \def \byear#1{#1}\fi
\ifx \bissue  \undefined \def \bissue#1{#1}\fi
\ifx \bfpage  \undefined \def \bfpage#1{#1}\fi
\ifx \blpage  \undefined \def \blpage #1{#1}\fi
\ifx \burl  \undefined \def \burl#1{\textsf{#1}}\fi
\ifx \doiurl  \undefined \def \doiurl#1{\url{https://doi.org/#1}}\fi
\ifx \betal  \undefined \def \betal{\textit{et al.}}\fi
\ifx \binstitute  \undefined \def \binstitute#1{#1}\fi
\ifx \binstitutionaled  \undefined \def \binstitutionaled#1{#1}\fi
\ifx \bctitle  \undefined \def \bctitle#1{#1}\fi
\ifx \beditor  \undefined \def \beditor#1{#1}\fi
\ifx \bpublisher  \undefined \def \bpublisher#1{#1}\fi
\ifx \bbtitle  \undefined \def \bbtitle#1{#1}\fi
\ifx \bedition  \undefined \def \bedition#1{#1}\fi
\ifx \bseriesno  \undefined \def \bseriesno#1{#1}\fi
\ifx \blocation  \undefined \def \blocation#1{#1}\fi
\ifx \bsertitle  \undefined \def \bsertitle#1{#1}\fi
\ifx \bsnm \undefined \def \bsnm#1{#1}\fi
\ifx \bsuffix \undefined \def \bsuffix#1{#1}\fi
\ifx \bparticle \undefined \def \bparticle#1{#1}\fi
\ifx \barticle \undefined \def \barticle#1{#1}\fi
\bibcommenthead
\ifx \bconfdate \undefined \def \bconfdate #1{#1}\fi
\ifx \botherref \undefined \def \botherref #1{#1}\fi
\ifx \url \undefined \def \url#1{\textsf{#1}}\fi
\ifx \bchapter \undefined \def \bchapter#1{#1}\fi
\ifx \bbook \undefined \def \bbook#1{#1}\fi
\ifx \bcomment \undefined \def \bcomment#1{#1}\fi
\ifx \oauthor \undefined \def \oauthor#1{#1}\fi
\ifx \citeauthoryear \undefined \def \citeauthoryear#1{#1}\fi
\ifx \endbibitem  \undefined \def \endbibitem {}\fi
\ifx \bconflocation  \undefined \def \bconflocation#1{#1}\fi
\ifx \arxivurl  \undefined \def \arxivurl#1{\textsf{#1}}\fi
\csname PreBibitemsHook\endcsname

%%% 1
\bibitem{lin2018bsn}
\begin{bchapter}
\bauthor{\bsnm{Lin}, \binits{T.}},
\bauthor{\bsnm{Zhao}, \binits{X.}},
\bauthor{\bsnm{Su}, \binits{H.}},
\bauthor{\bsnm{Wang}, \binits{C.}},
\bauthor{\bsnm{Yang}, \binits{M.}}:
\bctitle{Bsn: Boundary sensitive network for temporal action proposal
  generation}.
In: \bbtitle{ECCV}
(\byear{2018})
\end{bchapter}
\endbibitem

%%% 2
\bibitem{BSN++}
\begin{bchapter}
\bauthor{\bsnm{Su}, \binits{H.}},
\bauthor{\bsnm{Gan}, \binits{W.}},
\bauthor{\bsnm{Wu}, \binits{W.}},
\bauthor{\bsnm{Yan}, \binits{J.}},
\bauthor{\bsnm{Qiao}, \binits{Y.}}:
\bctitle{{BSN++:} complementary boundary regressor with scale-balanced relation
  modeling for temporal action proposal generation}.
In: \bbtitle{ACCV}
(\byear{2020})
\end{bchapter}
\endbibitem

%%% 3
\bibitem{bmn}
\begin{bchapter}
\bauthor{\bsnm{Lin}, \binits{T.}},
\bauthor{\bsnm{Liu}, \binits{X.}},
\bauthor{\bsnm{Li}, \binits{X.}},
\bauthor{\bsnm{Ding}, \binits{E.}},
\bauthor{\bsnm{Wen}, \binits{S.}}:
\bctitle{Bmn: Boundary-matching network for temporal action proposal
  generation}.
In: \bbtitle{ICCV}
(\byear{2019})
\end{bchapter}
\endbibitem

%%% 4
\bibitem{dbg}
\begin{botherref}
\oauthor{\bsnm{Lin}, \binits{C.}},
\oauthor{\bsnm{Li}, \binits{J.}},
\oauthor{\bsnm{Wang}, \binits{Y.}},
\oauthor{\bsnm{Tai}, \binits{Y.}},
\oauthor{\bsnm{Luo}, \binits{D.}},
\oauthor{\bsnm{Cui}, \binits{Z.}},
\oauthor{\bsnm{Wang}, \binits{C.}},
\oauthor{\bsnm{Li}, \binits{J.}},
\oauthor{\bsnm{Huang}, \binits{F.}},
\oauthor{\bsnm{Ji}, \binits{R.}}:
Fast learning of temporal action proposal via dense boundary generator.
AAAI,
11499--11506
(2020)
\end{botherref}
\endbibitem

%%% 5
\bibitem{xu2020gtad}
\begin{bchapter}
\bauthor{\bsnm{Xu}, \binits{M.}},
\bauthor{\bsnm{Zhao}, \binits{C.}},
\bauthor{\bsnm{Rojas}, \binits{D.S.}},
\bauthor{\bsnm{Thabet}, \binits{A.}},
\bauthor{\bsnm{Ghanem}, \binits{B.}}:
\bctitle{G-tad: Sub-graph localization for temporal action detection}.
In: \bbtitle{CVPR}
(\byear{2020})
\end{bchapter}
\endbibitem

%%% 6
\bibitem{KhoaVo_Access}
\begin{barticle}
\bauthor{\bsnm{Vo}, \binits{K.}},
\bauthor{\bsnm{Yamazaki}, \binits{K.}},
\bauthor{\bsnm{Truong}, \binits{S.}},
\bauthor{\bsnm{Tran}, \binits{M.-T.}},
\bauthor{\bsnm{Sugimoto}, \binits{A.}},
\bauthor{\bsnm{Le}, \binits{N.}}:
\batitle{Abn: Agent-aware boundary networks for temporal action proposal
  generation}.
\bjtitle{IEEE Access}
\bvolume{9},
\bfpage{126431}--\blpage{126445}
(\byear{2021})
\end{barticle}
\endbibitem

%%% 7
\bibitem{KhoaVo_ICASSP}
\begin{bchapter}
\bauthor{\bsnm{Vo-Ho}, \binits{V.-K.}},
\bauthor{\bsnm{Le}, \binits{N.}},
\bauthor{\bsnm{Kamazaki}, \binits{K.}},
\bauthor{\bsnm{Sugimoto}, \binits{A.}},
\bauthor{\bsnm{Tran}, \binits{M.-T.}}:
\bctitle{Agent-environment network for temporal action proposal generation}.
In: \bbtitle{ICASSP},
pp. \bfpage{2160}--\blpage{2164}
(\byear{2021})
\end{bchapter}
\endbibitem

%%% 8
\bibitem{anchor_2}
\begin{bchapter}
\bauthor{\bsnm{Shou}, \binits{Z.}},
\bauthor{\bsnm{Wang}, \binits{D.}},
\bauthor{\bsnm{Chang}, \binits{S.-F.}}:
\bctitle{Temporal action localization in untrimmed videos via multi-stage
  cnns}.
In: \bbtitle{CVPR}
(\byear{2016})
\end{bchapter}
\endbibitem

%%% 9
\bibitem{Jiyang2017}
\begin{botherref}
\oauthor{\bsnm{{Gao}}, \binits{J.}},
\oauthor{\bsnm{{Yang}}, \binits{Z.}},
\oauthor{\bsnm{{Nevatia}}, \binits{R.}}:
{Cascaded Boundary Regression for Temporal Action Detection}.
arXiv e-prints,
1705--01180
(2017)
{\href{https://arxiv.org/abs/1705.01180}{{arXiv:1705.01180}}}
{[cs.CV]}
\end{botherref}
\endbibitem

%%% 10
\bibitem{CTAP}
\begin{bchapter}
\bauthor{\bsnm{Gao}, \binits{J.}},
\bauthor{\bsnm{Chen}, \binits{K.}},
\bauthor{\bsnm{Nevatia}, \binits{R.}}:
\bctitle{Ctap: Complementary temporal action proposal generation}.
In: \bbtitle{ECCV}
(\byear{2018})
\end{bchapter}
\endbibitem

%%% 11
\bibitem{Gao_2018_CVPR}
\begin{bchapter}
\bauthor{\bsnm{Gao}, \binits{J.}},
\bauthor{\bsnm{Ge}, \binits{R.}},
\bauthor{\bsnm{Chen}, \binits{K.}},
\bauthor{\bsnm{Nevatia}, \binits{R.}}:
\bctitle{Motion-appearance co-memory networks for video question answering}.
In: \bbtitle{CVPR}
(\byear{2018})
\end{bchapter}
\endbibitem

%%% 12
\bibitem{caba2015activitynet}
\begin{bchapter}
\bauthor{\bsnm{Fabian Caba~Heilbron}, \binits{B.G.} \bsuffix{Victor~Escorcia}},
\bauthor{\bsnm{Niebles}, \binits{J.C.}}:
\bctitle{Activitynet: A large-scale video benchmark for human activity
  understanding}.
In: \bbtitle{CVPR},
pp. \bfpage{961}--\blpage{970}
(\byear{2015})
\end{bchapter}
\endbibitem

%%% 13
\bibitem{THUMOS14}
\begin{botherref}
\oauthor{\bsnm{Jiang}, \binits{Y.-G.}},
\oauthor{\bsnm{Liu}, \binits{J.}},
\oauthor{\bsnm{Roshan~Zamir}, \binits{A.}},
\oauthor{\bsnm{Toderici}, \binits{G.}},
\oauthor{\bsnm{Laptev}, \binits{I.}},
\oauthor{\bsnm{Shah}, \binits{M.}},
\oauthor{\bsnm{Sukthankar}, \binits{R.}}:
{THUMOS} Challenge: Action Recognition with a Large Number of Classes.
\url{http://crcv.ucf.edu/THUMOS14/}
(2014)
\end{botherref}
\endbibitem

%%% 14
\bibitem{krishna2017dense}
\begin{bchapter}
\bauthor{\bsnm{Krishna}, \binits{R.}},
\bauthor{\bsnm{Hata}, \binits{K.}},
\bauthor{\bsnm{Ren}, \binits{F.}},
\bauthor{\bsnm{Fei-Fei}, \binits{L.}},
\bauthor{\bsnm{Carlos~Niebles}, \binits{J.}}:
\bctitle{Dense-captioning events in videos}.
In: \bbtitle{ICCV},
pp. \bfpage{706}--\blpage{715}
(\byear{2017})
\end{bchapter}
\endbibitem

%%% 15
\bibitem{Kinetics}
\begin{botherref}
\oauthor{\bsnm{Kay}, \binits{W.}},
\oauthor{\bsnm{Carreira}, \binits{J.}},
\oauthor{}, et al.:
The kinetics human action video dataset.
arXiv preprint arXiv:1705.06950
(2017)
\end{botherref}
\endbibitem

%%% 16
\bibitem{actionproposal_2016}
\begin{bchapter}
\bauthor{\bsnm{Richard}, \binits{A.}},
\bauthor{\bsnm{Gall}, \binits{J.}}:
\bctitle{Temporal action detection using a statistical language model}.
In: \bbtitle{CVPR}
(\byear{2016})
\end{bchapter}
\endbibitem

%%% 17
\bibitem{FasterR_CNN_Action}
\begin{bchapter}
\bauthor{\bsnm{{Chao}}, \binits{Y.}},
\bauthor{\bsnm{{Vijayanarasimhan}}, \binits{S.}},
\bauthor{\bsnm{{Seybold}}, \binits{B.}},
\bauthor{\bsnm{{Ross}}, \binits{D.A.}},
\bauthor{\bsnm{{Deng}}, \binits{J.}},
\bauthor{\bsnm{{Sukthankar}}, \binits{R.}}:
\bctitle{Rethinking the faster r-cnn architecture for temporal action
  localization}.
In: \bbtitle{CVPR},
pp. \bfpage{1130}--\blpage{1139}
(\byear{2018})
\end{bchapter}
\endbibitem

%%% 18
\bibitem{anchor_1}
\begin{bchapter}
\bauthor{\bsnm{Heilbron}, \binits{F.C.}},
\bauthor{\bsnm{Niebles}, \binits{J.C.}},
\bauthor{\bsnm{Ghanem}, \binits{B.}}:
\bctitle{Fast temporal activity proposals for efficient detection of human
  actions in untrimmed videos}.
In: \bbtitle{CVPR}
(\byear{2016})
\end{bchapter}
\endbibitem

%%% 19
\bibitem{anchor_3}
\begin{bchapter}
\bauthor{\bsnm{Gao}, \binits{J.}},
\bauthor{\bsnm{Yang}, \binits{Z.}},
\bauthor{\bsnm{Chen}, \binits{K.}},
\bauthor{\bsnm{Sun}, \binits{C.}},
\bauthor{\bsnm{Nevatia}, \binits{R.}}:
\bctitle{Turn tap: Temporal unit regression network for temporal action
  proposals}.
In: \bbtitle{ICCV}
(\byear{2017})
\end{bchapter}
\endbibitem

%%% 20
\bibitem{C3D}
\begin{barticle}
\bauthor{\bsnm{{Ji}}, \binits{S.}},
\bauthor{\bsnm{{Xu}}, \binits{W.}},
\bauthor{\bsnm{{Yang}}, \binits{M.}},
\bauthor{\bsnm{{Yu}}, \binits{K.}}:
\batitle{3d convolutional neural networks for human action recognition}.
\bjtitle{IEEE TPAMI}
\bvolume{35}(\bissue{1}),
\bfpage{221}--\blpage{231}
(\byear{2013})
\end{barticle}
\endbibitem

%%% 21
\bibitem{i3d_2017}
\begin{bchapter}
\bauthor{\bsnm{Carreira}, \binits{J.}},
\bauthor{\bsnm{Zisserman}, \binits{A.}}:
\bctitle{Quo vadis, action recognition? a new model and the kinetics dataset}.
In: \bbtitle{CVPR},
pp. \bfpage{6299}--\blpage{6308}
(\byear{2017})
\end{bchapter}
\endbibitem

%%% 22
\bibitem{2_stream_1}
\begin{bchapter}
\bauthor{\bsnm{Simonyan}, \binits{K.}},
\bauthor{\bsnm{Zisserman}, \binits{A.}}:
\bctitle{Two-stream convolutional networks for action recognition in videos}.
In: \bbtitle{NIPS}.
\bsertitle{NIPS'14},
pp. \bfpage{568}--\blpage{576}.
\bpublisher{MIT Press},
\blocation{Cambridge, MA, USA}
(\byear{2014})
\end{bchapter}
\endbibitem

%%% 23
\bibitem{SlowFast}
\begin{bchapter}
\bauthor{\bsnm{Feichtenhofer}, \binits{C.}},
\bauthor{\bsnm{Fan}, \binits{H.}},
\bauthor{\bsnm{Malik}, \binits{J.}},
\bauthor{\bsnm{He}, \binits{K.}}:
\bctitle{Slowfast networks for video recognition}.
In: \bbtitle{ICCV}
(\byear{2019})
\end{bchapter}
\endbibitem

%%% 24
\bibitem{mei2020vision}
\begin{botherref}
\oauthor{\bsnm{Mei}, \binits{T.}},
\oauthor{\bsnm{Zhang}, \binits{W.}},
\oauthor{\bsnm{Yao}, \binits{T.}}:
Vision and language: from visual perception to content creation.
APSIPA TSIP
\textbf{9}
(2020)
\end{botherref}
\endbibitem

%%% 25
\bibitem{anderson2018bottom}
\begin{bchapter}
\bauthor{\bsnm{Anderson}, \binits{P.}},
\bauthor{\bsnm{He}, \binits{X.}},
\bauthor{\bsnm{Buehler}, \binits{C.}},
\bauthor{\bsnm{Teney}, \binits{D.}},
\bauthor{\bsnm{Johnson}, \binits{M.}},
\bauthor{\bsnm{Gould}, \binits{S.}},
\bauthor{\bsnm{Zhang}, \binits{L.}}:
\bctitle{Bottom-up and top-down attention for image captioning and visual
  question answering}.
In: \bbtitle{CVPR},
pp. \bfpage{6077}--\blpage{6086}
(\byear{2018})
\end{bchapter}
\endbibitem

%%% 26
\bibitem{radford2021learning}
\begin{botherref}
\oauthor{\bsnm{Radford}, \binits{A.}},
\oauthor{\bsnm{Kim}, \binits{J.W.}}, et al.:
Learning transferable visual models from natural language supervision.
arXiv preprint arXiv:2103.00020
(2021)
\end{botherref}
\endbibitem

%%% 27
\bibitem{heilbron2016fast}
\begin{bchapter}
\bauthor{\bsnm{Heilbron}, \binits{F.C.}},
\bauthor{\bsnm{Niebles}, \binits{J.C.}},
\bauthor{\bsnm{Ghanem}, \binits{B.}}:
\bctitle{Fast temporal activity proposals for efficient detection of human
  actions in untrimmed videos}.
In: \bbtitle{CVPR},
pp. \bfpage{1914}--\blpage{1923}
(\byear{2016})
\end{bchapter}
\endbibitem

%%% 28
\bibitem{FasterRCNN}
\begin{bchapter}
\bauthor{\bsnm{Ren}, \binits{S.}},
\bauthor{\bsnm{He}, \binits{K.}},
\bauthor{\bsnm{Girshick}, \binits{R.}},
\bauthor{\bsnm{Sun}, \binits{J.}}:
\bctitle{Faster r-cnn: Towards real-time object detection with region proposal
  networks}.
In: \bbtitle{NeurIPS},
pp. \bfpage{91}--\blpage{99}
(\byear{2015})
\end{bchapter}
\endbibitem

%%% 29
\bibitem{RetinaNet}
\begin{bchapter}
\bauthor{\bsnm{{Lin}}, \binits{T.}},
\bauthor{\bsnm{{Goyal}}, \binits{P.}},
\bauthor{\bsnm{{Girshick}}, \binits{R.}},
\bauthor{\bsnm{{He}}, \binits{K.}},
\bauthor{\bsnm{{Dollár}}, \binits{P.}}:
\bctitle{Focal loss for dense object detection}.
In: \bbtitle{ICCV},
pp. \bfpage{2999}--\blpage{3007}
(\byear{2017})
\end{bchapter}
\endbibitem

%%% 30
\bibitem{yolov3}
\begin{botherref}
\oauthor{\bsnm{Redmon}, \binits{J.}},
\oauthor{\bsnm{Farhadi}, \binits{A.}}:
Yolov3: An incremental improvement.
arXiv
(2018)
\end{botherref}
\endbibitem

%%% 31
\bibitem{C3D_3}
\begin{bchapter}
\bauthor{\bsnm{Tran}, \binits{D.}},
\bauthor{\bsnm{Bourdev}, \binits{L.}},
\bauthor{\bsnm{Fergus}, \binits{R.}},
\bauthor{\bsnm{Torresani}, \binits{L.}},
\bauthor{\bsnm{Paluri}, \binits{M.}}:
\bctitle{Learning spatiotemporal features with 3d convolutional networks}.
In: \bbtitle{ICCV},
pp. \bfpage{4489}--\blpage{4497}
(\byear{2015})
\end{bchapter}
\endbibitem

%%% 32
\bibitem{boundary_0}
\begin{bchapter}
\bauthor{\bsnm{Zhao}, \binits{Y.}},
\bauthor{\bsnm{Xiong}, \binits{Y.}},
\bauthor{\bsnm{Wang}, \binits{L.}},
\bauthor{\bsnm{Wu}, \binits{Z.}},
\bauthor{\bsnm{Tang}, \binits{X.}},
\bauthor{\bsnm{Lin}, \binits{D.}}:
\bctitle{Temporal action detection with structured segment networks}.
In: \bbtitle{ICCV}
(\byear{2017})
\end{bchapter}
\endbibitem

%%% 33
\bibitem{liu2019multi}
\begin{bchapter}
\bauthor{\bsnm{Liu}, \binits{Y.}},
\bauthor{\bsnm{Ma}, \binits{L.}},
\bauthor{\bsnm{Zhang}, \binits{Y.}},
\bauthor{\bsnm{Liu}, \binits{W.}},
\bauthor{\bsnm{Chang}, \binits{S.-F.}}:
\bctitle{Multi-granularity generator for temporal action proposal}.
In: \bbtitle{CVPR}
(\byear{2019})
\end{bchapter}
\endbibitem

%%% 34
\bibitem{itti1998model}
\begin{barticle}
\bauthor{\bsnm{Itti}, \binits{L.}},
\bauthor{\bsnm{Koch}, \binits{C.}},
\bauthor{\bsnm{Niebur}, \binits{E.}}:
\batitle{A model of saliency-based visual attention for rapid scene analysis}.
\bjtitle{IEEE TPAMI}
\bvolume{20}(\bissue{11}),
\bfpage{1254}--\blpage{1259}
(\byear{1998})
\end{barticle}
\endbibitem

%%% 35
\bibitem{chaudhari2021attentive}
\begin{barticle}
\bauthor{\bsnm{Chaudhari}, \binits{S.}},
\bauthor{\bsnm{Mithal}, \binits{V.}},
\bauthor{\bsnm{Polatkan}, \binits{G.}},
\bauthor{\bsnm{Ramanath}, \binits{R.}}:
\batitle{An attentive survey of attention models}.
\bjtitle{TIST}
\bvolume{12}(\bissue{5}),
\bfpage{1}--\blpage{32}
(\byear{2021})
\end{barticle}
\endbibitem

%%% 36
\bibitem{bahdanau2014neural}
\begin{botherref}
\oauthor{\bsnm{Bahdanau}, \binits{D.}},
\oauthor{\bsnm{Cho}, \binits{K.}},
\oauthor{\bsnm{Bengio}, \binits{Y.}}:
Neural machine translation by jointly learning to align and translate.
arXiv preprint arXiv:1409.0473
(2014)
\end{botherref}
\endbibitem

%%% 37
\bibitem{cho2015describing}
\begin{barticle}
\bauthor{\bsnm{Cho}, \binits{K.}},
\bauthor{\bsnm{Courville}, \binits{A.}},
\bauthor{\bsnm{Bengio}, \binits{Y.}}:
\batitle{Describing multimedia content using attention-based encoder-decoder
  networks}.
\bjtitle{IEEE Transactions on Multimedia}
\bvolume{17}(\bissue{11}),
\bfpage{1875}--\blpage{1886}
(\byear{2015})
\end{barticle}
\endbibitem

%%% 38
\bibitem{galassi2020attention}
\begin{botherref}
\oauthor{\bsnm{Galassi}, \binits{A.}},
\oauthor{\bsnm{Lippi}, \binits{M.}},
\oauthor{\bsnm{Torroni}, \binits{P.}}:
Attention in natural language processing.
IEEE Transactions on Neural Networks and Learning Systems
(2020)
\end{botherref}
\endbibitem

%%% 39
\bibitem{chaudhari2019attentive}
\begin{botherref}
\oauthor{\bsnm{Chaudhari}, \binits{S.}},
\oauthor{\bsnm{Polatkan}, \binits{G.}},
\oauthor{\bsnm{Ramanath}, \binits{R.}},
\oauthor{\bsnm{Mithal}, \binits{V.}}:
An attentive survey of attention models.
arXiv preprint arXiv:1904.02874
(2019)
\end{botherref}
\endbibitem

%%% 40
\bibitem{attention_is_all_you_need}
\begin{bchapter}
\bauthor{\bsnm{Vaswani}, \binits{A.}},
\bauthor{\bsnm{Shazeer}, \binits{N.}},
\bauthor{\bsnm{Parmar}, \binits{N.}},
\bauthor{\bsnm{Uszkoreit}, \binits{J.}},
\bauthor{\bsnm{Jones}, \binits{L.}},
\bauthor{\bsnm{Gomez}, \binits{A.N.}},
\bauthor{\bsnm{Kaiser}, \binits{L.u.}},
\bauthor{\bsnm{Polosukhin}, \binits{I.}}:
\bctitle{Attention is all you need}.
In: \bbtitle{NeurIPS}.
\bpublisher{Curran Associates, Inc.}, \blocation{???}
(\byear{2017})
\end{bchapter}
\endbibitem

%%% 41
\bibitem{xu2015show}
\begin{bchapter}
\bauthor{\bsnm{Xu}, \binits{K.}},
\bauthor{\bsnm{Ba}, \binits{J.}},
\bauthor{\bsnm{Kiros}, \binits{R.}},
\bauthor{\bsnm{Cho}, \binits{K.}},
\bauthor{\bsnm{Courville}, \binits{A.}},
\bauthor{\bsnm{Salakhudinov}, \binits{R.}},
\bauthor{\bsnm{Zemel}, \binits{R.}},
\bauthor{\bsnm{Bengio}, \binits{Y.}}:
\bctitle{Show, attend and tell: Neural image caption generation with visual
  attention}.
In: \bbtitle{ICML},
pp. \bfpage{2048}--\blpage{2057}
(\byear{2015}).
\bcomment{PMLR}
\end{bchapter}
\endbibitem

%%% 42
\bibitem{Saccader_NIPS2019}
\begin{bchapter}
\bauthor{\bsnm{Elsayed}, \binits{G.}},
\bauthor{\bsnm{Kornblith}, \binits{S.}},
\bauthor{\bsnm{Le}, \binits{Q.V.}}:
\bctitle{Saccader: Improving accuracy of hard attention models for vision}.
In: \bbtitle{NeurIPS}
(\byear{2019})
\end{bchapter}
\endbibitem

%%% 43
\bibitem{patro2018differential}
\begin{bchapter}
\bauthor{\bsnm{Patro}, \binits{B.}},
\bauthor{\bsnm{Namboodiri}, \binits{V.P.}}:
\bctitle{Differential attention for visual question answering}.
In: \bbtitle{CVPR},
pp. \bfpage{7680}--\blpage{7688}
(\byear{2018})
\end{bchapter}
\endbibitem

%%% 44
\bibitem{gtan_cvpr2019}
\begin{bchapter}
\bauthor{\bsnm{Long}, \binits{F.}},
\bauthor{\bsnm{Yao}, \binits{T.}},
\bauthor{\bsnm{Qiu}, \binits{Z.}},
\bauthor{\bsnm{Tian}, \binits{X.}},
\bauthor{\bsnm{Luo}, \binits{J.}},
\bauthor{\bsnm{Mei}, \binits{T.}}:
\bctitle{Gaussian temporal awareness networks for action localization}.
In: \bbtitle{CVPR},
pp. \bfpage{344}--\blpage{353}
(\byear{2019})
\end{bchapter}
\endbibitem

%%% 45
\bibitem{tsi_accv}
\begin{bchapter}
\bauthor{\bsnm{Liu}, \binits{S.}},
\bauthor{\bsnm{Zhao}, \binits{X.}},
\bauthor{\bsnm{Su}, \binits{H.}},
\bauthor{\bsnm{Hu}, \binits{Z.}}:
\bctitle{Tsi: Temporal scale invariant network for action proposal generation}.
In: \bbtitle{ACCV}
(\byear{2020})
\end{bchapter}
\endbibitem

%%% 46
\bibitem{bai2020boundary}
\begin{bchapter}
\bauthor{\bsnm{Bai}, \binits{Y.}},
\bauthor{\bsnm{Wang}, \binits{Y.}},
\bauthor{\bsnm{Tong}, \binits{Y.}},
\bauthor{\bsnm{Yang}, \binits{Y.}},
\bauthor{\bsnm{Liu}, \binits{Q.}},
\bauthor{\bsnm{Liu}, \binits{J.}}:
\bctitle{Boundary content graph neural network for temporal action proposal
  generation}.
In: \bbtitle{ECCV},
pp. \bfpage{121}--\blpage{137}
(\byear{2020}).
\bcomment{Springer}
\end{bchapter}
\endbibitem

%%% 47
\bibitem{tan2021relaxed}
\begin{botherref}
\oauthor{\bsnm{Tan}, \binits{J.}},
\oauthor{\bsnm{Tang}, \binits{J.}},
\oauthor{\bsnm{Wang}, \binits{L.}},
\oauthor{\bsnm{Wu}, \binits{G.}}:
Relaxed transformer decoders for direct action proposal generation.
ICCV
(2021)
\end{botherref}
\endbibitem

%%% 48
\bibitem{MaskRCNN_ICCV17}
\begin{bchapter}
\bauthor{\bsnm{He}, \binits{K.}},
\bauthor{\bsnm{Gkioxari}, \binits{G.}},
\bauthor{\bsnm{Dollar}, \binits{P.}},
\bauthor{\bsnm{Girshick}, \binits{R.}}:
\bctitle{Mask r-cnn}.
In: \bbtitle{ICCV}
(\byear{2017})
\end{bchapter}
\endbibitem

%%% 49
\bibitem{sennrich-etal-2016-neural}
\begin{bchapter}
\bauthor{\bsnm{Sennrich}, \binits{R.}},
\bauthor{\bsnm{Haddow}, \binits{B.}},
\bauthor{\bsnm{Birch}, \binits{A.}}:
\bctitle{Neural machine translation of rare words with subword units}.
In: \bbtitle{Proceedings of the 54th Annual Meeting of the Association for
  Computational Linguistics (Volume 1: Long Papers)},
pp. \bfpage{1715}--\blpage{1725}.
\bpublisher{Association for Computational Linguistics},
\blocation{Berlin, Germany}
(\byear{2016}).
\doiurl{10.18653/v1/P16-1162}.
\burl{https://aclanthology.org/P16-1162}
\end{bchapter}
\endbibitem

%%% 50
\bibitem{dosovitskiy2020image}
\begin{botherref}
\oauthor{\bsnm{Dosovitskiy}, \binits{A.}},
\oauthor{\bsnm{Beyer}, \binits{L.}}, et al.:
An image is worth 16x16 words: Transformers for image recognition at scale.
CVPR
(2021)
\end{botherref}
\endbibitem

%%% 51
\bibitem{adahard_eccv2018}
\begin{bchapter}
\bauthor{\bsnm{Malinowski}, \binits{M.}},
\bauthor{\bsnm{Doersch}, \binits{C.}},
\bauthor{\bsnm{Santoro}, \binits{A.}},
\bauthor{\bsnm{Battaglia}, \binits{P.}}:
\bctitle{Learning visual question answering by bootstrapping hard attention}.
In: \bbtitle{ECCV},
pp. \bfpage{3}--\blpage{20}
(\byear{2018})
\end{bchapter}
\endbibitem

%%% 52
\bibitem{khoavo_aei_bmvc21}
\begin{bchapter}
\bauthor{\bsnm{Vo}, \binits{K.}},
\bauthor{\bsnm{Joo}, \binits{H.}},
\bauthor{\bsnm{Yamazaki}, \binits{K.}},
\bauthor{\bsnm{Truong}, \binits{S.}},
\bauthor{\bsnm{Kitani}, \binits{K.}},
\bauthor{\bsnm{Tran}, \binits{M.-T.}},
\bauthor{\bsnm{Le}, \binits{N.}}:
\bctitle{Aei: Actors-environment interaction with adaptive attention for
  temporal action proposals generation}.
In: \bbtitle{32nd British Machine Vision Conference 2021, {BMVC} 2021, Virtual
  Event, UK, November 22-25, 2021}
(\byear{2021}).
\burl{https://www.bmvc2021-virtualconference.com/assets/papers/1095.pdf}
\end{bchapter}
\endbibitem

%%% 53
\bibitem{SoftNMS}
\begin{bchapter}
\bauthor{\bsnm{Bodla}, \binits{N.}},
\bauthor{\bsnm{Singh}, \binits{B.}},
\bauthor{\bsnm{Chellappa}, \binits{R.}},
\bauthor{\bsnm{Davis}, \binits{L.S.}}:
\bctitle{Soft-nms -- improving object detection with one line of code}.
In: \bbtitle{ICCV}
(\byear{2017})
\end{bchapter}
\endbibitem

%%% 54
\bibitem{NMS}
\begin{bchapter}
\bauthor{\bsnm{{Neubeck}}, \binits{A.}},
\bauthor{\bsnm{{Van Gool}}, \binits{L.}}:
\bctitle{Efficient non-maximum suppression}.
In: \bbtitle{ICPR},
vol. \bseriesno{3},
pp. \bfpage{850}--\blpage{855}
(\byear{2006})
\end{bchapter}
\endbibitem

%%% 55
\bibitem{damen2021rescaling}
\begin{botherref}
\oauthor{\bsnm{Damen}, \binits{D.}},
\oauthor{\bsnm{Doughty}, \binits{H.}}, et al.:
Rescaling egocentric vision: Collection, pipeline and challenges for
  epic-kitchens-100.
IJVC,
1--23
(2021)
\end{botherref}
\endbibitem

%%% 56
\bibitem{MR_eccv2020}
\begin{bchapter}
\bauthor{\bsnm{Zhao}, \binits{P.}},
\bauthor{\bsnm{Xie}, \binits{L.}},
\bauthor{\bsnm{Ju}, \binits{C.}},
\bauthor{\bsnm{Zhang}, \binits{Y.}},
\bauthor{\bsnm{Wang}, \binits{Y.}},
\bauthor{\bsnm{Tian}, \binits{Q.}}:
\bctitle{Bottom-up temporal action localization with mutual regularization}.
In: \bbtitle{ECCV},
pp. \bfpage{539}--\blpage{555}
(\byear{2020}).
\bcomment{Springer}
\end{bchapter}
\endbibitem

%%% 57
\bibitem{TCN}
\begin{bchapter}
\bauthor{\bsnm{Dai}, \binits{X.}},
\bauthor{\bsnm{Singh}, \binits{B.}},
\bauthor{\bsnm{Zhang}, \binits{G.}},
\bauthor{\bsnm{Davis}, \binits{L.S.}},
\bauthor{\bsnm{Qiu~Chen}, \binits{Y.}}:
\bctitle{Temporal context network for activity localization in videos}.
In: \bbtitle{ICCV}
(\byear{2017})
\end{bchapter}
\endbibitem

%%% 58
\bibitem{MSRA}
\begin{bchapter}
\bauthor{\bsnm{Yao}, \binits{T.}},
\bauthor{\bsnm{Li}, \binits{Y.}},
\bauthor{\bsnm{Qiu}, \binits{Z.}},
\bauthor{\bsnm{Long}, \binits{F.}},
\bauthor{\bsnm{Pan}, \binits{Y.}},
\bauthor{\bsnm{Li}, \binits{D.}},
\bauthor{\bsnm{Mei}, \binits{T.}}:
\bctitle{Msr asia msm at activitynet challenge 2017: Trimmed action
  recognition, temporal action proposals and densecaptioning events in videos}.
In: \bbtitle{CVPR Workshops}
(\byear{2017})
\end{bchapter}
\endbibitem

%%% 59
\bibitem{SSTAD_BMVC17}
\begin{bchapter}
\bauthor{\bsnm{Buch}, \binits{S.}},
\bauthor{\bsnm{Escorcia}, \binits{V.}},
\bauthor{\bsnm{Ghanem}, \binits{B.}},
\bauthor{\bsnm{Fei-Fei}, \binits{L.}},
\bauthor{\bsnm{Niebles}, \binits{J.C.}}:
\bctitle{End-to-end, single-stream temporal action detection in untrimmed
  videos}.
In: \bbtitle{BMVC}
(\byear{2017})
\end{bchapter}
\endbibitem

%%% 60
\bibitem{SRG}
\begin{botherref}
\oauthor{\bsnm{{Eun}}, \binits{H.}},
\oauthor{\bsnm{{Lee}}, \binits{S.}},
\oauthor{\bsnm{{Moon}}, \binits{J.}},
\oauthor{\bsnm{{Park}}, \binits{J.}},
\oauthor{\bsnm{{Jung}}, \binits{C.}},
\oauthor{\bsnm{{Kim}}, \binits{C.}}:
Srg: Snippet relatedness-based temporal action proposal generator.
IEEE Transactions on Circuits and Systems for Video Technology,
1--1
(2019)
\end{botherref}
\endbibitem

%%% 61
\bibitem{wang2021self}
\begin{bchapter}
\bauthor{\bsnm{Wang}, \binits{X.}},
\bauthor{\bsnm{Zhang}, \binits{S.}},
\bauthor{\bsnm{Qing}, \binits{Z.}},
\bauthor{\bsnm{Shao}, \binits{Y.}},
\bauthor{\bsnm{Gao}, \binits{C.}},
\bauthor{\bsnm{Sang}, \binits{N.}}:
\bctitle{Self-supervised learning for semi-supervised temporal action
  proposal}.
In: \bbtitle{CVPR},
pp. \bfpage{1905}--\blpage{1914}
(\byear{2021})
\end{bchapter}
\endbibitem

%%% 62
\bibitem{qing2021temporal}
\begin{bchapter}
\bauthor{\bsnm{Qing}, \binits{Z.}},
\bauthor{\bsnm{Su}, \binits{H.}},
\bauthor{\bsnm{Gan}, \binits{W.}},
\bauthor{\bsnm{Wang}, \binits{D.}},
\bauthor{\bsnm{Wu}, \binits{W.}},
\bauthor{\bsnm{Wang}, \binits{X.}},
\bauthor{\bsnm{Qiao}, \binits{Y.}},
\bauthor{\bsnm{Yan}, \binits{J.}},
\bauthor{\bsnm{Gao}, \binits{C.}},
\bauthor{\bsnm{Sang}, \binits{N.}}:
\bctitle{Temporal context aggregation network for temporal action proposal
  refinement}.
In: \bbtitle{CVPR},
pp. \bfpage{485}--\blpage{494}
(\byear{2021})
\end{bchapter}
\endbibitem

%%% 63
\bibitem{zheng2021boundary}
\begin{botherref}
\oauthor{\bsnm{Zheng}, \binits{J.}},
\oauthor{\bsnm{Chen}, \binits{D.}},
\oauthor{\bsnm{Hu}, \binits{H.}}:
Boundary adjusted network based on cosine similarity for temporal action
  proposal generation.
Neural Processing Letters,
1--16
(2021)
\end{botherref}
\endbibitem

%%% 64
\bibitem{cocodataset}
\begin{bchapter}
\bauthor{\bsnm{Lin}, \binits{T.-Y.}},
\bauthor{\bsnm{Maire}, \binits{M.}},
\bauthor{\bsnm{Belongie}, \binits{S.}},
\bauthor{\bsnm{Hays}, \binits{J.}},
\bauthor{\bsnm{Perona}, \binits{P.}},
\bauthor{\bsnm{Ramanan}, \binits{D.}},
\bauthor{\bsnm{Dollar}, \binits{P.}},
\bauthor{\bsnm{Zitnick}, \binits{L.}}:
\bctitle{Microsoft coco: Common objects in context}.
In: \bbtitle{ECCV}
(\byear{2014})
\end{bchapter}
\endbibitem

%%% 65
\bibitem{gao2020accurate}
\begin{bchapter}
\bauthor{\bsnm{Gao}, \binits{J.}},
\bauthor{\bsnm{Shi}, \binits{Z.}},
\bauthor{\bsnm{Wang}, \binits{G.}},
\bauthor{\bsnm{Li}, \binits{J.}},
\bauthor{\bsnm{Yuan}, \binits{Y.}},
\bauthor{\bsnm{Ge}, \binits{S.}},
\bauthor{\bsnm{Zhou}, \binits{X.}}:
\bctitle{Accurate temporal action proposal generation with relation-aware
  pyramid network}.
In: \bbtitle{AAAI},
vol. \bseriesno{34},
pp. \bfpage{10810}--\blpage{10817}
(\byear{2020})
\end{bchapter}
\endbibitem

%%% 66
\bibitem{pgcn_cvpr2020}
\begin{bchapter}
\bauthor{\bsnm{Zeng}, \binits{R.}},
\bauthor{\bsnm{Huang}, \binits{W.}},
\bauthor{\bsnm{Tan}, \binits{M.}},
\bauthor{\bsnm{Rong}, \binits{Y.}},
\bauthor{\bsnm{Zhao}, \binits{P.}},
\bauthor{\bsnm{Huang}, \binits{J.}},
\bauthor{\bsnm{Gan}, \binits{C.}}:
\bctitle{Graph convolutional networks for temporal action localization}.
In: \bbtitle{ICCV},
pp. \bfpage{7094}--\blpage{7103}
(\byear{2019})
\end{bchapter}
\endbibitem

%%% 67
\bibitem{action_protocol}
\begin{botherref}
\oauthor{\bsnm{Xiong}, \binits{Y.}},
\oauthor{\bsnm{Wang}, \binits{L.}},
\oauthor{\bsnm{Wang}, \binits{Z.}},
\oauthor{\bsnm{Zhang}, \binits{B.}},
\oauthor{\bsnm{Song}, \binits{H.}},
\oauthor{\bsnm{Li}, \binits{W.}},
\oauthor{\bsnm{Lin}, \binits{D.}},
\oauthor{\bsnm{Qiao}, \binits{Y.}},
\oauthor{\bsnm{Gool}, \binits{L.V.}},
\oauthor{\bsnm{Tang}, \binits{X.}}:
{CUHK} {\&} {ETHZ} {\&} {SIAT} submission to activitynet challenge 2016.
CoRR
\textbf{abs/1608.00797}
(2016)
\end{botherref}
\endbibitem

%%% 68
\bibitem{untrimmetNet}
\begin{bchapter}
\bauthor{\bsnm{Wang}, \binits{L.}},
\bauthor{\bsnm{Xiong}, \binits{Y.}},
\bauthor{\bsnm{Lin}, \binits{D.}},
\bauthor{\bsnm{Van~Gool}, \binits{L.}}:
\bctitle{Untrimmednets for weakly supervised action recognition and detection}.
In: \bbtitle{CVPR}
(\byear{2017})
\end{bchapter}
\endbibitem

%%% 69
\bibitem{zhao2020bottom}
\begin{bchapter}
\bauthor{\bsnm{Zhao}, \binits{P.}},
\bauthor{\bsnm{Xie}, \binits{L.}},
\bauthor{\bsnm{Ju}, \binits{C.}},
\bauthor{\bsnm{Zhang}, \binits{Y.}},
\bauthor{\bsnm{Wang}, \binits{Y.}},
\bauthor{\bsnm{Tian}, \binits{Q.}}:
\bctitle{Bottom-up temporal action localization with mutual regularization}.
In: \bbtitle{ECCV},
pp. \bfpage{539}--\blpage{555}
(\byear{2020}).
\bcomment{Springer}
\end{bchapter}
\endbibitem

\end{thebibliography}


\newpage
\section*{Appendix: Notations}
The following is a table that summarizes and briefly describes important symbols that are used throughout this manuscript:
\begin{table}[!ht]
    %\resizebox{0.8\linewidth}{!}{
    \begin{tabular}{c|l}
        Symbol & Description \\ \hline
        $\mathcal{V}$ & A sequence of all frames from the input video.\\ \hline
        $N$ & Total number of frames in the input video.\\ \hline
        $\delta$ & Length of a snippet, a sub-set of consecutive frames. \\ \hline
        $T$ & Total number of snippets from the input video.\\ \hline
        $s_i$ & The snippet at index $i\in T$.\\ \hline
        $I$ & The frame in the center of snippet $s_i$.\\ \hline
        $\mathcal{B}$ & A set of human bounding boxes detected in center frame $I$ of snippet $s_i$.\\ \hline
        $\phi(.)$ & An encoding function to encode a snippet $s_i$ into a feature vector.\\ \hline
        $\mathcal{F}^\mathcal{M}$ & A feature map extracted by a backbone network.\\ \hline
        $\mathcal{F}^a$ & A set of actors features in snippet $s_i$.\\ \hline
        $\mathcal{F}^o$ & A set of objects features in snippet $s_i$.\\ \hline
        $\mathcal{T}$ & Vocabulary used in Objects Beholder.\\ \hline
        $D$ & Total number of words in vocabulary $\mathcal{T}$.\\ \hline
        $\mathcal{T}^e$ & A set of embedded features for every word in $mathcal{T}$.\\ \hline
        $I^e$ & Embedded feature representing center frame $I$ of $s_i$.\\ \hline
        $K$ & A hyper-parameter, defines maximum number of words in $\mathcal{T}$ to be selected.\\ \hline
        $\mathcal{F}^o$ & Top $K$ embedded features in $\mathcal{T}^e$ that is best matched with $I$.\\ \hline
        $\hat{f}^e$ & Encoded feature of $f^e$, used in AAM.\\ \hline
        $\hat{\mathcal{F}}^a$ & Encoded feature of every actors feature in $\mathcal{F}^a$, used in AAM.\\ \hline
        $H^a$ & A set of scores of every actor feature in $\mathcal{F}^a$.\\ \hline
        $\tau$ & An adaptive threshold to filter out actor features that has lower scores.\\ \hline
        $\tilde{\mathcal{F}}^a$ & A set of main actor features that are selected.\\ \hline
        $f^e$ & Output feature vector of Environment Beholder.\\ \hline
        $f^a$ & Output feature vector of Actors Beholder.\\ \hline
        $f^o$ & Output feature vector of Objects Beholder.\\ \hline
        $\mathcal{F}^{aoe}$ & A stack of $f^a,f^o,f^e$ to serve Actors-Objects-Environment Beholder.\\ \hline
        $f^i$ & Output feature vector of Actors-Objects-Environment Beholder.\\ \hline
        $\mathcal{L}_s$ & Loss function to optimize for starting points classification.\\ \hline
        $L_s$ & Labels of starting points in the input.\\ \hline
        $\mathcal{L}_e$ & Loss function to optimize for ending points classification.\\ \hline
        $L_e$ & Labels of ending points in the input video.\\ \hline
        $\mathcal{L}_{act}$ & Loss function to optimize for the actions classification and regression.\\ \hline
        $L_{D}$ & Labels for the actions classification and regression.\\ \hline
        $\mathcal{L}_{wb}$ & Weighted binary cross-entropy loss.\\ \hline
        $\mathcal{L}_{2}$ & Mean squared error loss.\\ \hline
    \end{tabular}
    %}
    \caption{Descriptions of symbols used in our paper.}
    \label{tab:notations}
\end{table}

\end{document}

\title[AOE-Network - Supplementary Material]{Supplementary Material 

Actors-Objects-Environment: Entities Interactions Modeling with Adaptive Attention Mechanism for Temporal Action Proposals Generation}


\author*[1]{\mbox{\fnm{Khoa}~\sur{Vo}}}\email{khoavoho@uark.edu}

\author[1]{\mbox{\fnm{Sang}~\sur{Truong}}}\email{sangt@uark.edu}

\author[1]{\mbox{\fnm{Kashu}~\sur{Yamazaki}}}\email{kyamazak@uark.edu}

\author[2]{\fnm{Biksha}~\sur{Raj}}\email{bhiksha@cs.cmu.edu}

\author[3,4]{\\\fnm{Minh-Triet}~\sur{Tran}}\email{tmtriet@hcmus.edu.vn}

\author[1]{\fnm{Ngan}~\sur{Le}}\email{thile@uark.edu}

\affil*[1]{\orgdiv{AICV Lab}, \orgname{University of Arkansas}, \orgaddress{\city{Fayetteville}, \state{Arkansas}, \country{USA}}}

\affil[2]{\orgname{Carnegie Mellon University}, \orgaddress{\city{Pittsburgh}, \state{Pennsylvania}, \country{USA}}}

\affil[3]{\orgname{University of Science}, \orgaddress{\city{Ho Chi Minh City}, \country{Vietnam}}}

\affil[4]{\orgname{Vietnam National University}, \orgaddress{\city{Ho Chi Minh City}, \country{Vietnam}}}


\maketitle

\section{Source code of AOE-Net}

Please use the below URL to download source code (and its pre-trained models) due to its large size:

%{\small\noindent\url{https://drive.google.com/drive/folders/1_r3QkZYteQTrGEkwgT8aKgjlJa8w0fZL?usp=sharing}}
{\small\noindent\url{https://drive.google.com/file/d/1cT62LE2gt_KmbcIHmvSxqNhTIrmm5JZd/view?usp=sharing}}

In the source code, please refer to its README.txt for the instructions of usage. If there is any issue with running the code, please feel free to email the corresponding author.

\section{Video features extraction }
%As in the main paper, RGB frames of the input video are initially split into a sequence of snippets, in which each snippet is a sequence of $\delta$ frames ($\delta=16$ for all experiments). Afterwards, our proposed PMR is responsible for extracting important information of the snippet, forming a sequence of feature vectors representing the input video. Each feature vector of a snippet should be capable of containing relevant.
As described in the main submission, a video is represented by a sequence of feature vectors, each of which representing a snippet of $\delta$ frames. There are three types of features that are extracted in this phase, including environment features, actors features, and objects features. All features are extracted from raw video RGB frames with pre-trained backbone networks prior to the training phase of AOE-Net. Furthermore, we do not train the backbone network end-to-end with AOE-Net because of their heavy computational costs.

The environment feature and actors features are both extracted using a single 3D backbone network (e.g. C3D \cite{C3D}, 2Stream \cite{TSN2016ECCV}, SlowFast \cite{SlowFast}). Given a $\delta$-frames input snippet, a feature map $\mathcal{F}^\mathcal{M}$ is extracted from the final convolutional block of the backbone network, then, to extract the environment features, we simply apply average pooling onto $\mathcal{F}^\mathcal{M}$ and obtain a single feature vector representing the whole environment. Meanwhile, to extract the actors features, we firstly utilize a human detector to localize actors appearing in the center frame of input snippet, then, each bounding box is aligned \cite{MaskRCNN_ICCV17} onto $\mathcal{F}^\mathcal{M}$ to extract representation for corresponding actor. Finally, all actors in the snippet are represented by a set of feature vectors.

The objects features are extracted differently than the above two types of features. Instead of employing visual representation, we utilize linguistic representation to model the appearance of objects in the snippet. The main reason for this decision is because objects mainly pop up in very small spatial regions and are tremendously vanished in the feature map $\mathcal{F}^\mathcal{M}$. Therefore, we use CLIP to extract top K most likely objects appearing in the input snippet, with each object is represented by the linguistic feature extracted by the language encoder of CLIP. Finally, top K objects are represented by a set of K linguistic features.

\section{AOE-Net evaluation details}

As stated in the main submission, we have evaluated AOE-Net with both TAPG and TAD tasks on both ActivityNet-1.3 \cite{caba2015activitynet} and THUMOS-14 \cite{THUMOS14} datasets.

\subsection{TAPG evaluation}
Based on previous works \cite{lin2018bsn, bmn, dbg}, Soft-NMS \cite{SoftNMS} has been proved to work better than NMS \cite{NMS} on ActivityNet-1.3, thus,  Soft-NMS has been preferred on ActivityNet-1.3. Whereas both Soft-NMS and NMS have different benefits on THUMOS-14 and they both have been used on THUMOS-14.

On ActivityNet-1.3, the Soft-NMS \cite{SoftNMS} penalizes each proposal $p_i$ with score $s_i$ is defined as follows:
\begin{equation}
\begin{split}
    s_i = \begin{cases}
    s_i & \text{if  } IoU(p_{max}, p_i) < \\ & \theta_1 + \theta_2 ||p_{max} - p_i|| \\
    \alpha s_i  & \text{ otherwise} 
    \end{cases}
\end{split}
\end{equation}
where $\alpha = e^{-\frac{IoU(p_{max},p_i)}{\sigma}}$ and $p_{max}$ is the proposal having maximum IoU with $p_i$. Empirically, the best performance is obtained by $\theta_1=0.5$, $\theta_2=0.4$ and $\sigma=0.4$ on ActivityNet-1.3. Whereas Soft-NMS works best with $\theta_1=0.65, \theta_2=0.$, $\sigma=0.3$ and NMS works best with a threshold setting as 0.45 on THUMOS-14.

\subsection{TAD evaluation}

We use the best performed model trained and evaluated in TAPG to integrate with classification results of \cite{action_protocol} and \cite{untrimmetNet} for ActivityNet-1.3 and THUMOS-14, respectively. In this task, we use Soft-NMS with $\theta_1=\theta_2=0.$ and $\sigma=0.4$ for ActivityNet-1.3 and NMS with a threshold set at 0.45 for THUMOS-14.

\section{\textcolor{red}{Ablation study on $\lambda$ in $\mathcal{L}_{act}$}}
\textcolor{red}{While the loss functions of $\mathcal{L}_s$ and $\mathcal{L}_e$ help optimize our AOE-Net to localize the starting and ending timestamps of groundtruth actions in the video, $\mathcal{L}_{act}$ optimizes our AOE-Net to correctly pair those starting and ending timestamps. $\mathcal{L}_{act}$ is defined in Eq. (9) of our manuscript, which is as follows:}
\begin{equation*}
    \textcolor{red}{\mathcal{L}_{act} = \mathcal{L}_{wb}(P,L) + \lambda \mathcal{L}_2(P,L)}
\end{equation*}
\textcolor{red}{where $\mathcal{L}_{wb}$ and $\mathcal{L}_2$ are weighted binary cross-entropy and mean squared error loss functions, respectively, $P$ and $L$ are the prediction and the corresponding groundtruth label.}

\textcolor{red}{$\lambda$ in Eq. (9) plays a role of balancing between the weighted binary log-likelihood loss and the $L_2$ loss in actionness loss $\mathcal{L}_{act}$. It is set to 10 (i.e., $\lambda = 10$) for all of our experiments in ActivityNet-1.3, THUMOS-14 and EPIC-KITCHEN datasets as we followed BMN~\cite{bmn} and BSN~\cite{lin2018bsn}.
%Fortunately, this setting shows very desirable results. 
To investigate the impact of $\lambda$ in Eq. (9), we have conducted an ablation study on various values of $\lambda \in [1,5,10,15,20,30]$ on TAPG task of THUMOS-14~\cite{THUMOS14} as follows:}
\begin{table}
    \centering
    \begin{tabular}{c|ccccc}
        $\lambda$ & @50 & @100 & @200 & @500 & @1000 \\
        \toprule
        1  & 41.70 & 49.74 & 57.03 & 63.80 & 67.66 \\ % 7000, 0.4, 0.65, 0.65
        %5  & 42.94 & 50.15 & 56.90 & 64.01 & 67.89 \\ % 8000, 0.4, 0.65, 0.65
        5  & 43.87 & 50.21 & 57.09 & 64.10 & 68.00 \\ % 9000, 0.35, 0.65, 0.65
        10 & 44.56 & 50.26 & 57.30 & 64.32 & 68.19 \\
        %15 & 42.88 & 50.70 & 57.04 & 63.86 & 67.86 \\ % 7000, 0.4, 0.65, 0.65
        15 & 44.38 & 50.61 & 57.61 & 64.22 & 67.97 \\ % 9000, 0.35, 0.65, 0.65
        20 & 43.92 & 50.11 & 57.55 & 64.18 & 67.75 \\ % 9000, 0.3, 0.65, 0.65
        30 & 42.95 & 49.87 & 57.12 & 63.87 & 67.59 \\ % 7000, 0.35, 0.65, 0.65
    \end{tabular}
    \caption{\textcolor{red}{Ablation study of various values of $\lambda$ on TAPG task of THUMOS-14~\cite{THUMOS14}.}}
    \label{tab:ablation_lambda}
\end{table}

\textcolor{red}{From the aforementioned ablation study, we observe that the performance of our AOE-Net may reach its peak when lambda is in a range of $5 < \lambda < 15$ in THUMOS-14. As this finding enforces the recommendation of previous works~\cite{bmn, lin2018bsn}, we would suggest to choose value of $\lambda=10$. However, the optimal value for this hyper-parameter may vary a bit on different the datasets.}

\section{\textcolor{red}{Ablation study on $\hat{f}^e$ in Eq.~4}}
\textcolor{red}{Our Eq.~4 was defined as follows:}
\begin{equation}\nonumber
\textcolor{red}{h^a_i = \mid\mid \hat{f}^a_i \oplus \hat{f}^e \mid\mid_2}
\end{equation}
\textcolor{red}{where $\hat{f}^{e} &= MLP_{\theta_e}(f^e)$ and $\hat{f}_i^a$ is a vector of $\hat{F}^{a}$ and $\hat{F}^{a} = MLP_{\theta_a} ({F}^{a})$.}

\textcolor{red}{The intuition behind Eq.~4 is that by broadcasting $\hat{f}^{e}$ into each actor feature of $\hat{f}^{a}_{i}$, the $L^2$ norm of the combination of ($\hat{f}^{e}+ \hat{f}^{a}_{i}$) will reflect to which degree the environment agrees that $f^{a}_{i}$ is the main actor. Therefore, the role of $f^{e}$ is necessary in our AAM.}

\textcolor{red}{To prove the role of $f^e$ in AAM, we have conducted two more ablation studies on Actors Beholder while turning off the other two beholders (i.e., objects an environments) as follows:}
\begin{itemize}
    \item \textcolor{red}{(A1): In AAM, we remove $\hat{f}^e$ from Eq.~4, hence, Eq.~4 becomes $h^a_i=\mid\mid \hat{f}^a_i \mid\mid_2$.}
     \item \textcolor{red}{(A2): In AAM, we keep $\hat{f}^e$ and  Eq.~4 is defined as it was, i.e., $h^a_i=\mid\mid \hat{f}^a_i \oplus \hat{f}^e \mid\mid_2$.}
    %\item (B1): We only enable Objects Beholder and turn off the other two. In AAM, we follow the above procedure by removing $\hat{f}^e$ from Eq.~4.
    %\item (B2): We only enable Objects Beholder and turn off the other two. In AAM, we keep $\hat{f}^e$.
\end{itemize}

\textcolor{red}{The TAPG performance in THUMOS-1.4 of two experiments of with and without environment feature $\hat{f}^e$ is shown in Table \ref{tab:ablation_env}.}

\begin{table}[t]
  \centering
  \begin{tabular}{c|ccccc}
    Exp. & @50 & @100 & @200 & @500 & @1000 \\
    \toprule
    (A1) & 21.70 & 30.12 & 38.27 & 48.77 & 56.19 \\
    (A2) & 26.43 & 36.28 & 45.87 & 56.00 & 59.09 \\ \hline
    % \#1 & 25.96 & 35.14 & 43.48 & 52.37 & 58.47 \\ \hline
    % (b) & 15.24 & 24.08 & 33.86 & 45.88 & 53.71 \\
    % \#3 & 18.06 & 26.68 & 37.14 & 49.28 & 56.99 \\
  \end{tabular}
  \caption{\textcolor{red}{Ablation study of $\hat{f}^e$ in AAM, in the case of using Actors Beholder only.}}
  \label{tab:ablation_env}
\end{table}
\textcolor{red}{As we can see, without environment feature $f^e$ in the AAM, the performance is degraded compared to when $f^e$ is included. Therefore, $f^e$ is proved to have an important role in AAM.}

\section{Qualitative results}
The qualitative results of our proposed AOE-Net network in TAPG task on ActivityNet-1.3 \cite{caba2015activitynet} and THUMOS-14 \cite{THUMOS14} are illustrated in the videos submitted in the supplementary material.

\newpage
\bibliography{sn-bibliography}% common bib file
